%% file: main.tex
\documentclass[letterpaper, 10 pt, conference]{ieeeconf}
\IEEEoverridecommandlockouts
\let\labelindent\relax
\input{configs}

\title{\LARGE \bf Downwash-aware Control Allocation for Over-actuated UAV Platforms\vspace{-9pt}}

\author{Yao Su$^{1*}$, Chi Chu$^{1,2*}$, Meng Wang$^{1}$, Jiarui Li$^{1,3}$, Liu Yang$^{1,4}$, Yixin Zhu$^{5,6}$, Hangxin Liu$^{1\dagger}$\\%
Project Website: \url{https://marvel-uav.github.io}%
\vspace{-9pt}%
\thanks{\fontsize{7.5}{7.5}\selectfont $^{*}$ Y. Su and C. Chu contributed equally. $\dagger$ Corresponding author. $^{1}$Beijing Institute for General Artificial Intelligence (BIGAI). Emails: {\tt{\{suyao, chuchi, wangmeng, lijiarui, yangliu, liuhx\}@bigai.ai.}} $^{2}$Department of Automation, Tsinghua University. $^{3}$College of Engineering, Peking University. $^{4}$Academy of Arts \& Design, Tsinghua University. $^{5}$Institute for Artificial Intelligence, Peking University. Email: {\tt{yixin.zhu@pku.edu.cn.}} $^{6}$School of Artificial Intelligence, Peking University.}%
}%

\begin{document}

\maketitle
\thispagestyle{empty}
\pagestyle{empty}

\begin{abstract}
Tracking position and orientation independently affords more agile maneuver for over-actuated multirotor \acp{uav} while introducing undesired downwash effects; downwash flows generated by thrust generators may counteract others due to close proximity, which significantly threatens the stability of the platform. The complexity of modeling aerodynamic airflow challenges control algorithms from properly compensating for such a side effect. Leveraging the input redundancies in over-actuated \acp{uav}, we tackle this issue with a novel control allocation framework that considers downwash effects and explores the entire allocation space for an optimal solution. This optimal solution avoids downwash effects while providing high thrust efficiency within the hardware constraints. To the best of our knowledge, ours is the first formal derivation to investigate the downwash effects on over-actuated \acp{uav}. We verify our framework on different hardware configurations in both simulation and experiment.
\end{abstract}

\setstretch{0.98}

\section{Introduction}

Over-actuated \ac{uav} platforms with independent position and orientation tracking provide more agile maneuver compared with traditional multirotors. A straightforward realization is to tilt propellers~\cite{anzai2017multilinked,li2021flexibly,gerber2018twisting,csenkul2016system} and generate thrust forces in non-collinear directions. As a result, many platforms employ \textbf{actively tiltable thrust generators}~\cite{nguyen2018novel,pi2021simple,yu2021over}, achieving higher thrust efficiency and enabling omnidirectional flights~\cite{gerber2018twisting,yu2021over}. 

\begin{figure}[t!]
    \centering
    \begin{subfigure}[b]{\linewidth}
    \centering
        \includegraphics[width=\linewidth]{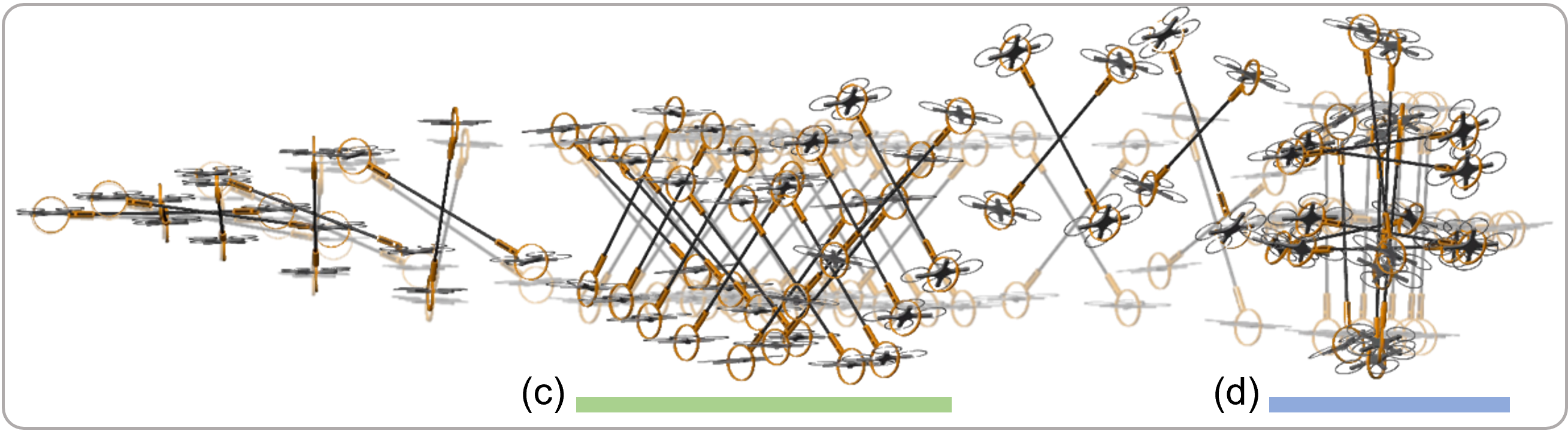}
        \caption{Conventional control allocation framework}
    \end{subfigure}%
    \\
    \begin{subfigure}[b]{\linewidth}
        \centering
        \includegraphics[width=\linewidth]{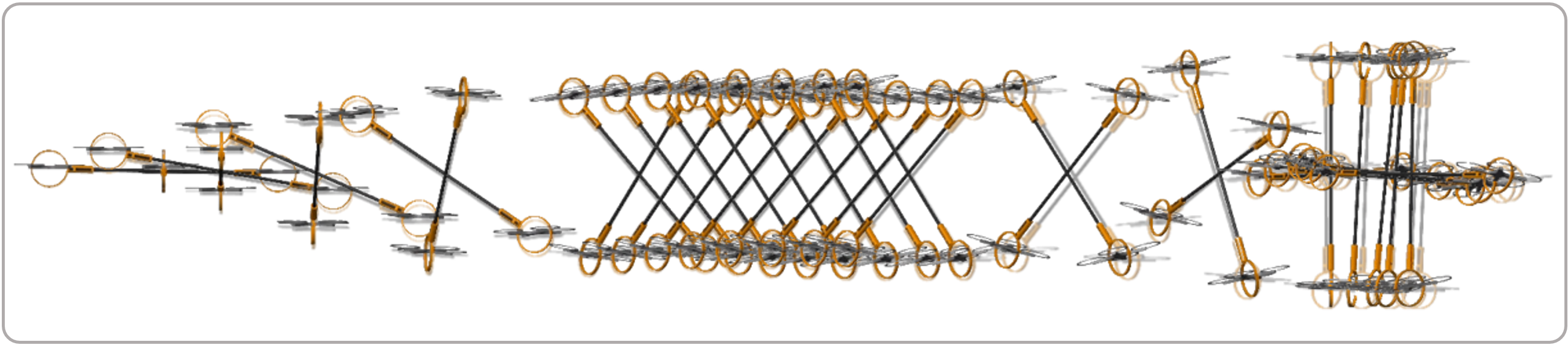}
        \caption{Proposed downwash-aware control allocation framework}
    \end{subfigure}%
    \\
    \begin{subfigure}[b]{0.25\linewidth}
        \centering
        \includegraphics[height=\linewidth]{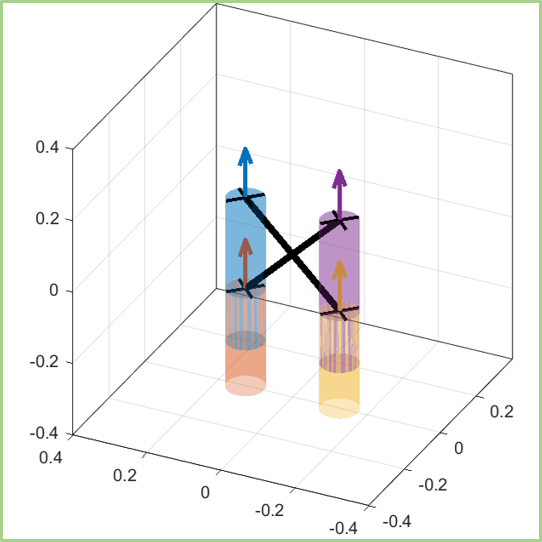}
        \caption{Config. 1}
    \end{subfigure}%
    \begin{subfigure}[b]{0.25\linewidth}
        \centering
        \includegraphics[height=\linewidth]{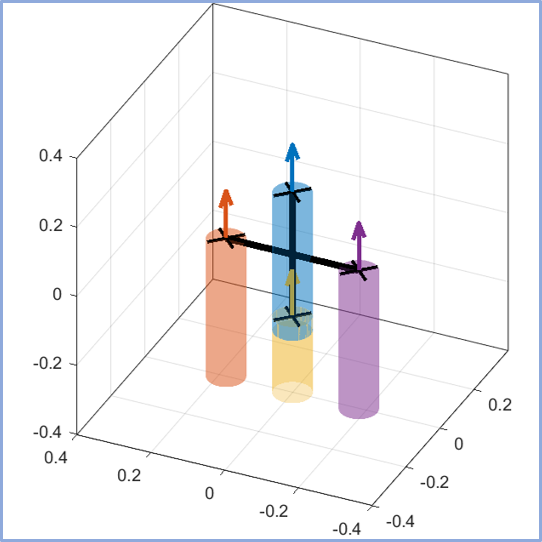}
        \caption{Config. 2}
    \end{subfigure}%
    \begin{subfigure}[b]{0.5\linewidth}
        \centering
        \includegraphics[height=0.55\linewidth]{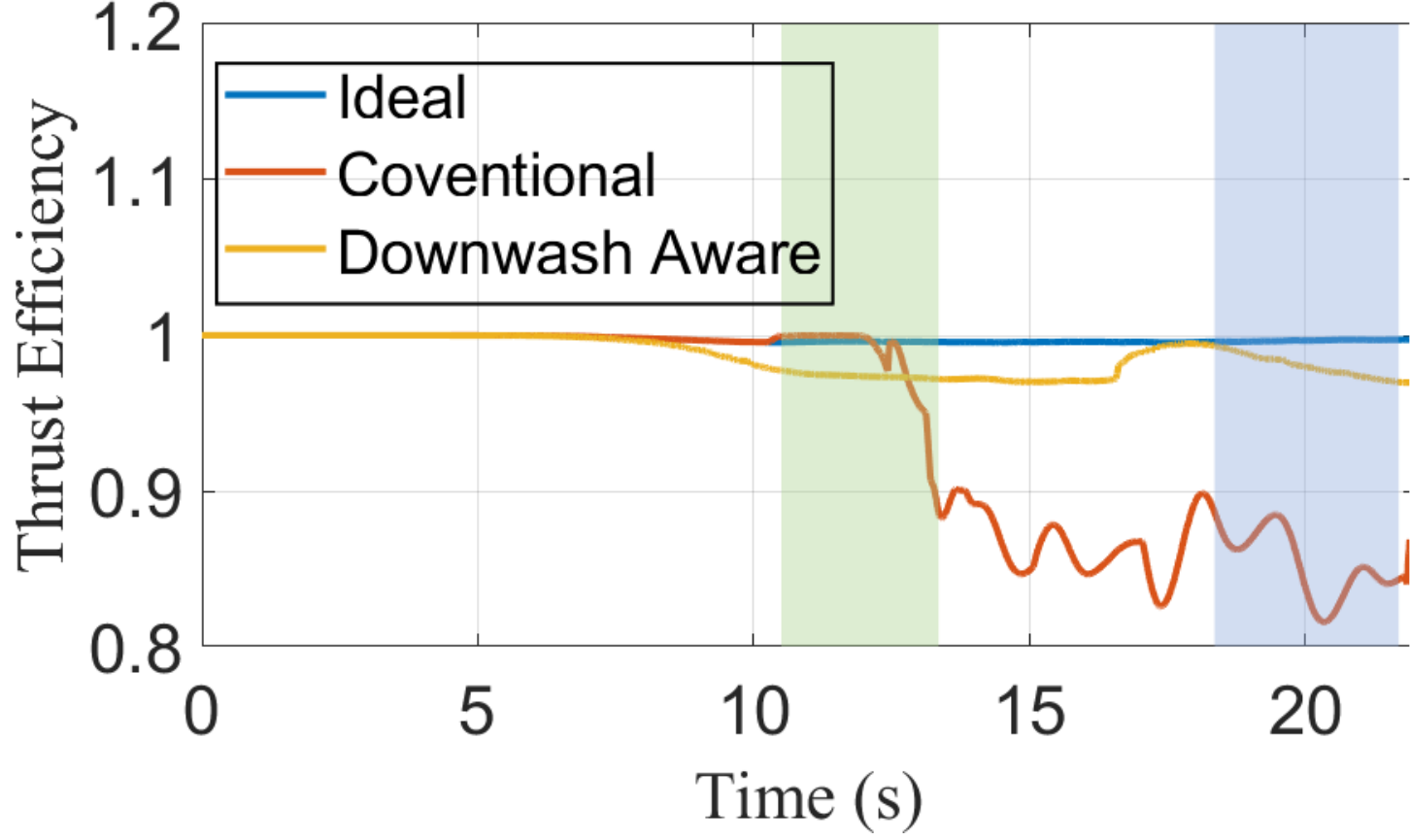}
        \caption{Trust efficiency over the flight}
    \end{subfigure}%
    \caption{\textbf{Comparison between the proposed control allocation framework with conventional ones when tracking the reference trajectory indicated by the light grey.} (a) Conventional control allocation framework fails to track stably as downwash effects appear twice, highlighted in green and blue. Exemplar configurations solved by the conventional control allocation framework may lead to (c) two and (d) one pair of downwash effects, where arrows and cylinders stand for the thrust forces and downwash flows, respectively. (b) The proposed framework avoids downwash effects and thus maintains both stable tracking performance and high trust efficiency over the challenging flight. (e) We further compare the thrust efficiency of ideal (the reference trajectory), conventional, and our proposed allocation frameworks.}
    \label{fig:downwash_illustration}
\end{figure}

Adopting tiltable thrust generators unfortunately also introduces a common side effect\textemdash{}the downwash effect~\cite{michael2010grasp}, which has been rarely studied in the context of over-actuated \acp{uav}. This effect occurs when the airflow generated by one thrust generator/propeller passes through and interacts with the other(s), resulting in deteriorated trajectory tracking performance and lower trust efficiency; see \cref{fig:downwash_illustration} for an illustration. In the literature, the downwash effects are primarily treated by compensation~\cite{su2021fast,lee2021low,zhang2020learning,yang2021learning,yu2022over} or as disturbances to be slowly attenuated by adding integrators into trajectory tracking controller~\cite{su2021nullspace,yu2021over}. However, the former approach needs numerous experimental data to learn the platform-specific compensator, which cannot be generalized to other platforms. The latter solution is slow in response and hence has undesirable transitional behavior (\eg, obvious drop in the flow direction). Critically, both approaches only handle the downwash effect \textit{after} it occurs and are \textit{inefficient} in terms of energy, requiring extra thrusts to compensate.

In this paper, we tackle the downwash effects from a novel control allocation perspective for over-actuated \acp{uav} with actively tiltable thrust generators. Due to input redundancy, there exists an infinite number of solutions to allocate desired force and torque commands to the low-level commands of thrust generators. This observation makes it possible to find a proper allocation such that no air flows would counteract with other thrust generators through a flight, thus reducing or even eliminating downwash effects \textit{beforehand}.

We first incorporate the aerodynamics model for downwash effect analysis and investigate the relationship between \textbf{downwash effect avoidance} and \textbf{thrust efficiency}. Next, we extend our nullspace-based control allocation framework~\cite{su2021nullspace} by adding downwash avoidance constraints and a thrust efficiency index in the objective function. In simulation, we verify the proposed downwash-aware control allocation framework on different over-actuated \ac{uav} platforms. In experiment, we build physical platforms that combine commercial quadcopters with passive gimbal joints as 3-\ac{dof} thrust generators and verify the proposed framework. Collectively, we demonstrate that the proposed framework can fully explore the entire allocation space and find the optimal allocation solution that avoids downwash effects and maintains a high thrust efficiency.

\setstretch{1}

\subsection{Related Work}\label{sec:related}

\textbf{Downwash effects} have recently drawn an increasing attention, primarily on computational models of two \acp{uav}~\cite{khan2013propeller,zheng2018computational,jain2019modeling} to achieve better motion cooperation~\cite{miyazaki2018airborne,brinkman2020post}. Downwash effects for multi-\ac{uav} systems are more challenging to handle. The most straightforward solution is to keep enough safety distance among \acp{uav} to avoid the interference introduced by downwash effects~\cite{preiss2017downwash}. Learning-based method has also been proposed to compensate for the downwash effects among multirobot swarm~\cite{shi2020neural,shi2021neural}. Different from the above work in multi-agent scenarios, we study over-actuated \ac{uav} platforms, wherein several thrust generators are physically connected to a common frame. By developing a centralized control allocation framework, our framework avoids the downwash effects by exploiting input redundancy when generating low-level control commands of thrust generators.

Commanding each actuator given the desired total wrench of the platform, the \textbf{control allocation} of over-actuated \ac{uav} platforms is a constrained nonlinear optimization problem and is generally difficult to solve with high efficiency. Prior work leverages gradient-descent~\cite{ryll2014novel}, force decomposition~\cite{kamel2018voliro}, iterative approach~\cite{zhao2020enhanced}, separation method~\cite{santos2021fast}, and linear approximation~\cite{johansen2004constrained} to reduce the computational complexity. However, none can incorporate input constraints while providing exact solutions with satisfactory efficiency. This limitation was first solved by Su \etal~\cite{su2021nullspace}, who devised a \textit{nullspace-based} control allocation framework; henceforth, we referred to this framework as the \textbf{conventional allocation framework}. This paper extends this framework by incorporating a downwash effect avoidance constraint and adding a thrust efficiency index to the objective function. As a result, various \ac{uav} platforms with 3-\acs{dof} thrust generators~\cite{gerber2018twisting,yu2021over,anzai2017multilinked,li2021flexibly,da2020drone,csenkul2016system} can achieve any arbitrary attitude without downwash effects while maintaining high thrust efficiency along the entire possible configuration space.

\subsection{Overview}

We organize the remainder of the paper as follows. \cref{sec:dynamics} presents the dynamics model of the \ac{uav} system with downwash effect modeling. 
We analyze downwash effects and study the relation between downwash effect avoidance and thrust efficiency in \cref{sec:analysis}. \cref{sec:control} describes the hierarchical control structure and the proposed downwash-aware allocation framework. \cref{sec:setup} and \cref{sec:result} show the simulation and experiment results with comprehensive evaluations. Finally, we conclude the paper in \cref{sec:conclusion}.

\section{Platform Model with Aerodynamics}\label{sec:dynamics}

The over-actuated \ac{uav} system discussed in this paper adopts regular quadcopters with 2-\ac{dof} passive gimbal mechanism, serving as 3-\ac{dof} thrust generators~\cite{yu2021over}. This system has demonstrated various configurations depending on the number of thrust generators and mainframe design~\cite{wang2020walkingbot}, and its dynamics is mathematically equivalent to some seminal platforms~\cite{gerber2018twisting,csenkul2016system,yu2021over,li2021flexibly}.

\subsection{System Frames and Configuration}

\begin{figure}[t!]
    \centering
    \includegraphics[width=\linewidth]{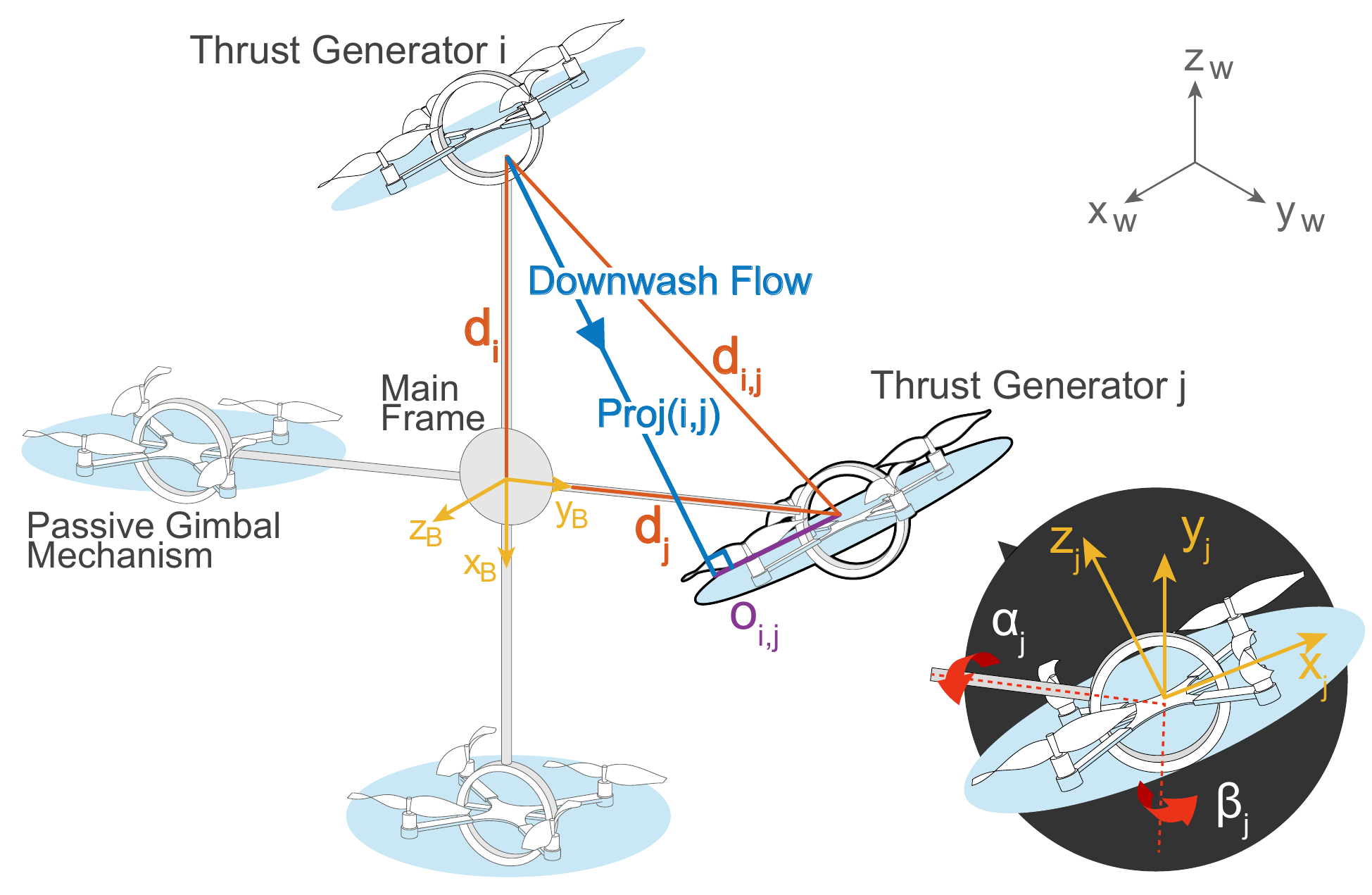}
    \caption{\textbf{Coordinate systems of the over-actuated \ac{uav} platform.} Regular quadcopters are connected to the mainframe by 2-\ac{dof} passive gimbal mechanism, serving as 3-\ac{dof} thrust generators. Each quadcopter generates downwash flow in thrust's opposite direction.}
    \label{fig:platform_downwash}
\end{figure}

\cref{fig:platform_downwash} outlines the system frames and configurations. Let $\mathcal{F}_W$ denote the world coordinate frame and attach the platform frame $\mathcal{F}_B$ to the geometric center of the \ac{uav} platform. We define the central position of the main frame as $\pmb{\xi}=[x,\, y, \,z]^\mathsf{T}$, the attitude in the roll-pitch-yaw convention as $\pmb{\eta}=[\phi,\,\theta,\,\psi]^\mathsf{T}$, and the platform angular velocity in $\mathcal{F}_B$ as $\pmb{\nu} = [p,\,q,\,r]^\mathsf{T}$. Actuator frames $\mathcal{F}_{i}$s are attached to the geometric center of the $i$th 3-\ac{dof} thrust generator. 


\subsection{Platform Dynamics}

The dynamics model of this over-actuated \ac{uav} platform can be described as in Yu \etal~\cite{yu2021over}.
\begin{equation}
    \small
    \begin{bmatrix}
        m \, \prescript{W}{}{\pmb{\ddot{\xi}}} \\
        \prescript{B}{}{\pmb{J}} \, \prescript{B}{}{\pmb{\dot{\nu}}} \\
    \end{bmatrix} = 
    \begin{bmatrix}
        \prescript{W}{B}{\pmb{R}}&0\\
        0&I_3
    \end{bmatrix}
    \pmb{u} + 
    \begin{bmatrix}
        m g \pmb{\hat{z}} \\
        \prescript{B}{}{\pmb{\tau}_g} \\
    \end{bmatrix}
    +\prescript{ext}{}{\pmb{u}},
    \label{eq:dyna}
\end{equation}
where the translational dynamics are expressed in the world frame $\mathcal{F}_W$, whereas the rotational dynamics are described in body-fix frame $\mathcal{F}_B$. $m$ and $\pmb{J}$ are the total mass and inertia matrix of the platform, respectively. $\ddot{\pmb{\xi}}$ and $\dot{\pmb{\nu}}$ are the linear and angular acceleration of the central frame, respectively. $g$ is the acceleration due to gravity, $\prescript{B}{}{\pmb{\tau}_g}$ is the gravity torque due to the displacement of its \ac{com} from the geometric center~\cite{gerber2018twisting}, $\hat{\pmb{z}} = \left[0,\,0,\,1\right]^\mathsf{T}$, and
\begin{equation} 
    \small
    \pmb{u} = 
    \begin{bmatrix}
        \sum_{i=1}^{N} \prescript{B}{i}{\pmb{R}}\,T_i \pmb{\hat{z}} \\
        \sum_{i=1}^{N} (\pmb{d}_i \times \prescript{B}{i}{\pmb{R}}\,T_i \pmb{\hat{z}}) \\
    \end{bmatrix}
    =    
    \begin{bmatrix}
        \pmb{J}_\xi(\pmb{\alpha},\pmb{\beta}) \\ \pmb{J}_\nu(\pmb{\alpha},\pmb{\beta})
    \end{bmatrix}
    \pmb{T},
    \label{eq:bu_u}
\end{equation}
where $T_i$, $\alpha_i$, and $\beta_i$ denote the magnitude of thrust, tilting, and twisting angles of the $i$th thrust generator. $N$ is the number of thrust generators, $\pmb{d}_i$ the distance vector from $\mathcal{F}_B$'s center to each $\mathcal{F}_{i}$, and $\prescript{ext}{}{\pmb{u}}$ the external force/torque input, assumed to be caused by downwash effects.

\subsection{Downwash Effect Modeling}\label{sec:downwashmodel}

As elaborated by Khan \etal~\cite{khan2013propeller}, for the zone of flow establishment (ZFE), the velocity field of a quadcopter follows a Gaussian distribution,
\begin{equation}
    \small
    V(z,r)=V_{\textit{ZFE,max}}(z)\,e^{-\frac{1}{2}\left(\frac{r-R_{m0}}{0.5R_{m0}+0.075(z-z_0-R_0)/K_{\textit{visc}}}\right)^2},
\end{equation}
with
\begin{equation}
    \small
    V_{\textit{ZFE,max}}(z)=V_0\left[c_1-c_2K_{\textit{visc}}(z-z_0)/R_0\right], 
\end{equation}
where $z$ and $r$ are the vertical and radial separations, respectively. $K_{\textit{visc}}$ is the viscosity constant. $z_0$, $R_{0}$, and $V_0$ are the position, contracted radius, and induced velocity of the efflux plane, respectively.  $R_{m0}$ is the radial location of the maximum velocity at each cross-section. $c_1$ and $c_2$ are two parameters, which can be experimentally determined. 

With the model introduced in Jain \etal~\cite{jain2019modeling}, the thrust change caused by oncoming flows for every propeller is estimated:
\begin{equation}
    \small
    {\Delta} t_{i,j}=-b_v\sum_{k=1}^{N}V(z_{i,j,k},r_{i,j,k})\,t_{i,j}, \quad \forall{j=1,\cdots,4}
\end{equation}
where $t_{i,j}$ is the thrust generated by the $j$th propeller of $i$th quadcopter module, defined by: 
\begin{equation}
    \small
    t_{i,j} = K_T \omega_{i,j}^2,
    \label{eq:omega}
\end{equation}
where $\omega_{i,j}$ is the rotational speed, $z_{i,j,k}$ and $r_{i,j,k}$ are the vertical and radial separations between $i$th quadcopter's $j$th propeller and $k$th quadcopter's downwash flow, and $b_v$ is the thrust decay coefficient, obtained experimentally. 

We calculate the $i$th quadcopter's thrust and torque disturbance caused by the downwash effects as in Ruan \etal~\cite{ruan2020independent}:
\begin{equation}
    \small
    \begin{bmatrix}
    {\Delta} T_{i}\\ {\Delta}M_i 
    \end{bmatrix}=
    \begin{bmatrix}
    1  &  1  &  1 & 1 \\
    b &  -b &  -b & b \\
    -b & -b &  b & b \\
    -c_\tau & c_\tau & -c_\tau & c_\tau
    \end{bmatrix}
    \begin{bmatrix}
     {\Delta} t_{i,1}\\ {\Delta} t_{i,2}\\{\Delta} t_{i,3}\\{\Delta} t_{i,4}
    \end{bmatrix},
    \label{eq:quad_input_decouple}
\end{equation}
where ${\Delta}M_i$ affects the low-level attitude control of $i$th quadcopter. $M_i=[M^x_i,M^y_i,M^z_i]^\mathsf{T}$ are the torque outputs in $\mathcal{F}_{i}$. $b$ is a constant defined as $b = {a}/{\sqrt{2}}$, where $a$ is the distance of each propeller to the quadcopter center. $c_\tau$ is a constant defined as $c_\tau = {K_\tau}/{K_T}$, where ${K_\tau}$ is the propeller drag constant, and $K_T$ the standard propeller thrust constant. ${\Delta} T_{i}$ mainly influences the high-level control as external force, and we can have
\begin{equation}
    \small
    \prescript{ext}{}{\pmb{u}}=
    \begin{bmatrix}
       \prescript{W}{B}{\pmb{R}} (\sum_{i=1}^{N} \prescript{B}{i}{\pmb{R}} \, {\Delta} T_i \pmb{\hat{z}})\\
        \sum_{i=1}^{N} (\pmb{d}_i \times \prescript{B}{i}{\pmb{R}}\, {\Delta}T_i \pmb{\hat{z}})
    \end{bmatrix}.
\end{equation}
\cref{sec:result} adopts this downwash effect model for simulation with the parameters acquired from experimental data. 

\section{Downwash Effect Analysis}\label{sec:analysis}

\subsection{Downwash Constraint Derivation}

As shown in \cref{fig:platform_downwash}, the radial distance between $i$th quadcopter's downwash flow and $j$th quadcopter's center is defined as $O_{i,j}$, which be calculated by
\begin{align}
    \small
    O_{i,j}&=\sqrt{\|d_{i,j}\|^2-\|\textit{proj}(i,j)\|^2},\\
    \pmb{d}_{i,j}&=\pmb{d}_j-\pmb{d}_i,\\
    \textit{proj}(i,j)&=\textit{dot}(\pmb{d}_{i,j},\prescript{B}{i}{\pmb{R}}\,\pmb{\hat{z}}),
\end{align}
where \textit{dot} refers to the dot product of two vectors. By having $\prescript{B}{i}{\pmb{R}}$, $O_{i,j}$ is a function of $\alpha_i$ and $\beta_i$. If we build a vector $\pmb{O}(\alpha,\beta)=[O_{1,2}^2;...;O_{N,N-1}^2]\in\mathbb{R}^{N(N-1)\times1}$ by stacking $O_{i,j}^2$, we can calculate a minimum distance vector $\pmb{O}_{\textit{min}}$ to constraint $\pmb{O}$. As a result, the downwash effect avoidance can be achieved by requiring
\begin{equation}
    \small
    \pmb{O}(\alpha,\beta)\,\geq \pmb{O}_{\textit{min}}.
    \label{eq:dist_constraint}
\end{equation}%
\setstretch{0.98}%
Of note, as shown in \cref{alg:downwash}, we need only this constraint when the downwash flows go through other quadcopters in the positive direction. As this inequality constraint is highly nonlinear, we approximately include this constraint into the nullspace-based allocation framework by first-order linearization, to be detailed in \cref{sec:allo}.

\begin{algorithm}[t]
    \small
    \caption{Downwash Constraint Calculation} 
    \label{alg:downwash}  
    \KwData{$d_i, \prescript{B}{i}{\pmb{R}}, N, o_{\textit{min}}$ constant}
    \KwResult {$\pmb{O}_{\textit{min}}$}  
    $i \gets 1, j \gets 2, k \gets 0$\; 
    $\pmb{O}_{\textit{min}} \gets zeros(N(N-1),1)$\;
    \For{$i=1 \cdots N$}{
        \For{$j=1 \cdots N$}{
            \If{$i \neq j$}{
                $k \gets k+1$\;
                $d_{i,j} \gets d_j-d_i$\;
                $\textit{proj}(i,j)\gets \textit{dot}(d_{i,j},\prescript{B}{i}{\pmb{R}}\,\pmb{\hat{z}})$\;
                \eIf{$\textit{proj}(i,j)\leq0$}{
                 $\pmb{O}_{\textit{min}}(k) \gets 0$}
                {
                 $\pmb{O}_{\textit{min}}(k) \gets o_{\textit{min}}^2$
                }
            }
        }
    }
\end{algorithm}

\begin{figure}[t!]    
    \centering
    \begin{subfigure}[b]{0.32\linewidth}
        \centering
        \includegraphics[width=\linewidth,trim=5.1cm 9.2cm 5.1cm 9.2cm, clip]{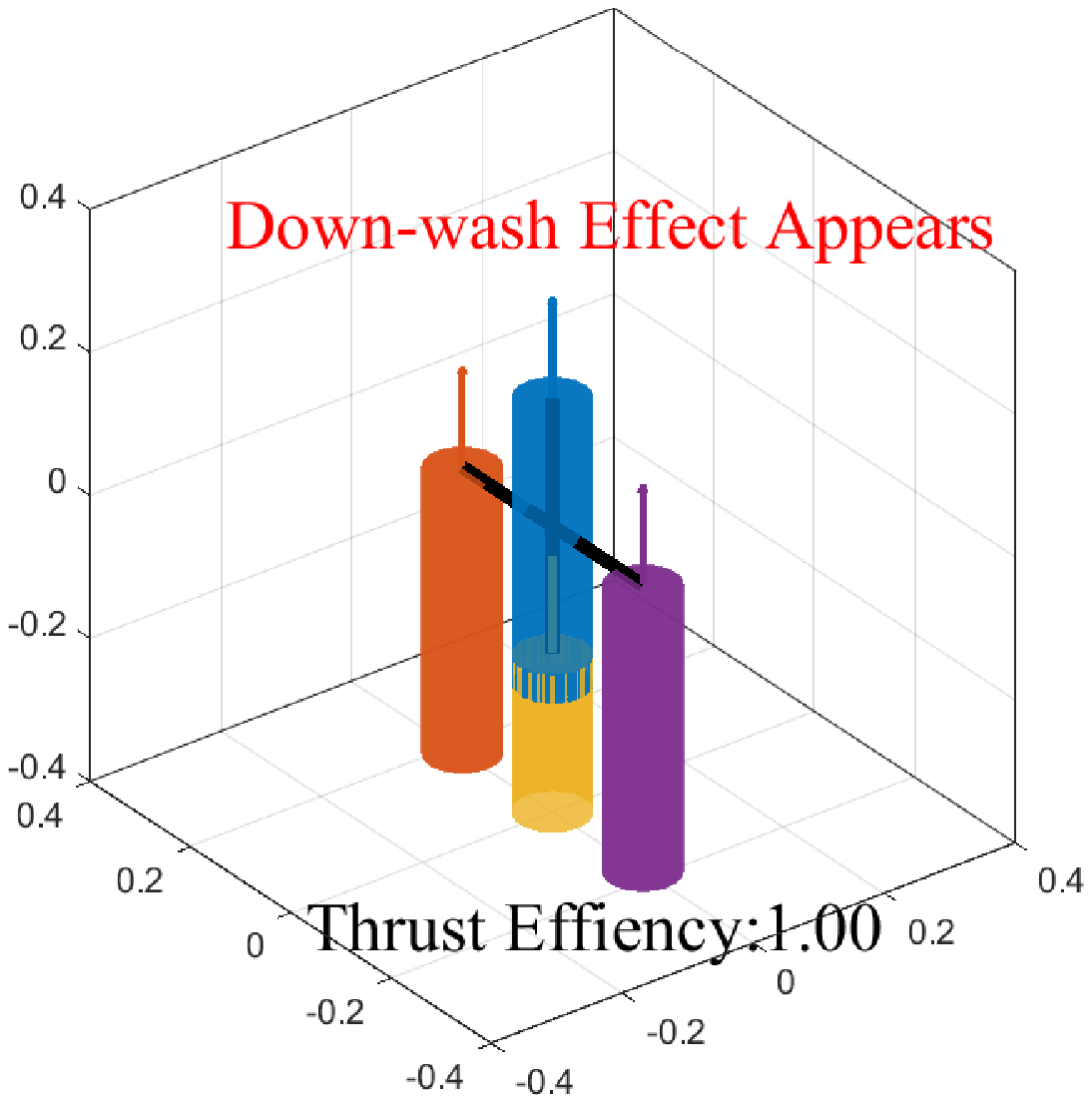}
        \caption{Four: Config. 1}
        \label{fig:quad_1}
    \end{subfigure}
    \begin{subfigure}[b]{0.32\linewidth}
        \centering
        \includegraphics[width=\linewidth,trim=5.1cm 9.2cm 5.1cm 9.2cm, clip]{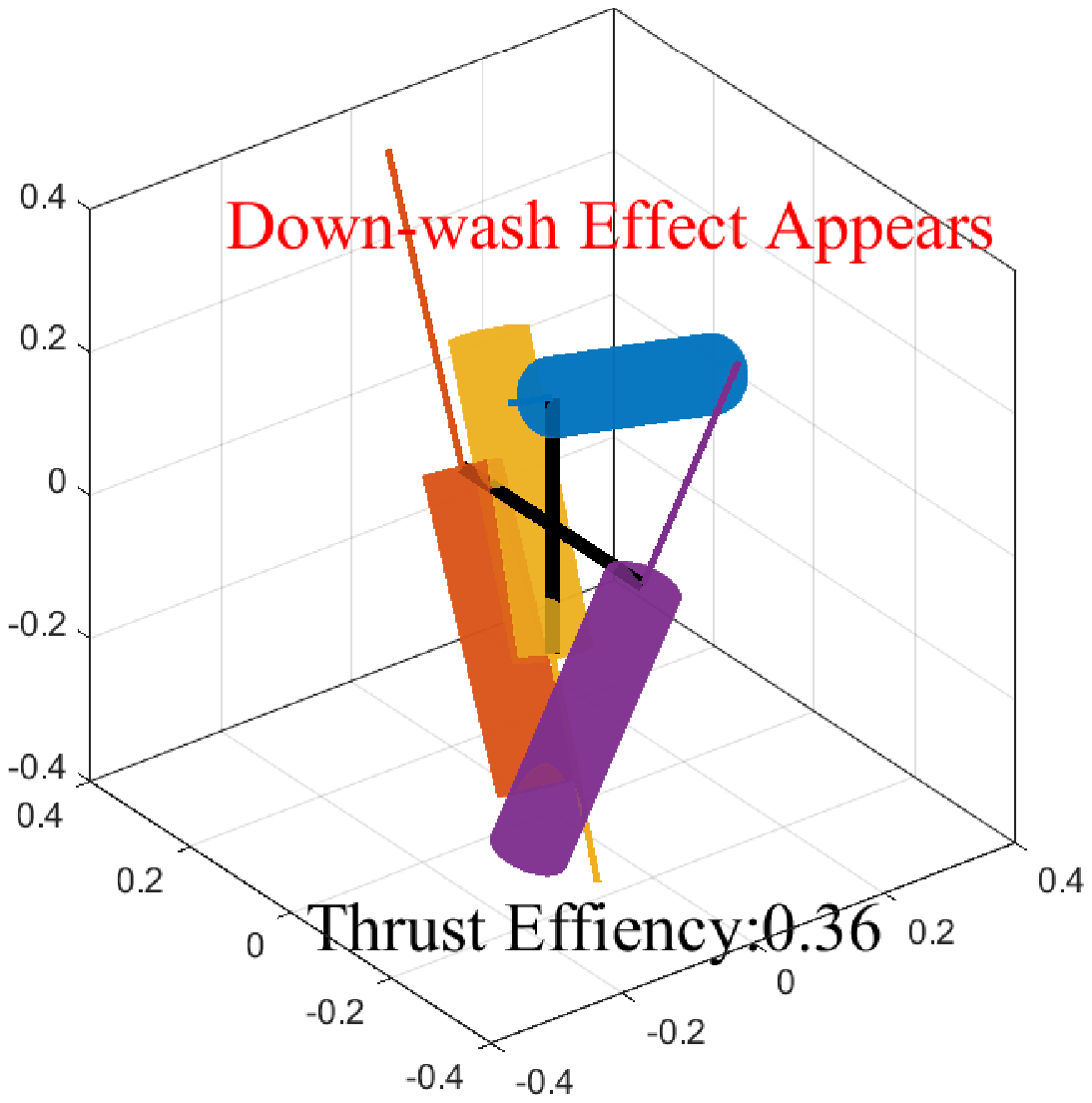}
        \caption{Four: Config. 2}
        \label{fig:quad_2}
    \end{subfigure}
    \begin{subfigure}[b]{0.32\linewidth}
        \centering
        \includegraphics[width=\linewidth,trim=5.1cm 9.2cm 5.1cm 9.2cm, clip]{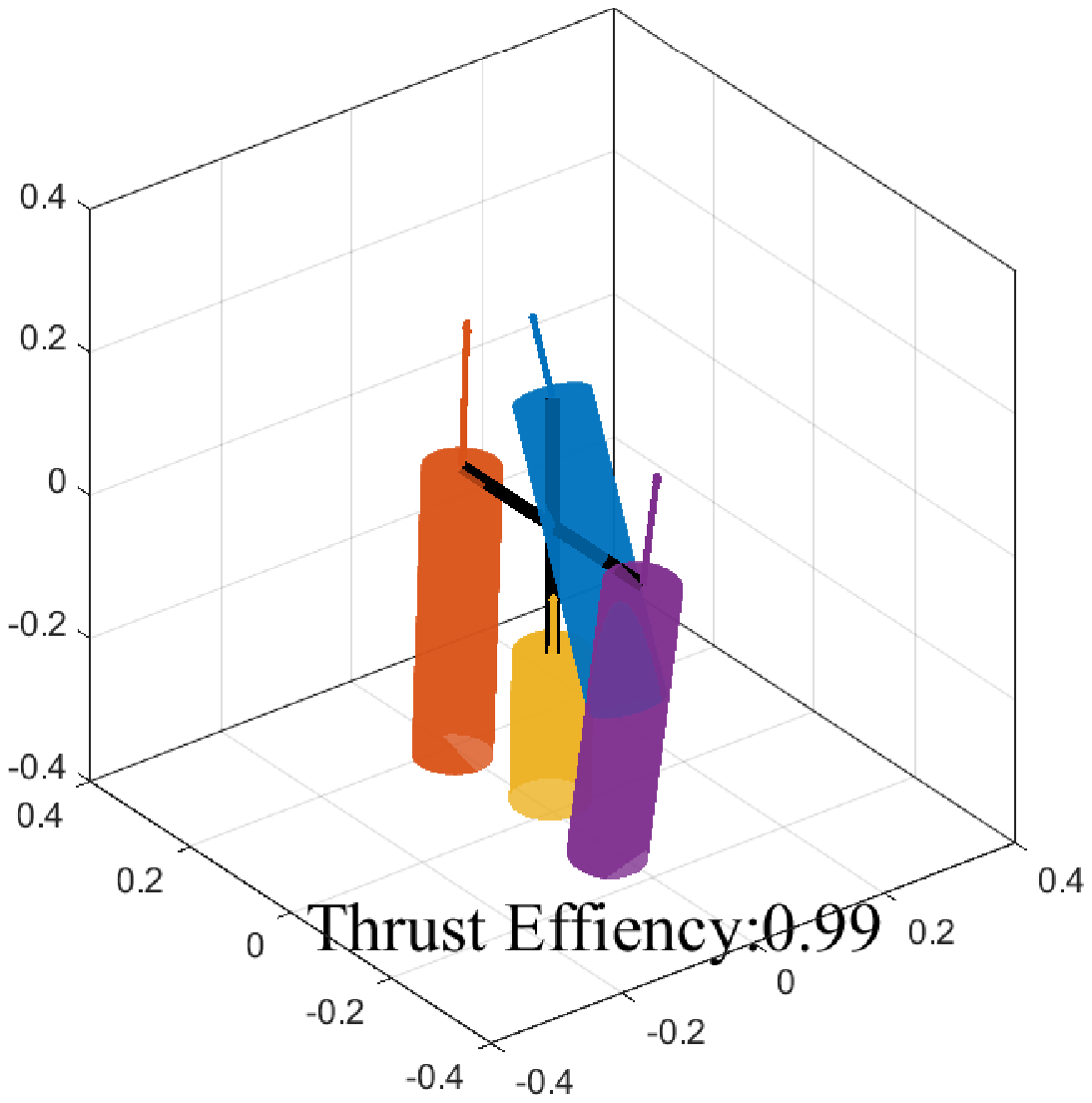}
        \caption{Four: Config. 3}
        \label{fig:quad_3}
    \end{subfigure}\\   
    \begin{subfigure}[b]{0.32\linewidth}
        \centering
        \includegraphics[width=\linewidth,trim=5.1cm 9.2cm 5.1cm 9.2cm, clip]{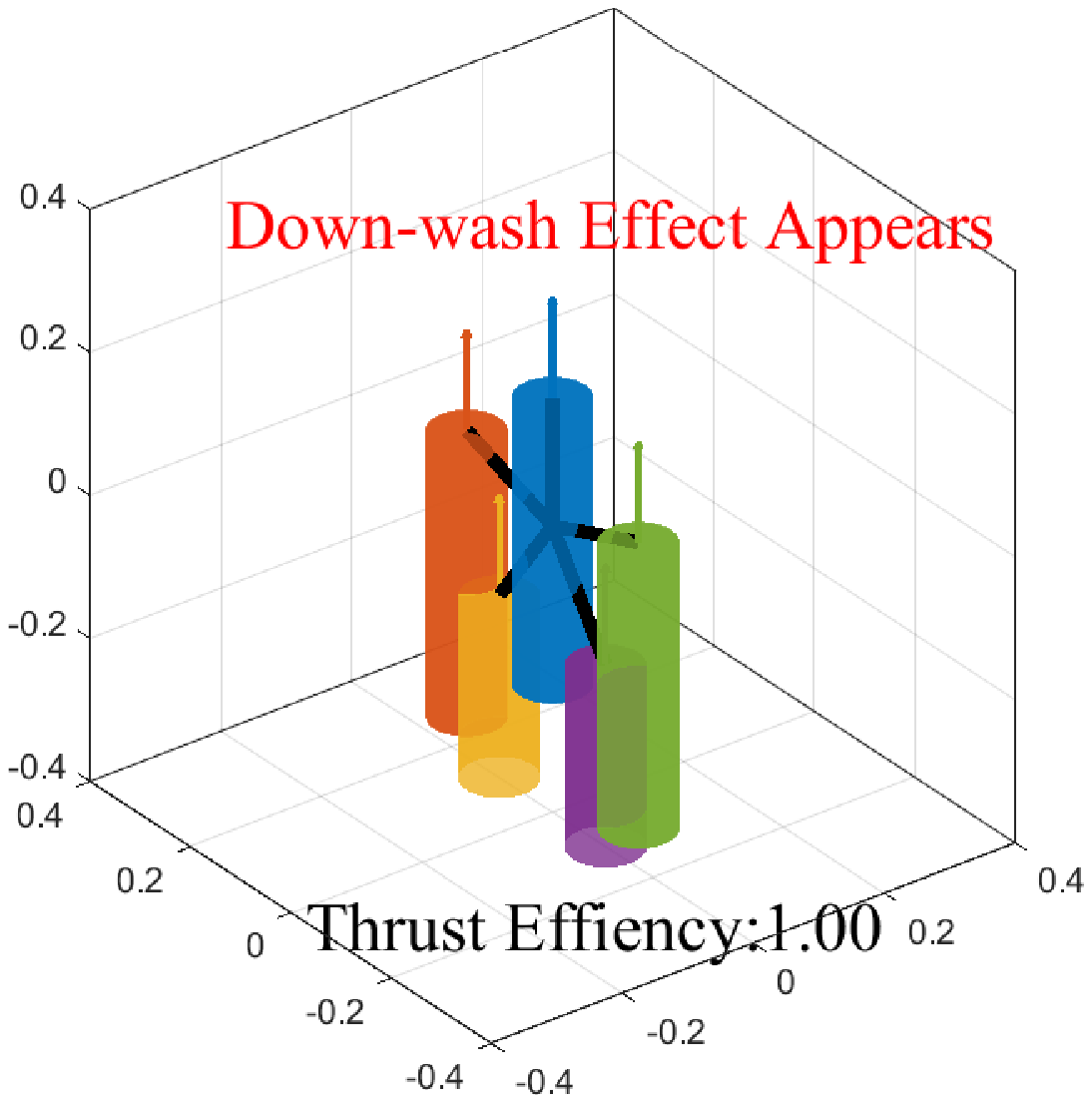}
        \caption{Five: Config. 1}
        \label{fig:penta_1}
    \end{subfigure}
    \begin{subfigure}[b]{0.32\linewidth}
        \centering
        \includegraphics[width=\linewidth,trim=5.1cm 9.2cm 5.1cm 9.2cm, clip]{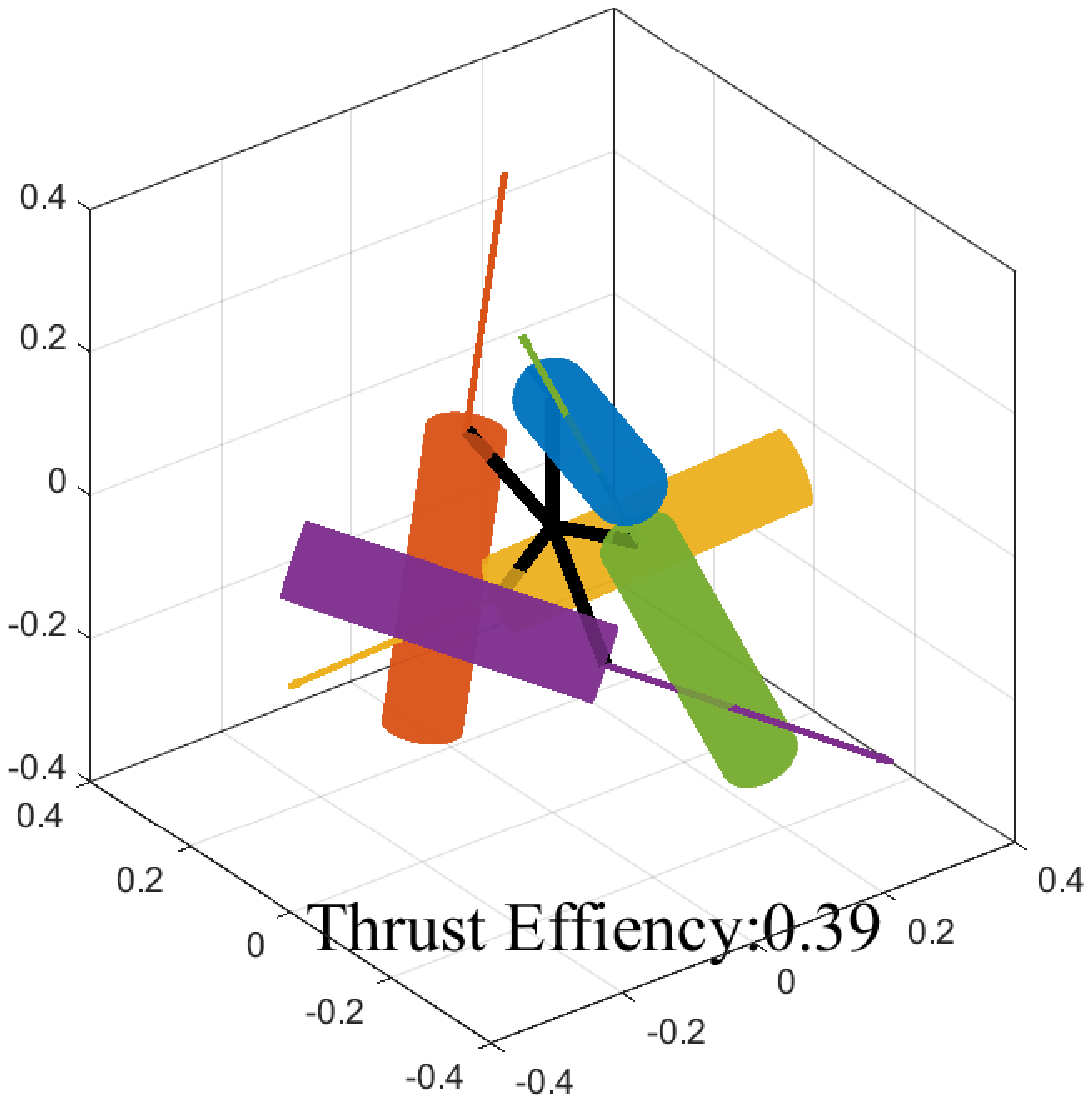}
        \caption{Five: Config. 2}
        \label{fig:penta_2}
    \end{subfigure}
    \begin{subfigure}[b]{0.32\linewidth}
        \centering
        \includegraphics[width=\linewidth,trim=5.1cm 9.2cm 5.1cm 9.2cm, clip]{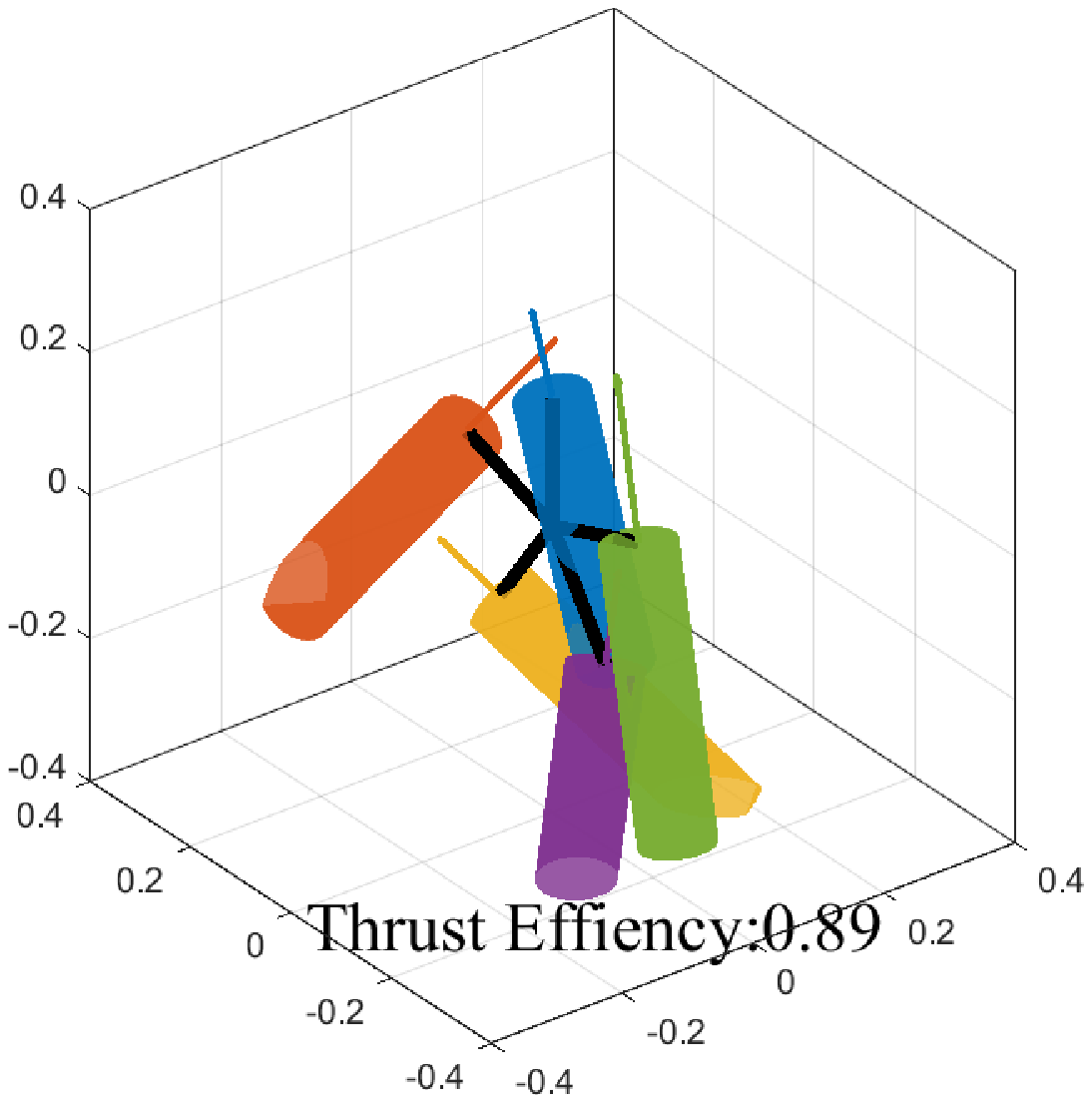}
        \caption{Five: Config. 3}
        \label{fig:penta_3}
    \end{subfigure}\\
    \begin{subfigure}[b]{0.32\linewidth}
        \centering
        \includegraphics[width=\linewidth,trim=5.1cm 9.2cm 5.1cm 9.2cm, clip]{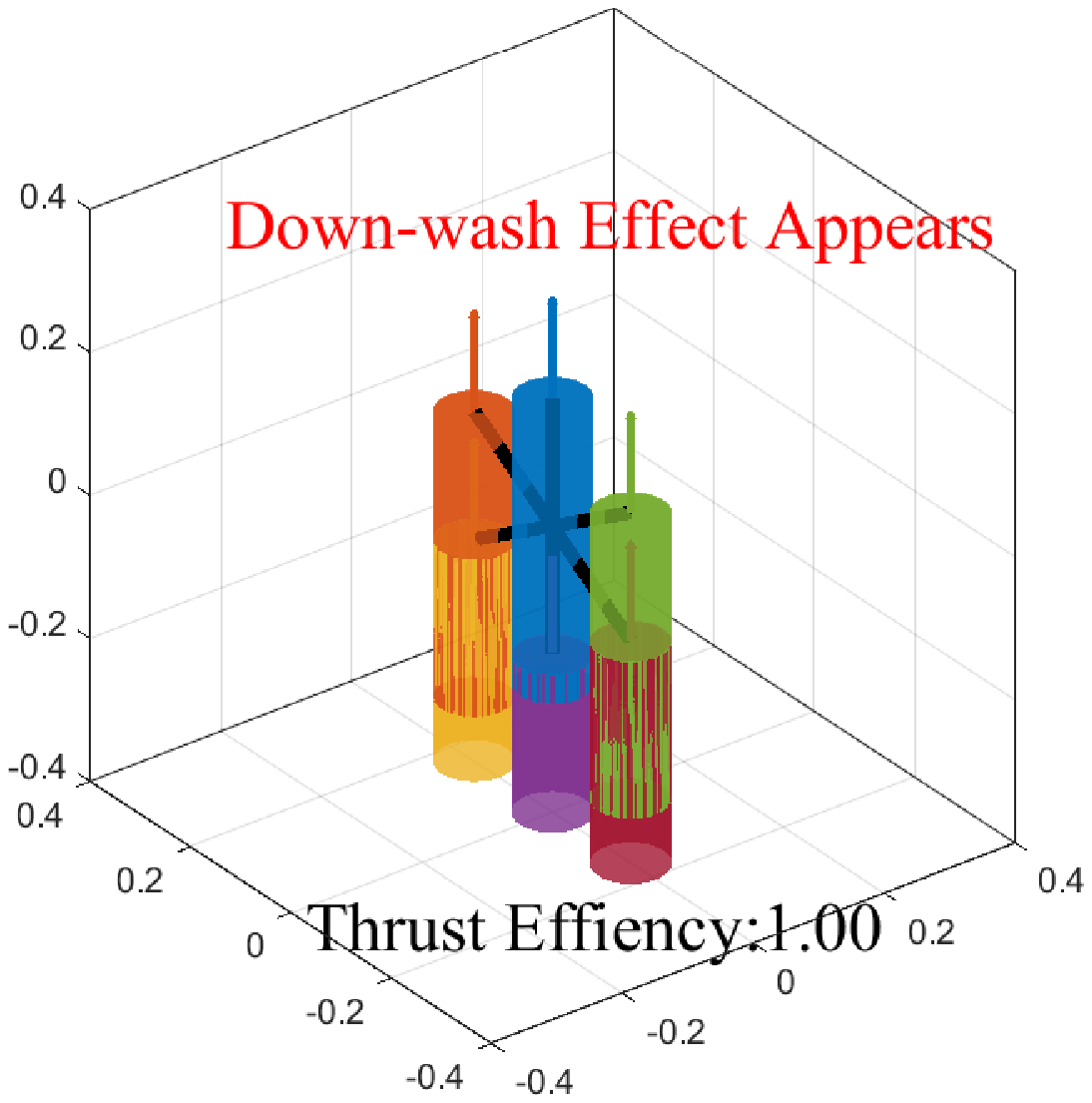}
        \caption{Six: Config. 1}
        \label{fig:hexa_1}
    \end{subfigure}
    \begin{subfigure}[b]{0.32\linewidth}
        \centering
        \includegraphics[width=\linewidth,trim=5.1cm 9.2cm 5.1cm 9.2cm, clip]{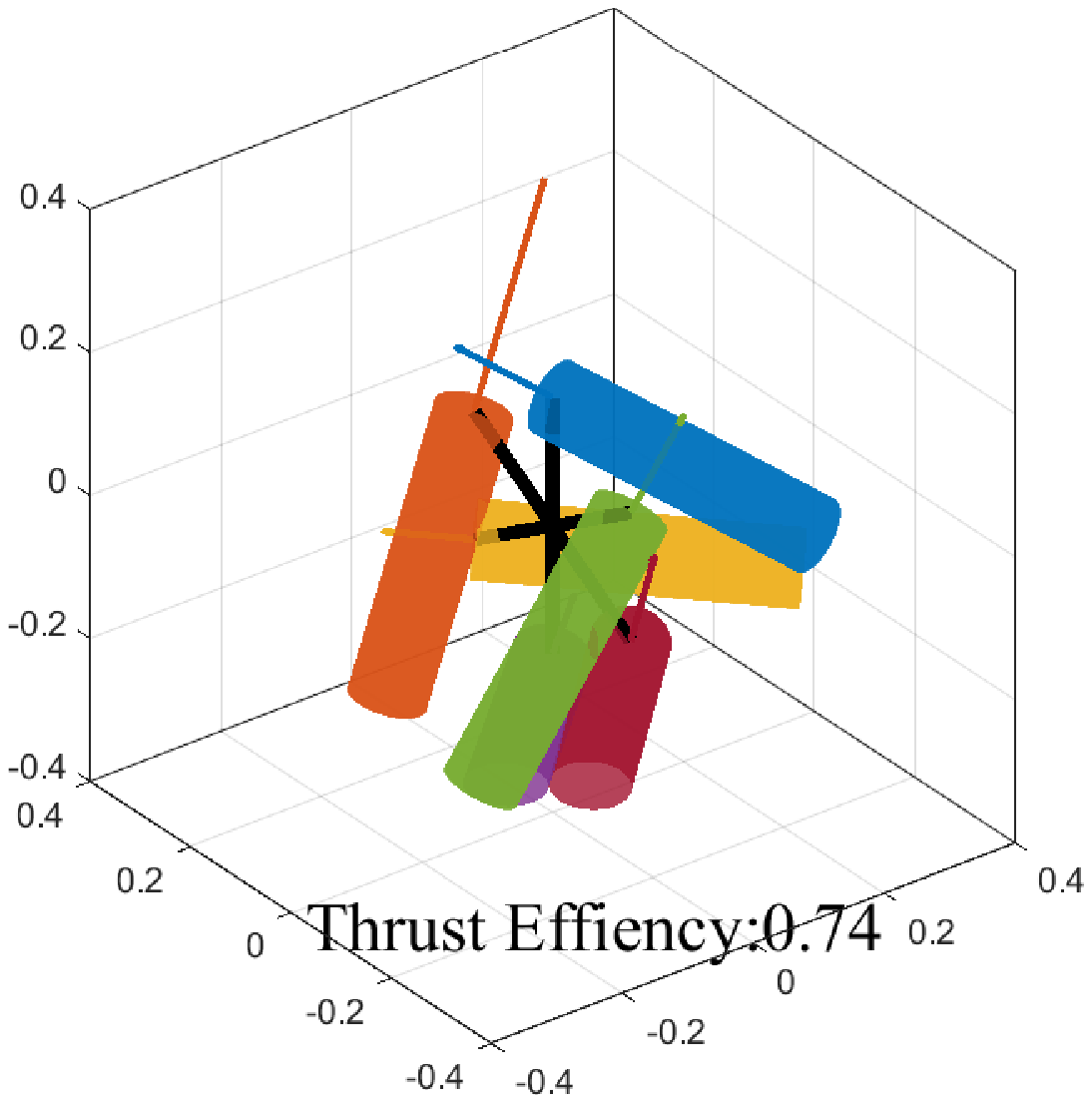}
        \caption{Six: Config. 2}
        \label{fig:hexa_2}
    \end{subfigure}
    \begin{subfigure}[b]{0.32\linewidth}
        \centering
        \includegraphics[width=\linewidth,trim=5.1cm 9.2cm 5.1cm 9.2cm, clip]{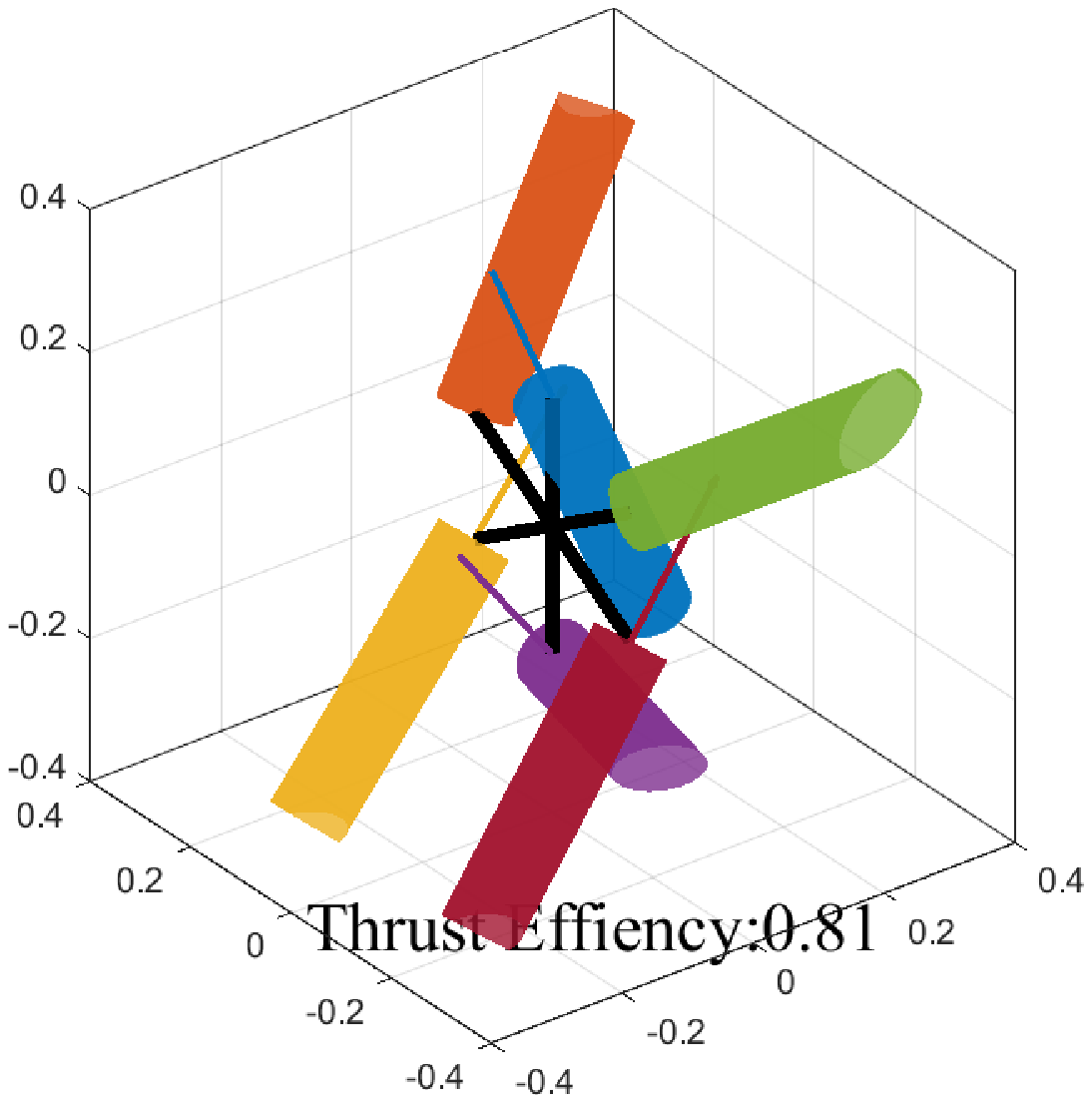}
        \caption{Six: Config. 3}
        \label{fig:hexa_3}
    \end{subfigure}
    \caption{\textbf{Thrust efficiency and downwash effect avoidance for different over-actuated \ac{uav} platforms.} Infinite number of thrust force configurations can generate the same required wrench command with different thrust efficiencies. For each platform, three configurations are provided as examples. Four, five, six refer to the platform with 4, 5, or 6 3-\ac{dof} thrust generators, respectively. Same notations are applied for the rest of this paper.}
    \label{fig:efficiency_and_downwash} 
\end{figure}     

\subsection{Downwash Effect Avoidance and Thrust Efficiency}\label{sec:efficiency}

The ``thrust efficiency index'' was defined by Ryll \etal~\cite{ryll2016modeling} to quantify wasted internal forces in over-actuated multirotor systems. Formally, it is defined as
\begin{equation} 
    \small
    \begin{aligned}
        \eta_f &= \frac{\| \sum_{i=1}^{N} \prescript{B}{i}{\pmb{R}}\, T_i \pmb{\hat{z}} \|}{\sum_{i=1}^{N} T_i}
        =\frac{\|\pmb{J}_\xi(\pmb{\alpha},\pmb{\beta})\pmb{T}\|}{\sum_{i=1}^{N}
        T_i}\\
        &=\frac{\|\pmb{u}(1:3,1)\|}{\sum_{i=1}^{N}
        T_i},
        \quad\eta_f\in\left[0,1\right] 
    \end{aligned}
    \label{eq:nf}
\end{equation}
where $\eta_f$ is a configuration-dependent ratio between the sum of vectored thrusts and the sum of total thrust magnitudes. 

We study the relation between downwash avoidance and thrust efficiency for three different over-actuated \ac{uav} platforms with four, five, and six 3-\ac{dof} thrust generators; \cref{fig:efficiency_and_downwash} summarizes the results. When the platforms fly vertically (see \cref{fig:quad_1,fig:penta_1,fig:hexa_1}), downwash effects still appear as most prior allocation frameworks~\cite{su2021nullspace,yu2021over} if we only try to maintain maximum thrust efficiency ($\eta_f=1$). By exploring the entire configuration space, other feasible configurations might both avoid downwash effects and maintain high thrust efficiency (see \cref{fig:quad_3,fig:penta_3,fig:hexa_3}). This finding motivates us to propose a new allocation framework that efficiently finds such a configuration for the over-actuated \ac{uav} platforms.

In \cref{eq:nf}, the numerator of $\eta_f$ is provided by wrench command $\pmb{u}$ of tracking controller, which can be treated as a constant value in allocation. To include thrust efficiency index into the objective function of the nullspace-based allocation framework, we choose to minimize the denominator of \cref{eq:nf} ($\sum_{i=1}^{N} T_i$); please refer to \cref{sec:allo} for details.

\setstretch{1}

\section{Downwash-aware Controller Design}\label{sec:control}

\begin{figure}[t!]
    \centering
    \begin{subfigure}[b]{\linewidth}
        \centering
        \includegraphics[width=\linewidth,trim=6cm 7.2cm 6cm 5cm,clip]{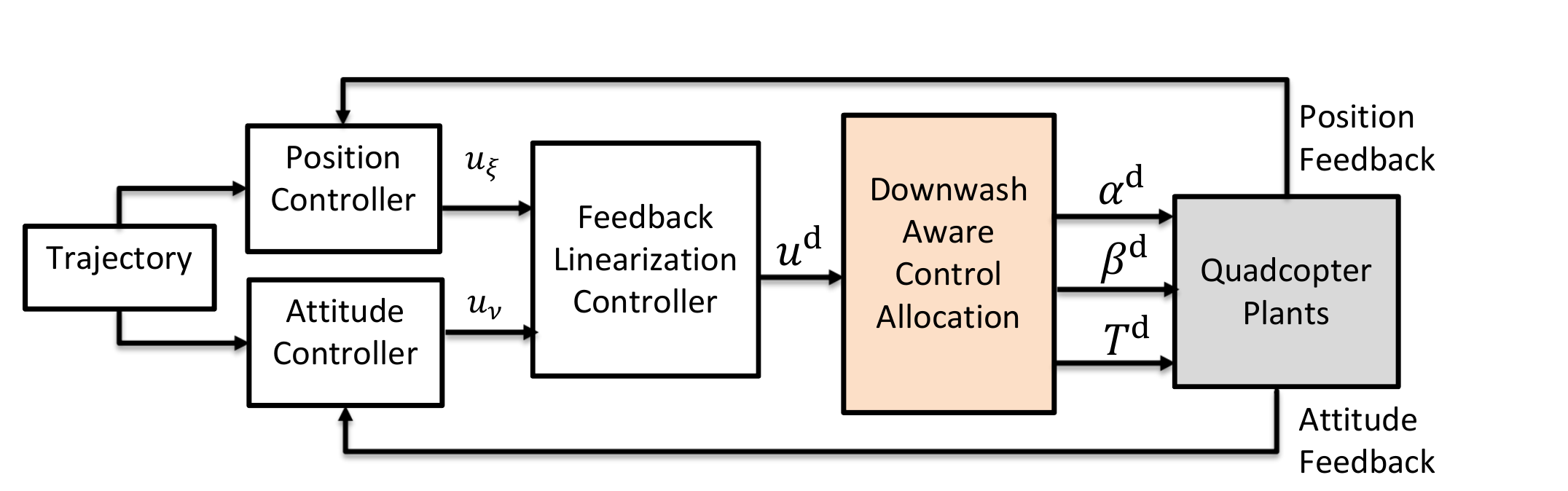}
        \caption{High-level platform trajectory tracking controller (100~$Hz$)}
        \label{fig:control_high}
    \end{subfigure}%
    \\
    \begin{subfigure}[b]{\linewidth}
        \centering
        \includegraphics[width=\linewidth,trim=5cm 7.5cm 4cm 6.8cm, clip]{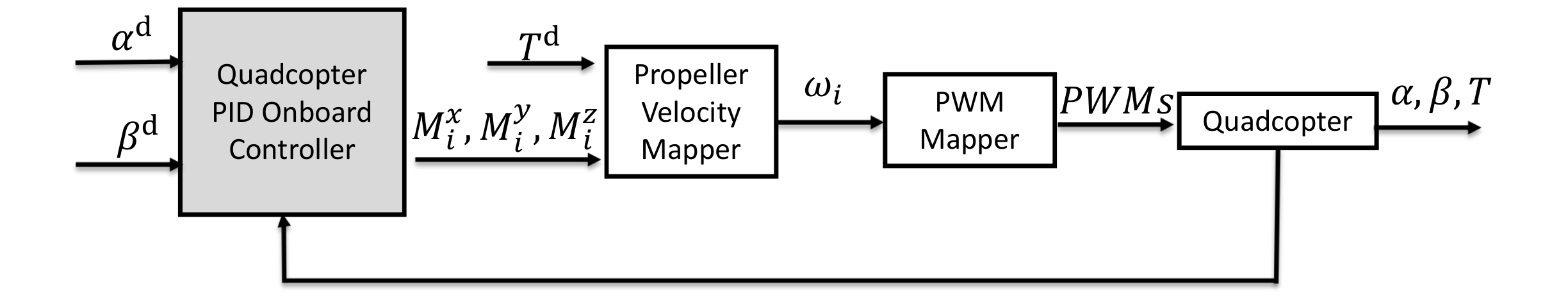}
        \caption{Low-level 3-\ac{dof} thrust generator controller (500~$Hz$)}
        \label{fig:control_low}
    \end{subfigure}
    \caption{\textbf{Hierarchical control architecture.} (a) The high-level position and attitude tracking controller gives desired 6-\ac{dof} wrench command $\pmb{u}^d$ to the downwash-aware control allocation through feedback linearization. $\pmb{u}^d$ is then allocated as the desired thrusts and joint angles for each 3-\ac{dof} thrust generator to maintain high thrust efficiency and avoid downwash effects. (b) In low-level control, each quadcopter module regulates its joint angles and thrust with an onboard PID controller. The angular velocity commands are converted to PWM signals for motor actuation.}
    \label{fig:controller}
\end{figure}

The overall controller has a hierarchical architecture, shown in \cref{fig:controller}. The high-level trajectory tracking controller (see \cref{fig:control_high}) (i) calculates the desired force/torque command (6-\ac{dof} wrench command) for the entire platform, and (ii) allocates the force/torque command to tilting angle $\alpha_i$, twisting angle $\beta_i$, and thrust $T_i$ of each 3-\ac{dof} thrust generator. The low-level controller (see \cref{fig:control_low}) of each quadcopter (i) regulates the individual attitude to the desired values and (ii) provides the required thrust force.

\subsection{High-level Control}

Without downwash effects, the dynamics equation (\ie, \cref{eq:dyna}) can be rewritten following Su \etal~\cite{su2021compensation}
\begin{equation}
    \small
    \begin{bmatrix}
        \prescript{W}{}{\pmb{\ddot{\xi}}} \\
        \prescript{B}{}{\pmb{\dot{\nu}}} \\
    \end{bmatrix} = 
    \begin{bmatrix}
       {\frac{1}{m}}\prescript{W}{B}{\pmb{R}} &0\\
       0 & \prescript{B}{}{\pmb{J}}^{-1}
    \end{bmatrix}\pmb{u}
    + 
    \begin{bmatrix}
        g \pmb{\hat{z}} \\
    \prescript{B}{}{\pmb{J}}^{-1}\prescript{B}{}{\pmb{\tau}_g} \\
    \end{bmatrix}.
    \label{eq:simple_dyna}
\end{equation}
We design the feedback-linearization controller as
\begin{equation}
    \small
    \pmb{u}^d=
    \begin{bmatrix}
        {m}\prescript{W}{B}{\pmb{R}}^\mathsf{T} & \pmb{0}
        \\
        \pmb{0} & \prescript{B}{}{\pmb{J}}
    \end{bmatrix}    
    \left(
        \begin{bmatrix}
            \pmb{u}_{\xi}
            \\
            \pmb{u}_{\nu}
        \end{bmatrix}-    
        \begin{bmatrix}
            g \pmb{\hat{z}} \\
            \prescript{B}{}{\pmb{\tau}_g} \\
        \end{bmatrix}
    \right),
\end{equation}
where the superscript $d$ indicates the desired values. Our above controller design transfers platform dynamics expressed by \cref{eq:simple_dyna} into a simple double integrator~\cite{ruan2022control},
\begin{equation}
    \small
    \begin{bmatrix}
        \prescript{W}{}{\pmb{\ddot{\xi}}}
        \\
        \prescript{B}{}{\pmb{\dot{\nu}}}
    \end{bmatrix} = 
    \begin{bmatrix}
        \pmb{u}_{\xi} 
        \\ 
        \pmb{u}_{\nu}
    \end{bmatrix}.
\end{equation}
Two virtual inputs $\pmb{u}_\xi$ and $\pmb{u}_\nu$ can be designed with translational and rotational errors to track predefined reference position and attitude trajectory. We close this control loop by an LQR controller that considers communication delay and improves system robustness~\cite{ruan2022control,su2022fault}.

\subsection{Downwash-aware Control Allocation}\label{sec:allo}

The nullspace-based control allocation framework of over-actuated \acp{uav} has been proposed in Su \etal~\cite{su2021nullspace} to solve $\pmb{\alpha}$, $\pmb{\beta}$, and $\pmb{T}$ from $\pmb{u}^d$ while maintaining defined input constraints. To avoid downwash effects and maintain high thrust efficiency, we modify the framework and reformulate the \ac{qp} problem as described below.

An intermediate variable $\pmb{F}$ is defined as
\begin{equation}
    \small
    \pmb{F}(\pmb{\alpha},\pmb{\beta},\pmb{T})=
    \begin{bmatrix}
        \pmb{F}_1^T & \cdots & \pmb{F}_N^T
    \end{bmatrix}
    ^T\in\mathbb{R}^{3N\times1},
\end{equation}
where
\begin{equation}
    \small
    \pmb{F}_{i}(\alpha_i,\beta_i,T_i)=T_{i} 
    \begin{bmatrix}
        \sin\beta_{i}\\
        -\sin\alpha_{i}\cos\beta_{i}\\
        \cos\alpha_{i}\cos\beta_{i} 
    \end{bmatrix}.
\end{equation}
With $\pmb{F}$, we can transform the nonlinear allocation problem to a linear one, 
\begin{equation}
    \small
    \pmb{u}^d = 
    \begin{bmatrix}
        \pmb{J}_\xi(\pmb{\alpha},\pmb{\beta}) \\ \pmb{J}_\nu(\pmb{\alpha},\pmb{\beta})
    \end{bmatrix}
    \pmb{T}
    = 
    \pmb{WF},
    \label{eq:static}
\end{equation}
where $\pmb{W}\in\mathbb{R}^{6\times3N}$ is a constant allocation matrix with full row rank. Therefore, $\pmb{F}$ can be solved from $\pmb{u}^d$ with a general solution form,
\begin{equation}
    \small
    \pmb{F}(\pmb{\alpha},\pmb{\beta},\pmb{T})=\pmb{W}^\dagger\pmb{u}^d+\pmb{N}_W\pmb{Z},
    \label{eq:general_solution}
\end{equation}
where $\pmb{N}_W\in\mathbb{R}^{3N\times(3N-6)}$ is the nullspace of $\pmb{W}$, and $\pmb{Z}\in \mathbb{R}^{(3N-6)\times1}$ is an arbitrary vector.

As discussed in Su \etal~\cite{su2021nullspace}, \cref{eq:general_solution} is linearized with the first-order Taylor expansion and relaxed with slack variable $\pmb{s}\in\mathbb{R}^{3N\times1}$,
\begin{equation}
    \small
    \pmb{s}+\pmb{F}(\pmb{X}_0)+\left.\frac{\partial{\pmb{F}}}{\partial\pmb{X}}\right\vert_{\pmb{X}=\pmb{X}_0}
    \Delta\pmb{X}=\pmb{W}^{\dagger}\pmb{u}^d+\pmb{N}_W\pmb{Z},
    \label{eq:constraint}
\end{equation}
where $\pmb{X}$ is defined as $\pmb{X}=[\pmb{\alpha}^\mathsf{T}, \, \pmb{\beta}^\mathsf{T}, \, \pmb{T}^\mathsf{T}]^\mathsf{T}$, $[\cdot]_0$ is the value of a variable at last time step, and 
$\Delta[\cdot]$ is the difference \wrt the previous time step of a variable.

Similarly, the downwash avoidance constraint (see \cref{eq:dist_constraint}) can be approximated by another linear equation as a linear inequality constraint,
\begin{equation}
    \small
    \pmb{O}(\pmb{X}_0)+\left.\frac{\partial{\pmb{O}}}{\partial\pmb{X}}\right\vert_{\pmb{X}=\pmb{X}_0}
    \Delta\pmb{X}\geq\pmb{O}_{\text{min}}.
    \label{eq:down_wash_constraint}
\end{equation}

The physical constraints of the platform are designed as  
\begin{equation}
    \small
    \pmb{X}_{\text{min}}-\pmb{X}_o\leq {\Delta}\pmb{X} \leq \pmb{X}_{\text{max}}-\pmb{X}_o,
    \label{eq:deltax1}  
\end{equation}
\begin{equation}
    \small
    {\Delta}\pmb{X}_{\text{min}}\leq {\Delta}\pmb{X} \leq {\Delta}\pmb{X}_{\text{max}}.
    \label{eq:deltax2}  
\end{equation}

The objective function is designed as
\begin{equation}
    \small
    {\Delta}\pmb{X}^\mathsf{T}\pmb{Q}_1{\Delta}\pmb{X}+\pmb{s}^\mathsf{T}\pmb{Q}_2\pmb{s}+\pmb{Z}^\mathsf{T}\pmb{Q}_3\pmb{Z}+\pmb{P}^\mathsf{T}{\Delta}\pmb{X},
    \label{eq:objective_function}
\end{equation}
where $\pmb{Q}_{1-3}$ are three positive semi-definite weighting matrices. As introduced in \cref{sec:efficiency}, the thrust efficiency index is included as $\pmb{P}^\mathsf{T}(\pmb{X}_o+{\Delta}\pmb{X})$, with
\begin{equation}
    \small
    \pmb{P}^\mathsf{T}=
    \begin{bmatrix}
        \pmb{0}_{1{\times}2N} & \gamma\pmb{1}_{1{\times}N}
    \end{bmatrix}
    \in\mathbb{R}^{1\times3N}.
\end{equation}
Then we have
\begin{equation}
    \small
    \pmb{P}^\mathsf{T}(\pmb{X}_o+{\Delta}\pmb{X})=\gamma \sum_{i=1}^{N} T_i,
\end{equation}
where $\gamma$ is the scaling factor. Of note, $\pmb{P}^\mathsf{T}\pmb{X}_o$ is a constant, thus removed from the objective function.

After solving this optimization problem \cref{eq:constraint,eq:down_wash_constraint,eq:deltax1,eq:deltax2,eq:objective_function}, we can approximately calculate the desired $\pmb{X}$ for next step with discrete integration,   
\begin{equation}
    \small
    \pmb{X}=\pmb{X}_o+{\Delta}\pmb{X}.
    \label{eq:solve_alpha_T}
\end{equation}
To eliminate the approximation errors, we utilize nullspace projection with
\begin{align}
    \small
    \pmb{Z}^*& = \pmb{N}_W^\dagger(\pmb{F}(\pmb{X})-\pmb{W}^\dagger{\pmb{u}^d}),\\
    \pmb{F}^*& = \pmb{W}^\dagger{\pmb{u}^d}+\pmb{N}_W\pmb{Z}^*.
    \label{eq:proj}
\end{align}
Finally, with exact solution $\pmb{F}^*$, low-level commands $\pmb{\alpha}^d$, $\pmb{\beta}^d$, and $\pmb{T}^d$ can be recovered with inverse kinematics: 
\begin{align}
    \small
    T_{i}^d&=\sqrt{F_{ix}^2+F_{iy}^2+F_{iz}^2},\\
    \alpha_i^d&=\text{atan2}(-F_{iy},F_{iz}),\\
    \beta_i&=\text{asin}(\frac{F_{ix}}{T_{i}}).
    \label{eq:determine_beta}
\end{align}

\subsection{Low-level Control}

The joint angles of each quadcopter module are controlled by separate PID controllers based on the error dynamics:
\begin{equation}
    \small
    \begin{gathered}
        \Ddot{\alpha}_{i}^d = k_{P\alpha}e_{\alpha} + k_{I\alpha}\int{e_{\alpha}}dt+ k_{D\alpha}\dot{e}_{\alpha},  \\
        \Ddot{\beta}_{i}^d =  k_{P\beta}e_{\beta} + k_{I\beta}\int{e_{\beta}}dt+ k_{D\beta}\dot{e}_{\beta},
    \end{gathered}
\end{equation}
where $k_{[\cdot]\alpha}$ and $k_{[\cdot]\beta}$ are constant PID gains, and 
\begin{equation}
    \small
    \begin{gathered}  
    e_{\alpha}=\alpha_{i}^d-\alpha_{i}^e,\\ e_{\beta}=\beta_{i}^d-\beta_{i}^e,
    \end{gathered}
\end{equation}
are error terms with joint angle feedback $\alpha_{i}^e$, $\beta_{i}^e$ from onboard IMU. The related torque commands are determined by 
\begin{equation}
    \small
    \begin{split}
        M_{i}^x &= \prescript{B}{}{J_{i}^x} \Ddot{\alpha}_{i}^d \cos \beta_i,
        \\
        M_{i}^y &= \prescript{B}{}{J_{i}^y} \Ddot{\beta}_{i}^d,
        \\
        M_{i}^z &= \prescript{B}{}{J_{i}^x} \Ddot{\alpha}_{i}^d \sin \beta_i.
    \end{split}
    \label{eq:joint_torque}
\end{equation}
For each quadcopter module, with \cref{eq:quad_input_decouple,eq:omega}, the angular velocity $\omega_{i,j}$ of each propeller can be calculated, later converted to the PWM signal to drive the motor.

\section{Simulation and Experiment Setups}\label{sec:setup}

\subsection{Simulation Setup}

Before conducting physical experiments, we develop a simulation platform in Matlab Simulink/Simscape to evaluate and characterize the proposed downwash-aware control allocation framework. In addition to the \ac{uav}'s physical parameters obtained from system identification, the dynamics of propeller motors and saturation, control frequencies, measurement noise, and communication noise and delays, the simulator also incorporates the downwash aerodynamics model introduced in \cref{sec:downwashmodel} based on experimental data.

The proposed allocation framework was verified on two over-actuated platforms with four and six 3-\ac{dof} thrust generators, respectively. \cref{tab:setup} summarizes the physical and software properties acquired from the physical system used in simulation, where $m_0$ and $I_0$ refer to the mass and inertia matrix of the mainframe, and $m_i$ and $I_i$ refer to the mass and inertia matrix of each 3-\ac{dof} thrust generator.

\begin{table}[ht!]
    \small
    \centering
    \caption{\textbf{Physical and Software Properties in Simulation}}
    \resizebox{\linewidth}{!}{
        \begin{tabular}{lcc}
            \toprule
            \textbf{Parameter} & \textbf{Four}  & \textbf{Six}\\
            \midrule
            $m_0/kg$ & $0.020$ & $0.030$\\
            $m_i/kg$ & $0.050$ & $0.036$\\
            $\text{diag}(I_0)/kg \cdot cm^2$ & $[3.20\ 3.20 \ 4.70]$ & $[4.50\ 4.50 \ 6.20]$\\
            $\text{diag}(I_i)/kg \cdot cm^2$ & $[0.35\ 0.35\ 0.55]$ & $[0.16\ 0.16\ 0.29]$ \\
            $l/m$ & $0.21$ & $0.18$\\
            $a/m$ & $0.068$ & $0.032$\\
            $t_{\textit{max}}/N$ & $0.30$ & $0.15$ \\
            Communication delay/sec & $0.02$ & $0.02$\\
            Remote PC control rate/$Hz$ & $100$ & $100$\\
            Onboard control rate/$Hz$ & $500$ & $500$\\
            \bottomrule
        \end{tabular}%
    }
    \label{tab:setup}
\end{table}

\subsection{Experiment Setup}

\begin{figure}[t!]
    \centering 
    \begin{subfigure}{0.448\linewidth}
        \centering
        \includegraphics[width=\linewidth,trim=5cm 1.7cm 5cm 1cm, clip]{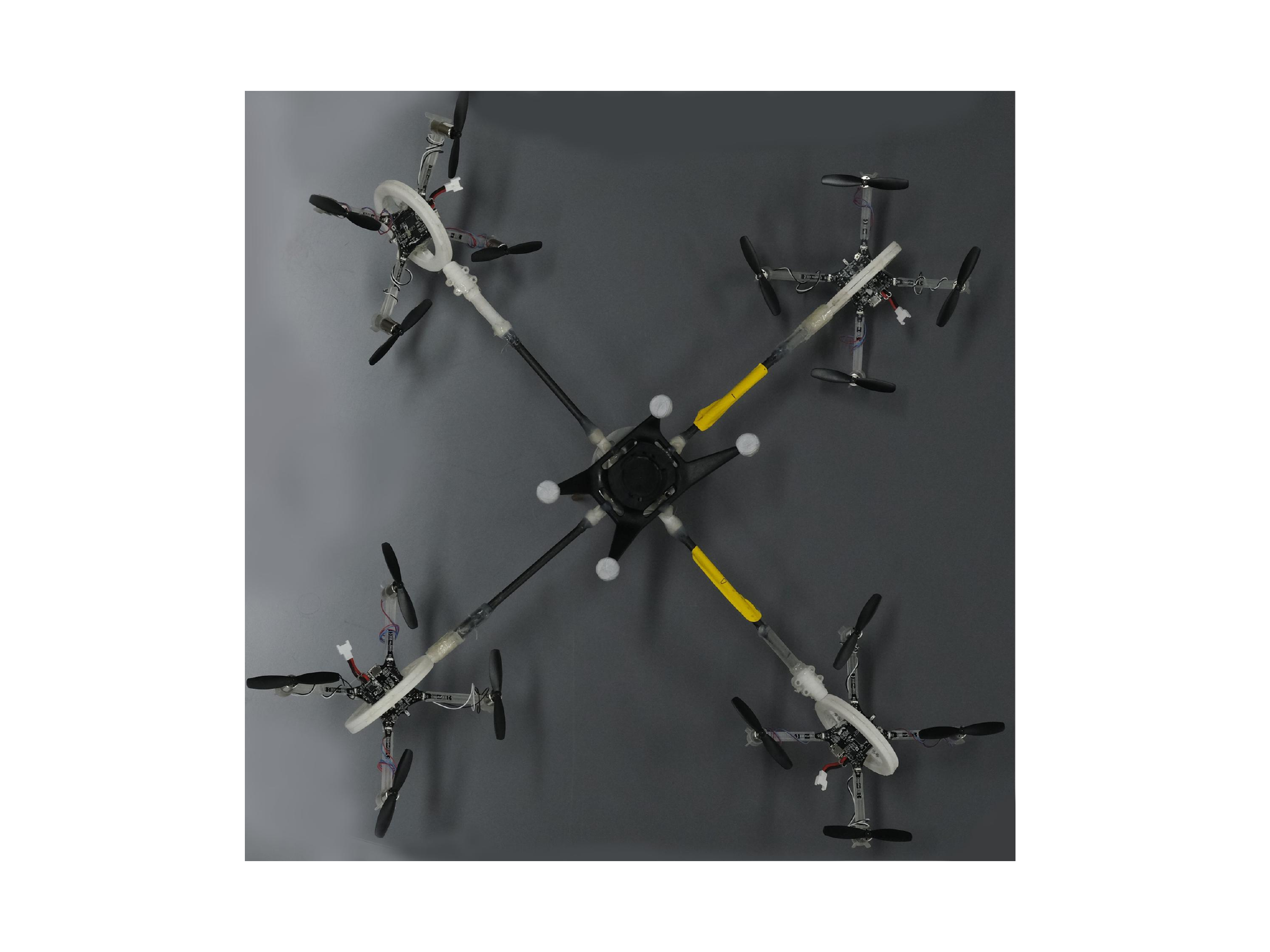}
        \caption{Four}
        \label{fig:four_hardware}
    \end{subfigure}%
    \hfill
    \begin{subfigure}{0.52\linewidth}
        \centering
        \includegraphics[width=\linewidth,trim=8cm 1.8cm 8cm 1cm, clip]{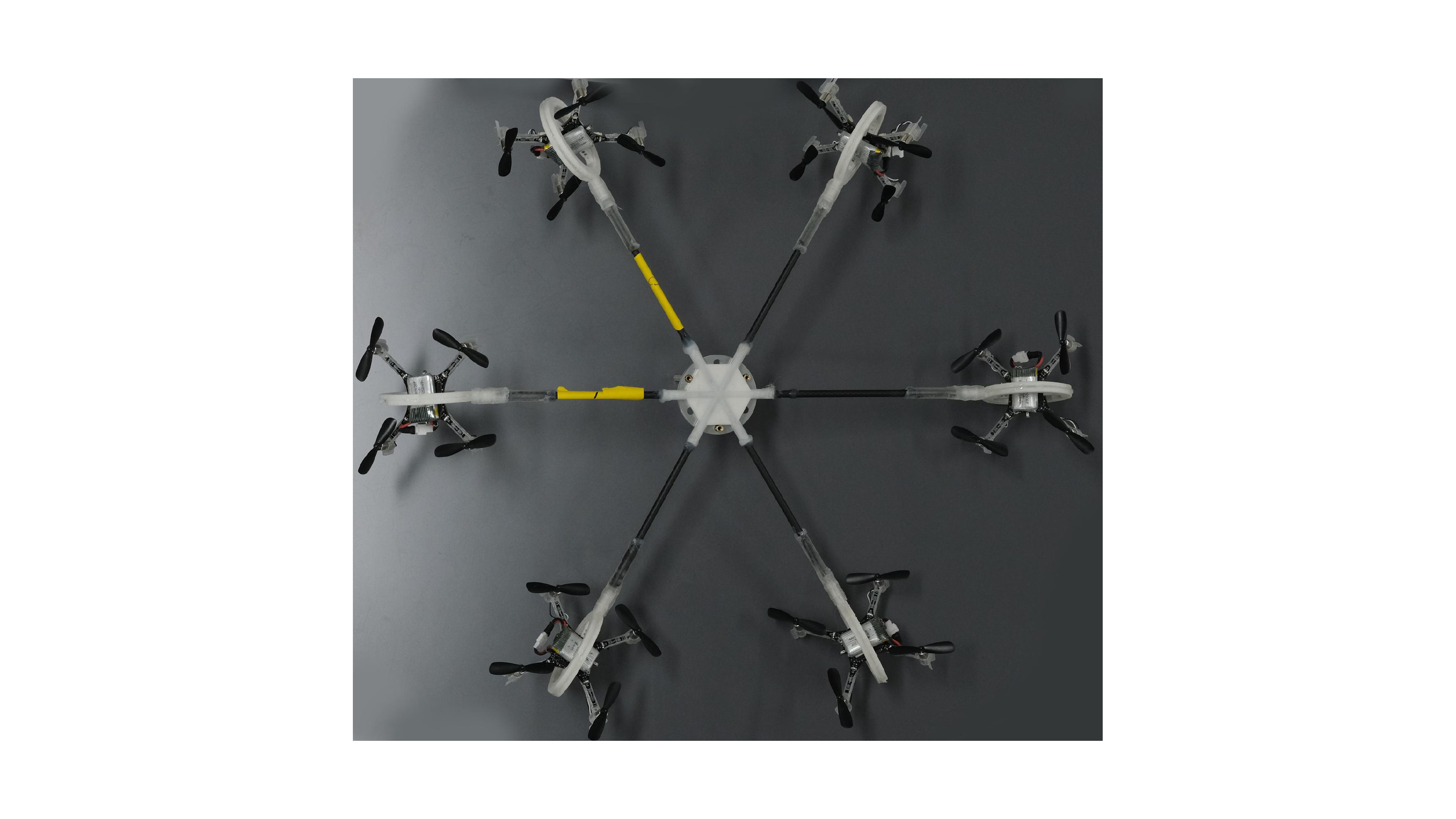}
        \caption{Six}
        \label{fig:fix_hardware}
    \end{subfigure}%
    \caption{\textbf{Hardware prototypes of the over-actuated \ac{uav} platforms.} The central frame is a rigid body made by carbon-fiber tubes and 3-D printed parts. Commercial quadcopter Crazyflie 2.1 from Bitcraze is combined with 3-D printed 2-\ac{dof} passive gimbal mechanism as the 3-\ac{dof} thrust generator. The platforms have (a) four and (b) six thrust generators, respectively.}
    \label{fig:prototype}
\end{figure} 

As shown in \cref{fig:prototype}, the quadcopters are connected to the central frame by 2-\ac{dof} passive gimbal mechanism, which have no rotation-angle limitations, thus can be utilized as 3-\ac{dof} thrust generators. We use Crazyflie 2.1 as the quadcopter module. The weight of Crazyflie 2.1 is $27g$ with a maximum $60g$ total payload. For the platform with four 3-\ac{dof} thrust generators, we upgraded the motors, propellers, and batteries of the Crazyflie for larger thrust force.
 
\begin{figure}[t!]
    \centering
    \includegraphics[width=0.9\linewidth]{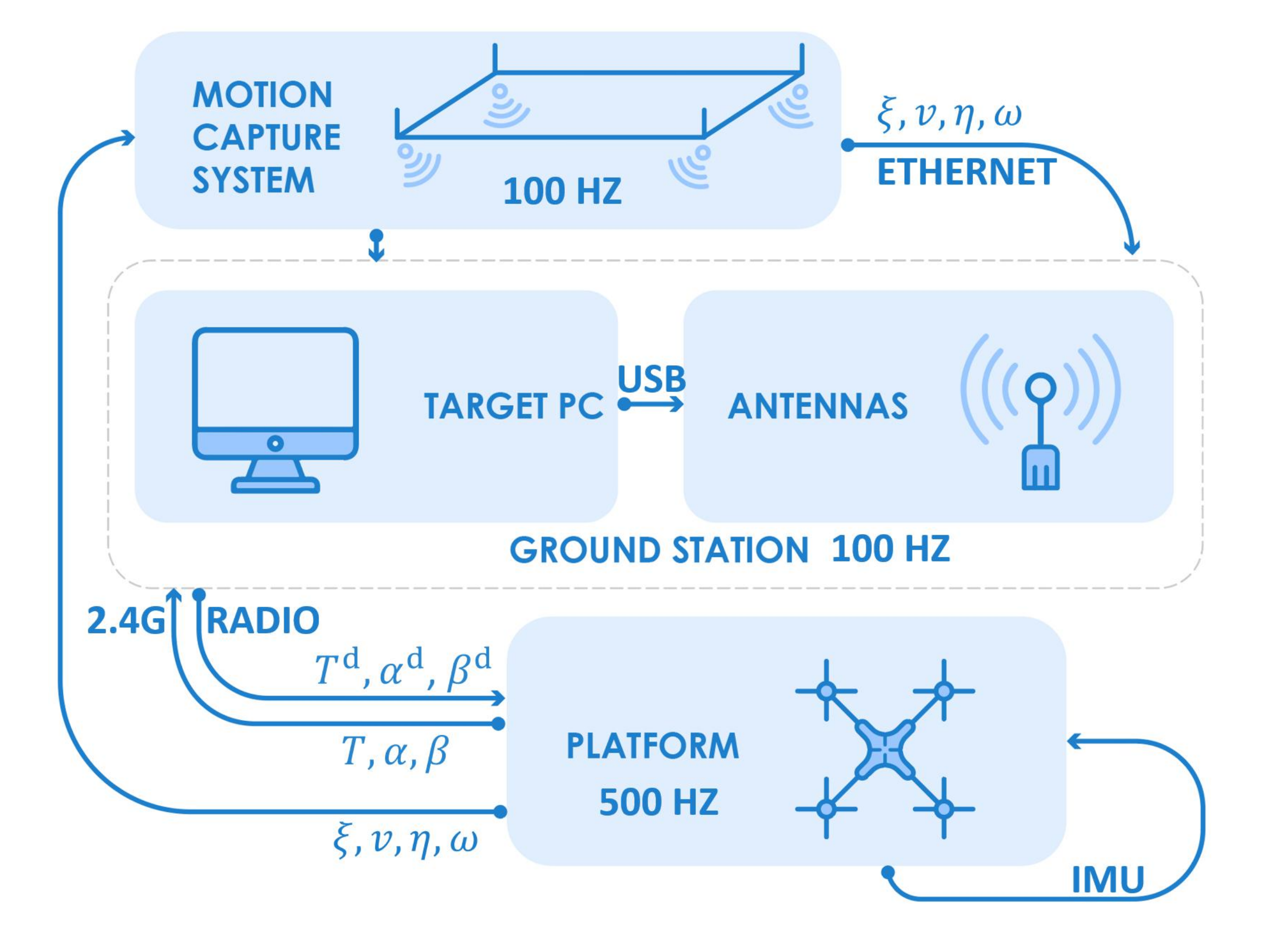}
    \caption{\textbf{Platform communication setup in experiment}. The remote PC takes position and attitude feedback from motion capture system, runs the high-level controller, and sends commands to each quadcopter through radio communication.}
    \label{fig:communication}
\end{figure}

In the experiment, we use the Noitom motion capture system to measure the position and attitude of the central frame. The main controller runs on a remote PC, which communicates with the motion capture system through Ethernet. The main controller calculates the desired thrust $\pmb{T}^d$, tilting angles $\pmb{\alpha}^d$, and twisting angles $\pmb{\beta}^d$ for all quadcopter modules. The communication between the remote PC and each quadcopter is achieved by Crazy Radio PA antennas (2.4G~$Hz$). Each quadcopter is embedded with an onboard IMU module, estimating the rotation angle given the attitude of central frame $\pmb{\eta}$. Meanwhile, the onboard controller regulates the tilting and twisting angles to desired values and provides the required thrust. The measurement rate of the motion capture system, the remote PC controller, and the data communication with each quadcopter are all set to 100~$Hz$. The quadcopter's onboard controller is set to 500~$Hz$ for fast low-level response. \cref{fig:communication} shows the software architecture.

\section{Simulation and Experiment Results}\label{sec:result}

\subsection{Simulation Results}

\begin{figure*}[t!]
    \centering
    \begin{subfigure}{0.2\linewidth}
        \centering
        \includegraphics[width=\linewidth,trim=3.2cm 10.2cm 3cm 9.4cm, clip]{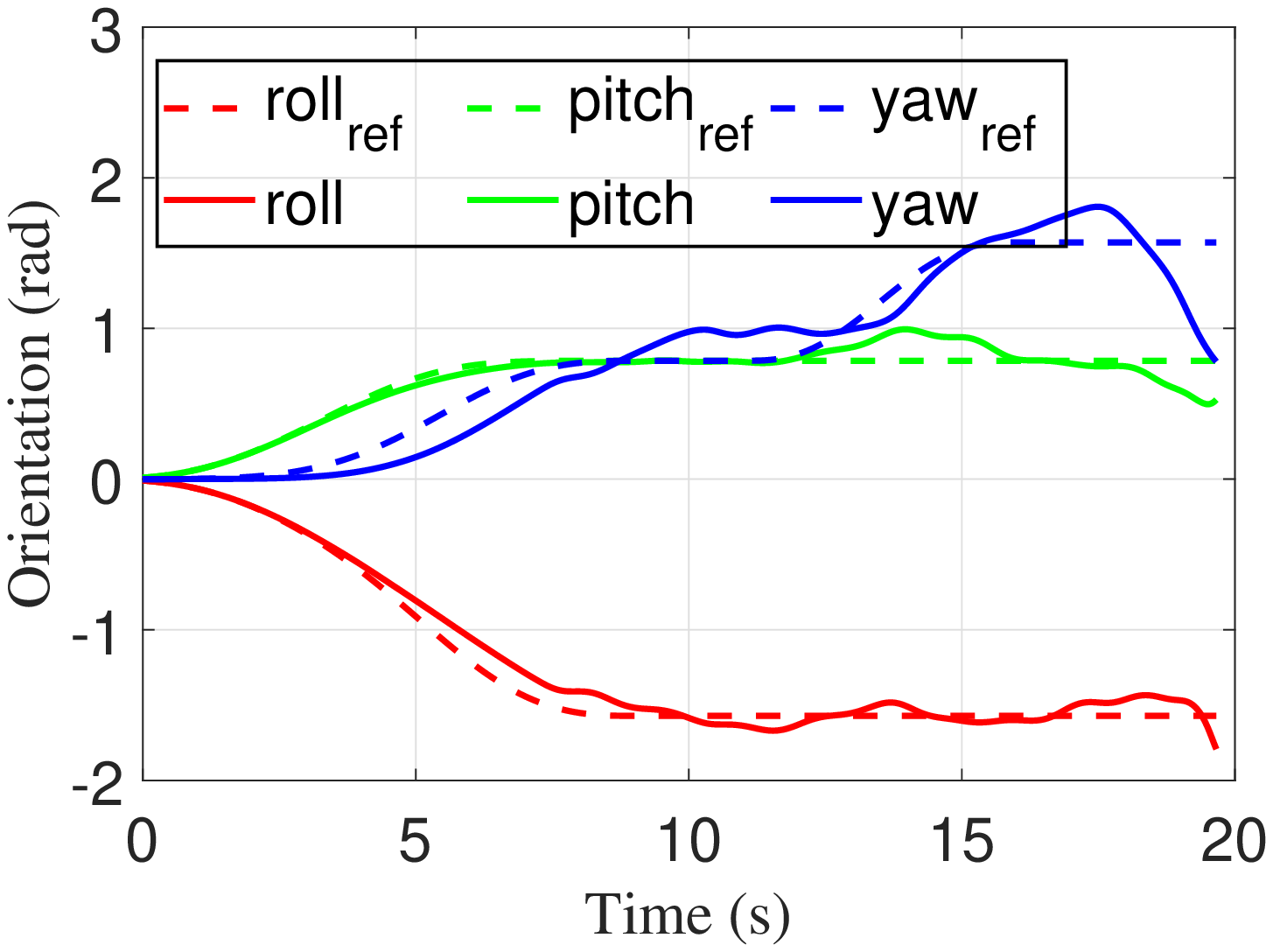}
        \caption{Four (C): Attitude.}
        \label{fig:four_sim_fail_rpy}
    \end{subfigure}%
    \begin{subfigure}{0.2\linewidth}
        \centering
        \includegraphics[width=\linewidth,trim=3.2cm 10.2cm 3cm 9.4cm, clip]{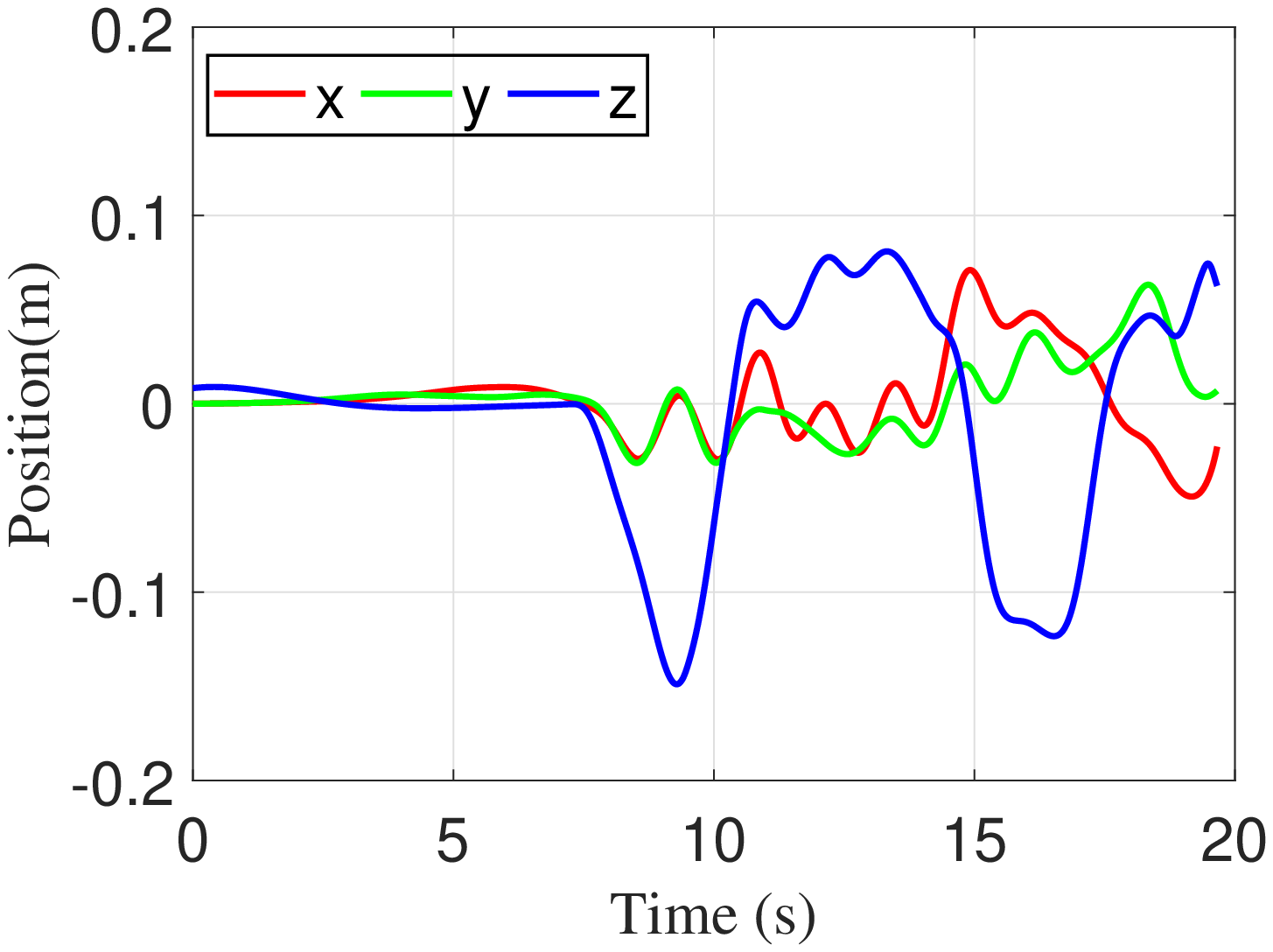}
        \caption{Four (C): Position.}
        \label{fig:four_sim_fail_xyz}
    \end{subfigure}%
    \begin{subfigure}{0.2\linewidth}
        \centering
        \includegraphics[width=\linewidth,trim=3.2cm 10.2cm 3cm 9.4cm, clip]{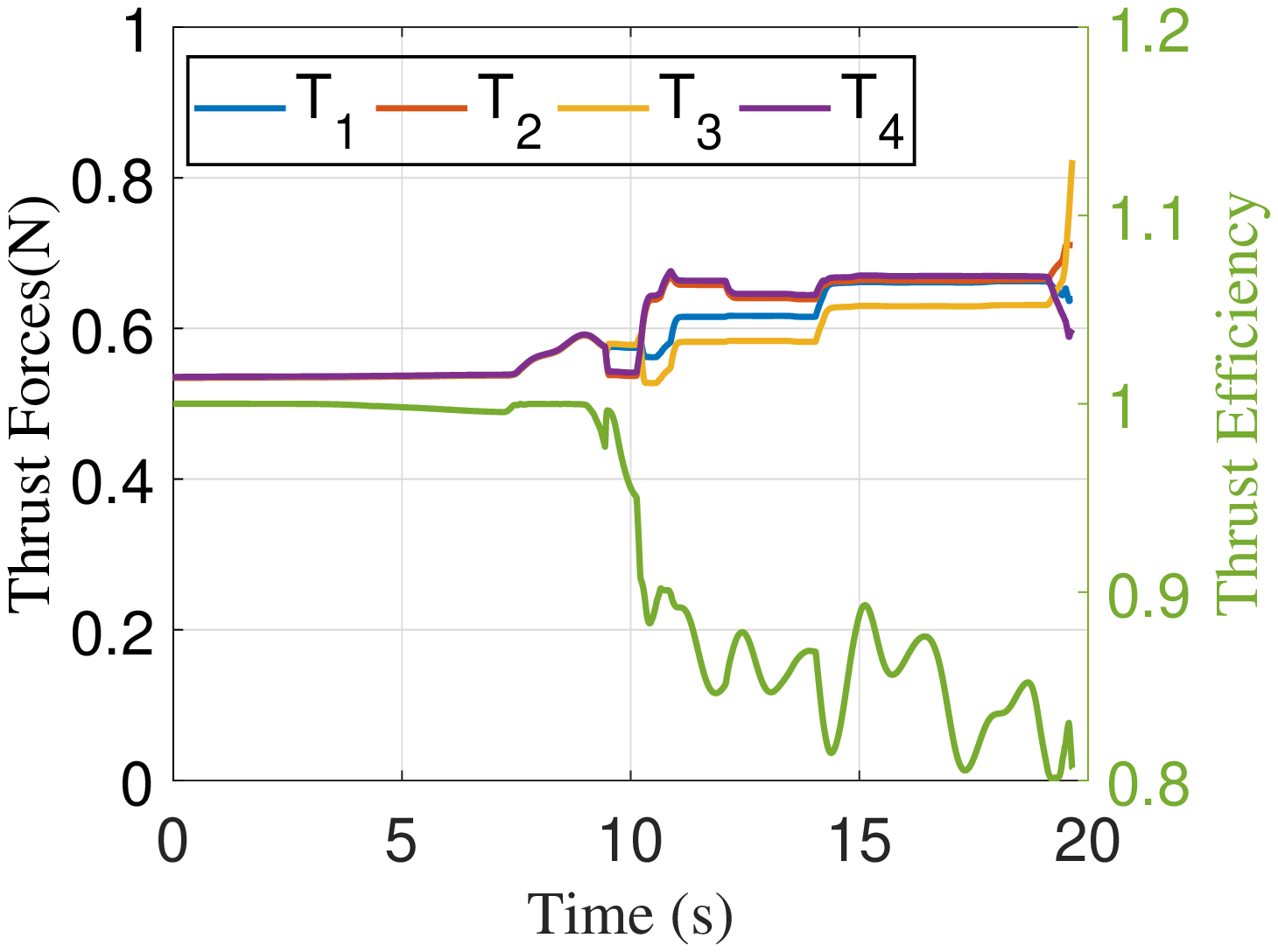}
        \caption{Four (C): Thrust.}
        \label{fig:four_sim_fail_thrust}
    \end{subfigure}%
    \begin{subfigure}{0.2\linewidth}
        \centering
        \includegraphics[width=\linewidth,trim=3.2cm 10.2cm 3cm 9.4cm, clip]{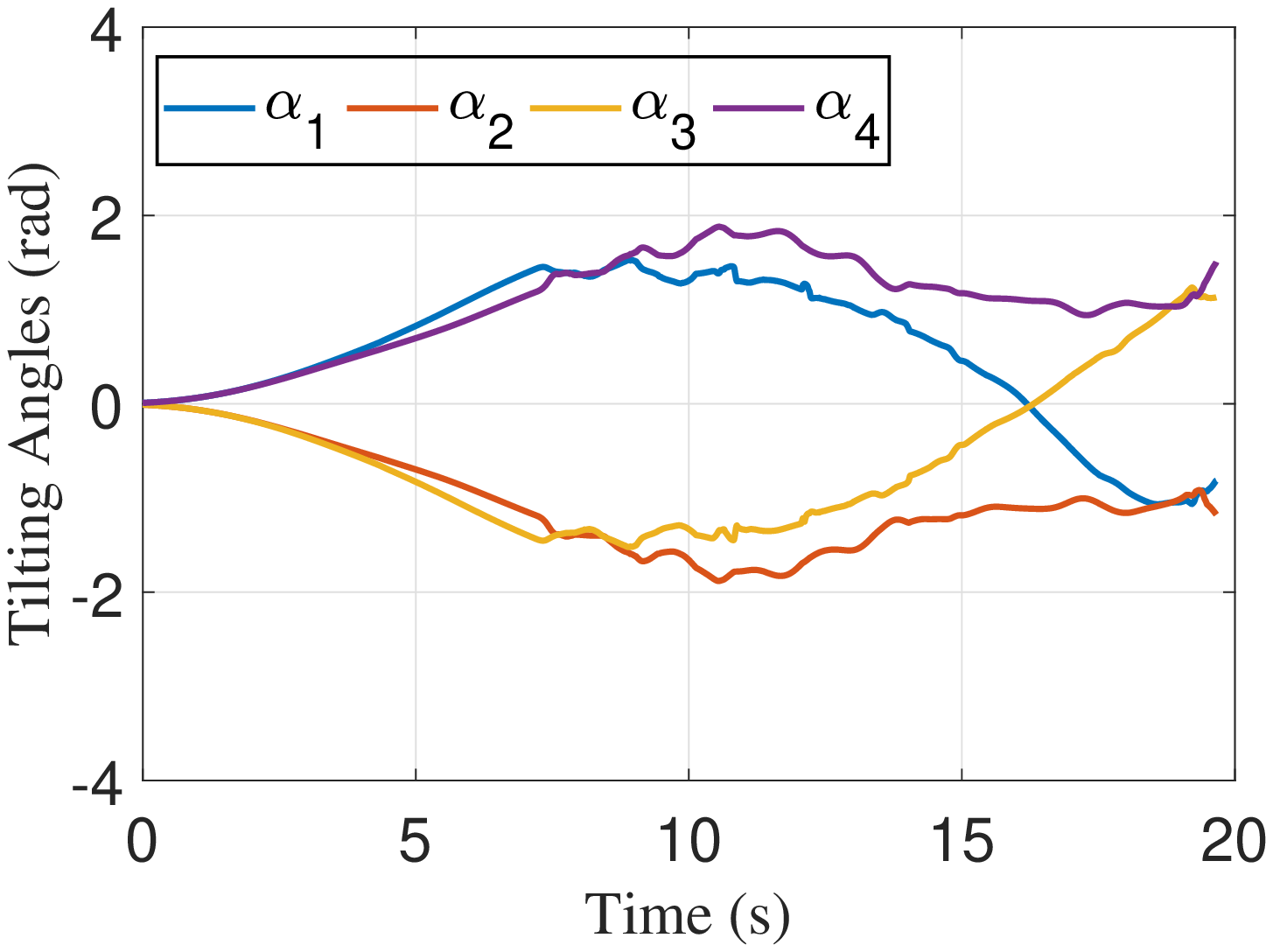}
        \caption{Four (C): Tilting Angles.}
        \label{fig:four_sim_fail_alpha}
    \end{subfigure}%
    \begin{subfigure}{0.2\linewidth}
        \centering
        \includegraphics[width=\linewidth,trim=3.2cm 10.2cm 3cm 9.4cm, clip]{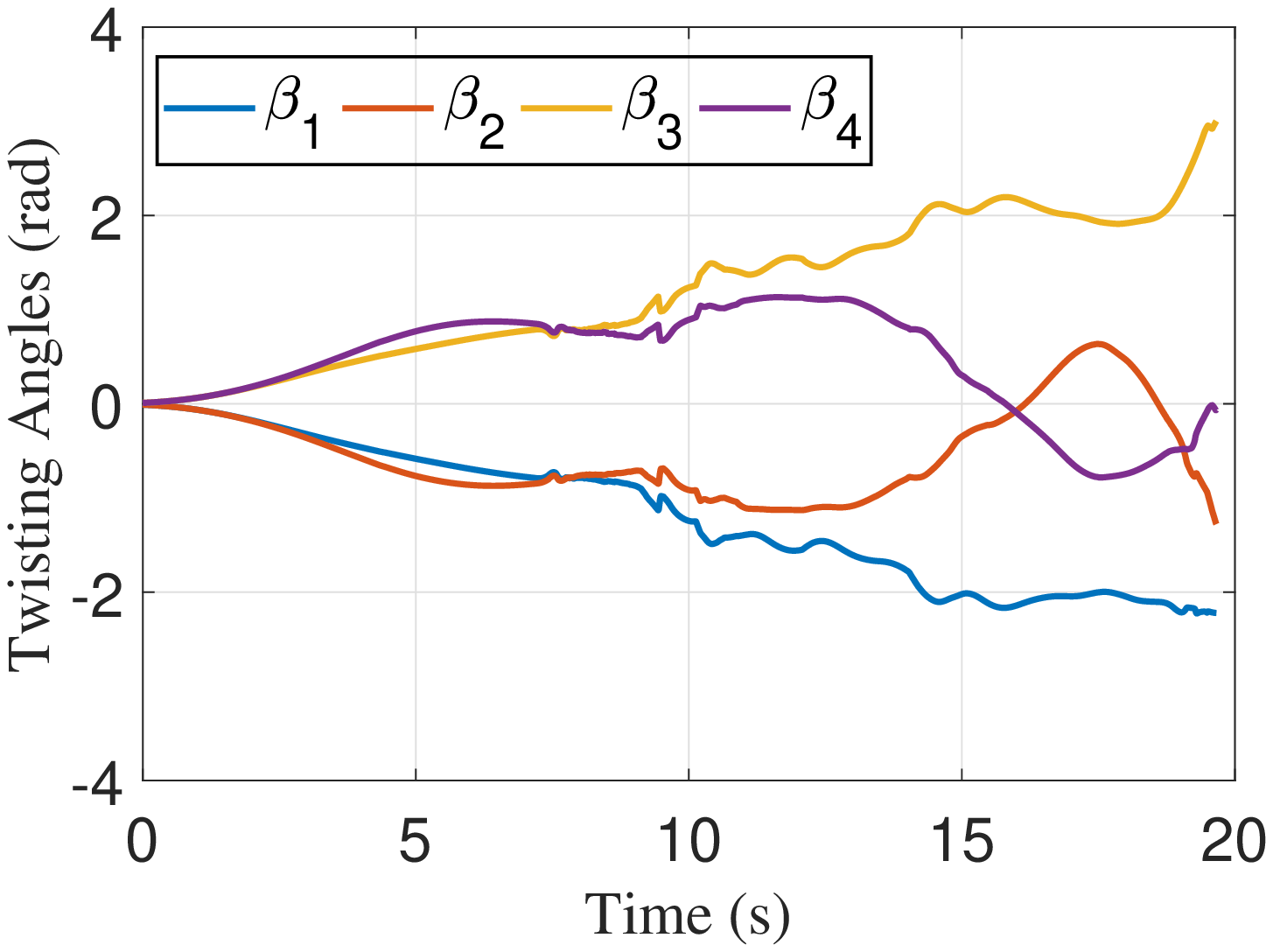}
        \caption{Four (C): Twisting Angles.}
        \label{fig:four_sim_fail_beta}
    \end{subfigure}%
    \\
    \begin{subfigure}{0.2\linewidth}
        \centering
        \includegraphics[width=\linewidth,trim=3.2cm 10.2cm 3cm 9.4cm, clip]{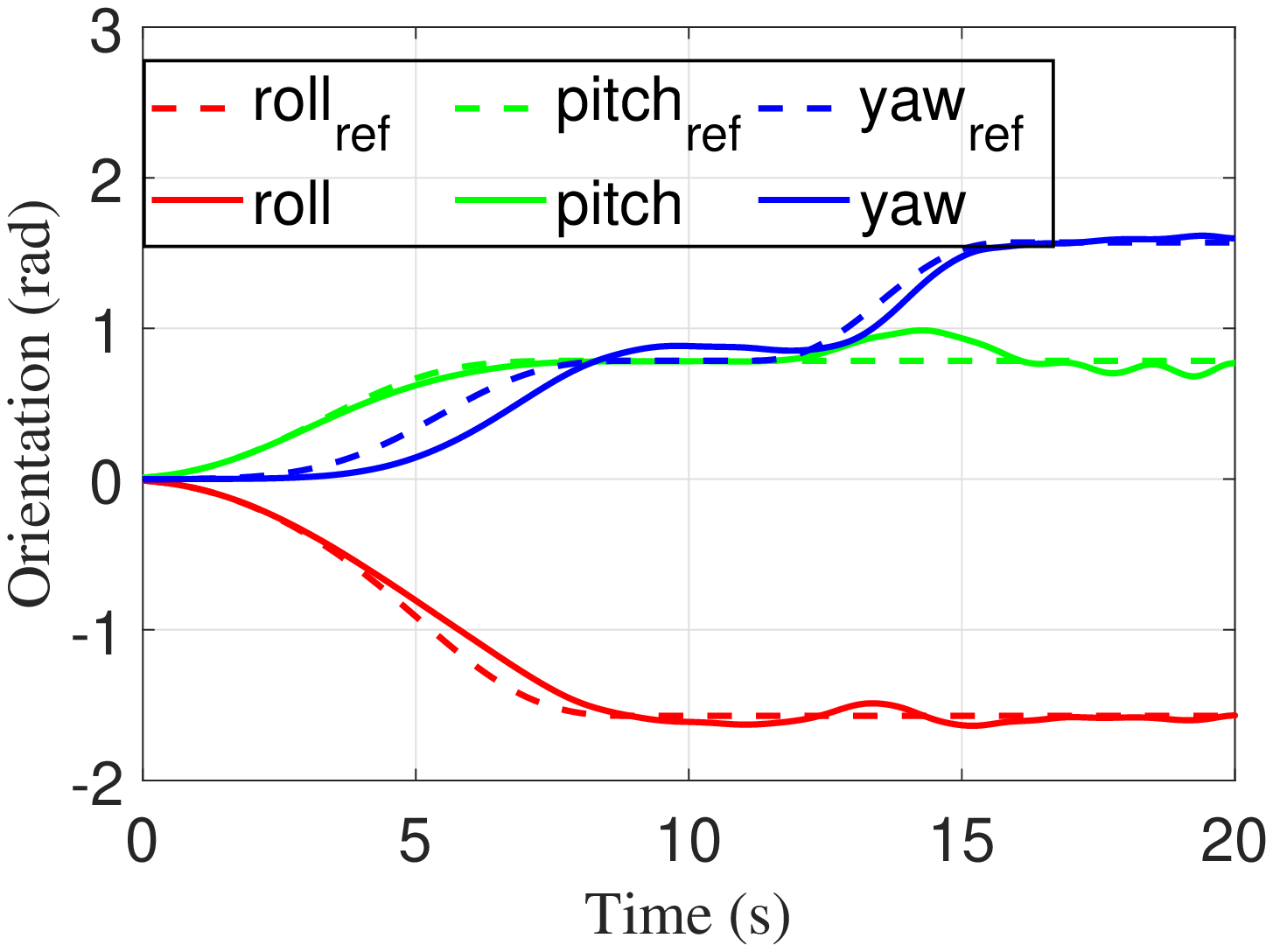}
        \caption{Four (D): Attitude.}
        \label{fig:four_sim_success_rpy}
    \end{subfigure}%
    \begin{subfigure}{0.2\linewidth}
        \centering
        \includegraphics[width=\linewidth,trim=3.2cm 10.2cm 3cm 9.4cm, clip]{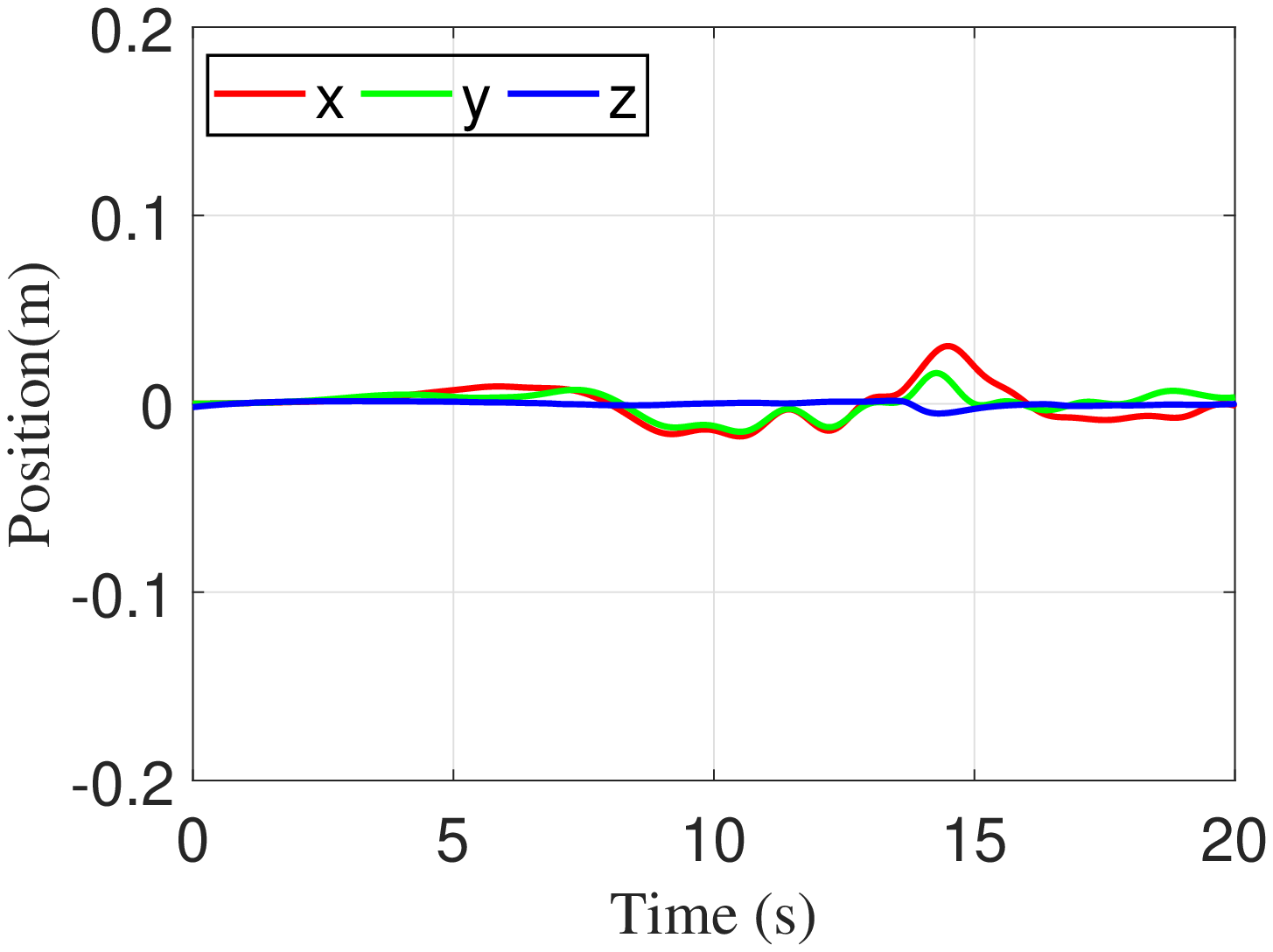}
        \caption{Four (D): Position.}
        \label{fig:four_sim_success_xyz}
    \end{subfigure}%
    \begin{subfigure}{0.2\linewidth}
        \centering
        \includegraphics[width=\linewidth,trim=3.2cm 10.2cm 3cm 9.4cm, clip]{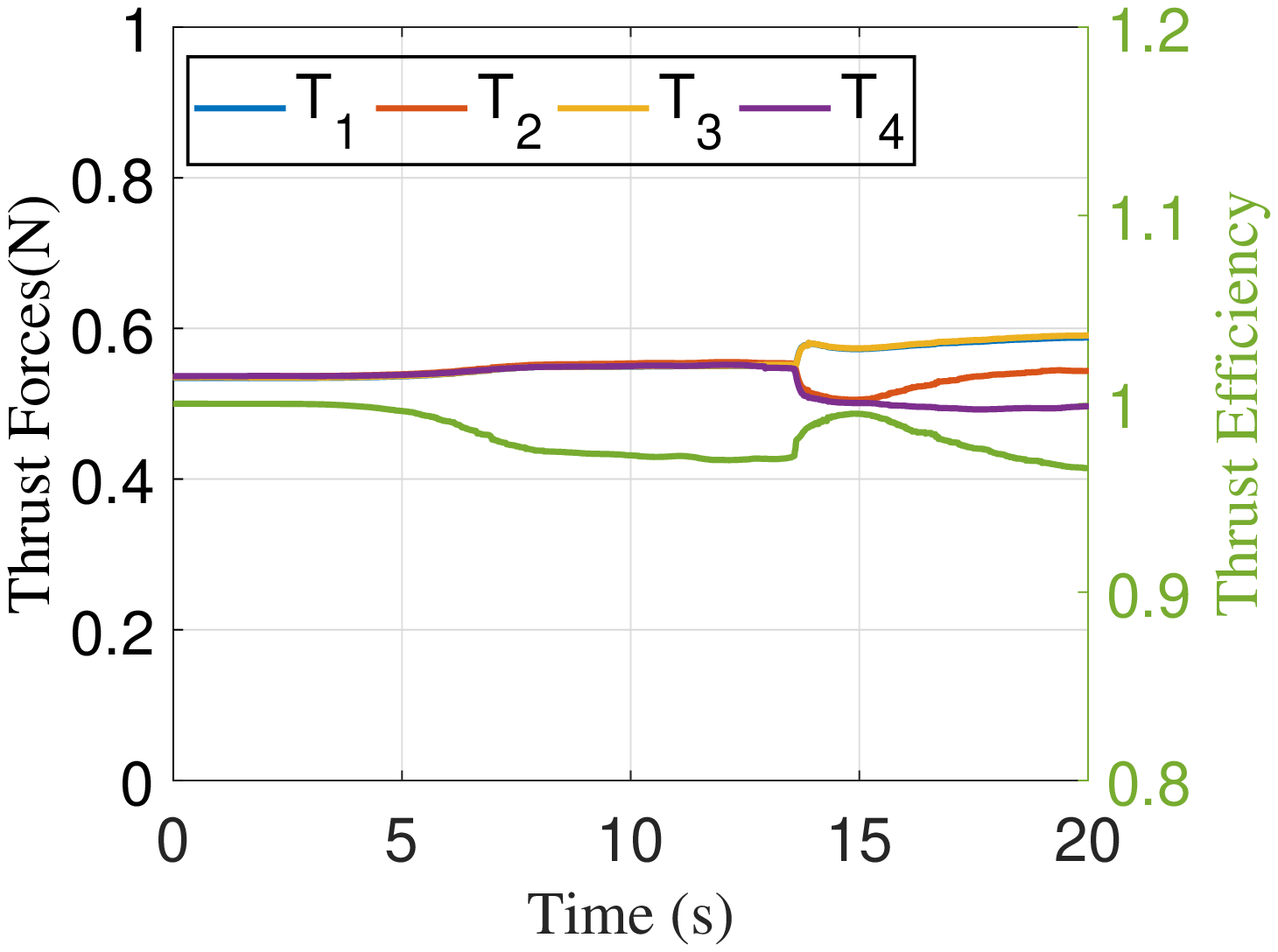}
        \caption{Four (D): Thrust.}
        \label{fig:four_sim_success_thrust}
    \end{subfigure}%
    \begin{subfigure}{0.2\linewidth}
        \centering
        \includegraphics[width=\linewidth,trim=3.2cm 10.2cm 3cm 9.4cm, clip]{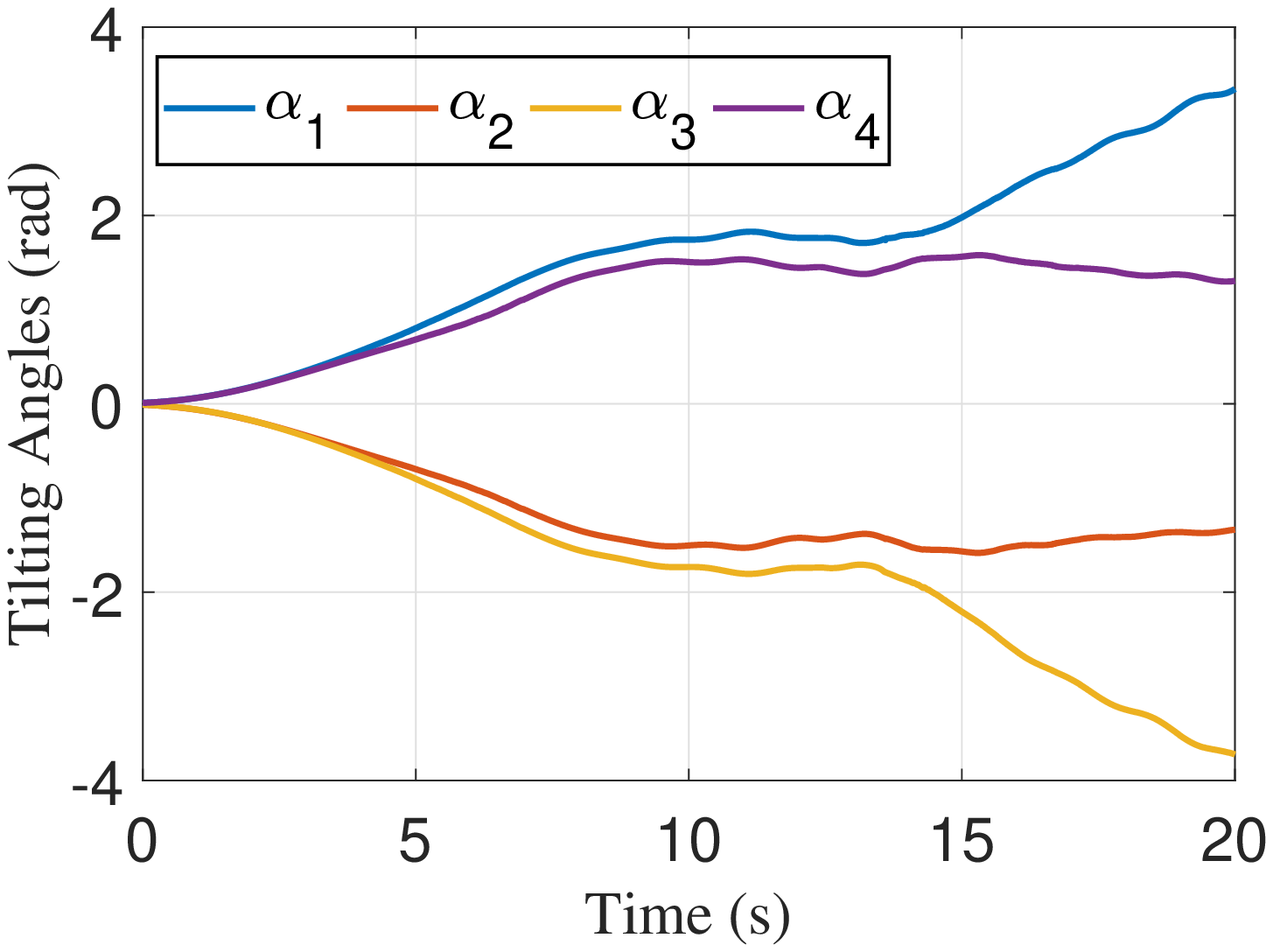}
        \caption{Four (D): Tilting Angles.}
        \label{fig:four_sim_success_alpha}
    \end{subfigure}%
    \begin{subfigure}{0.2\linewidth}
        \centering
        \includegraphics[width=\linewidth,trim=3.2cm 10.2cm 3cm 9.4cm, clip]{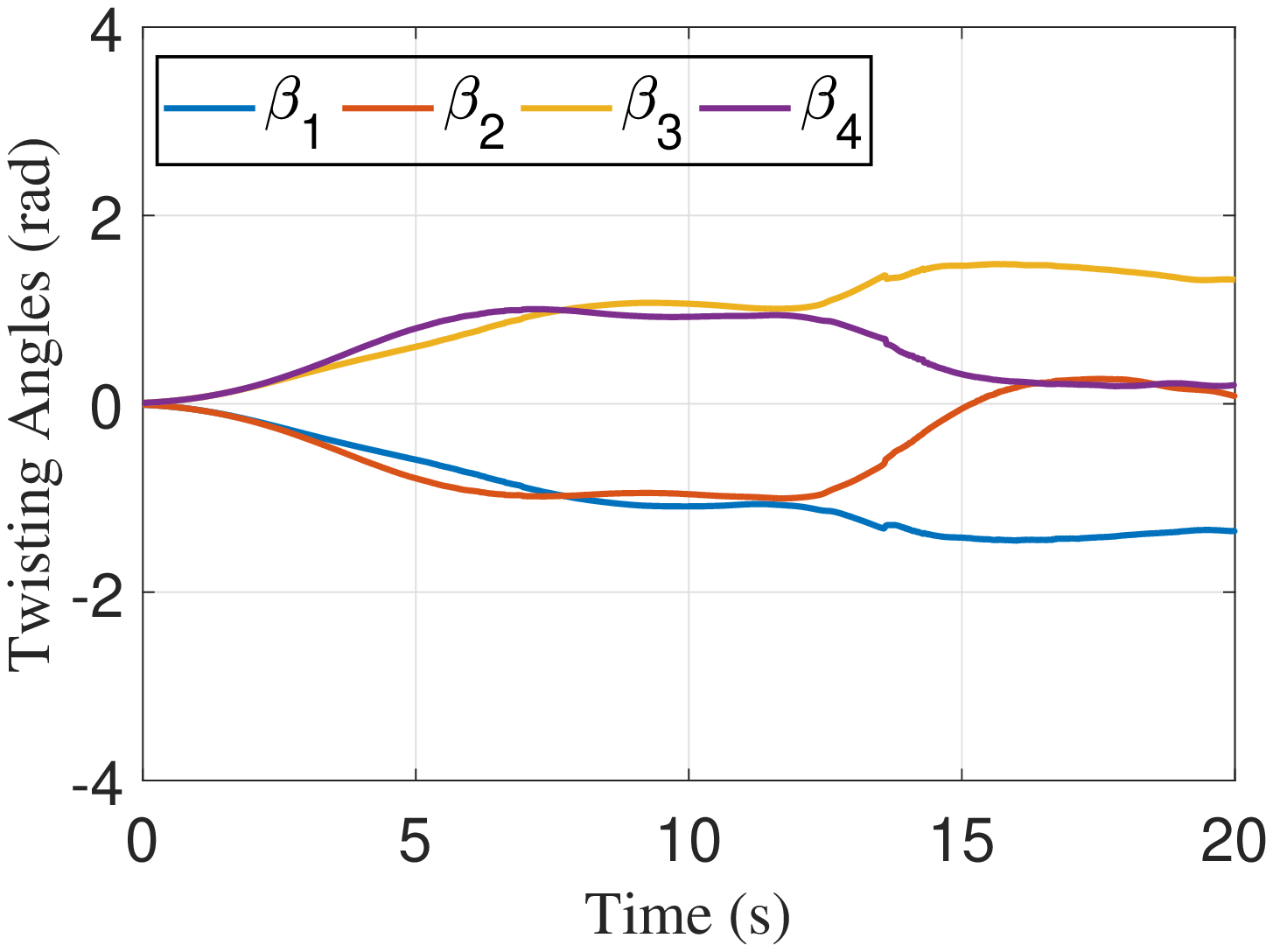}
        \caption{Four (D): Twisting Angles.}
        \label{fig:four_sim_success_beta}
    \end{subfigure}%
    \\
    \begin{subfigure}{0.2\linewidth}
        \centering
        \includegraphics[width=\linewidth,trim=3.2cm 10.2cm 3cm 9.4cm, clip]{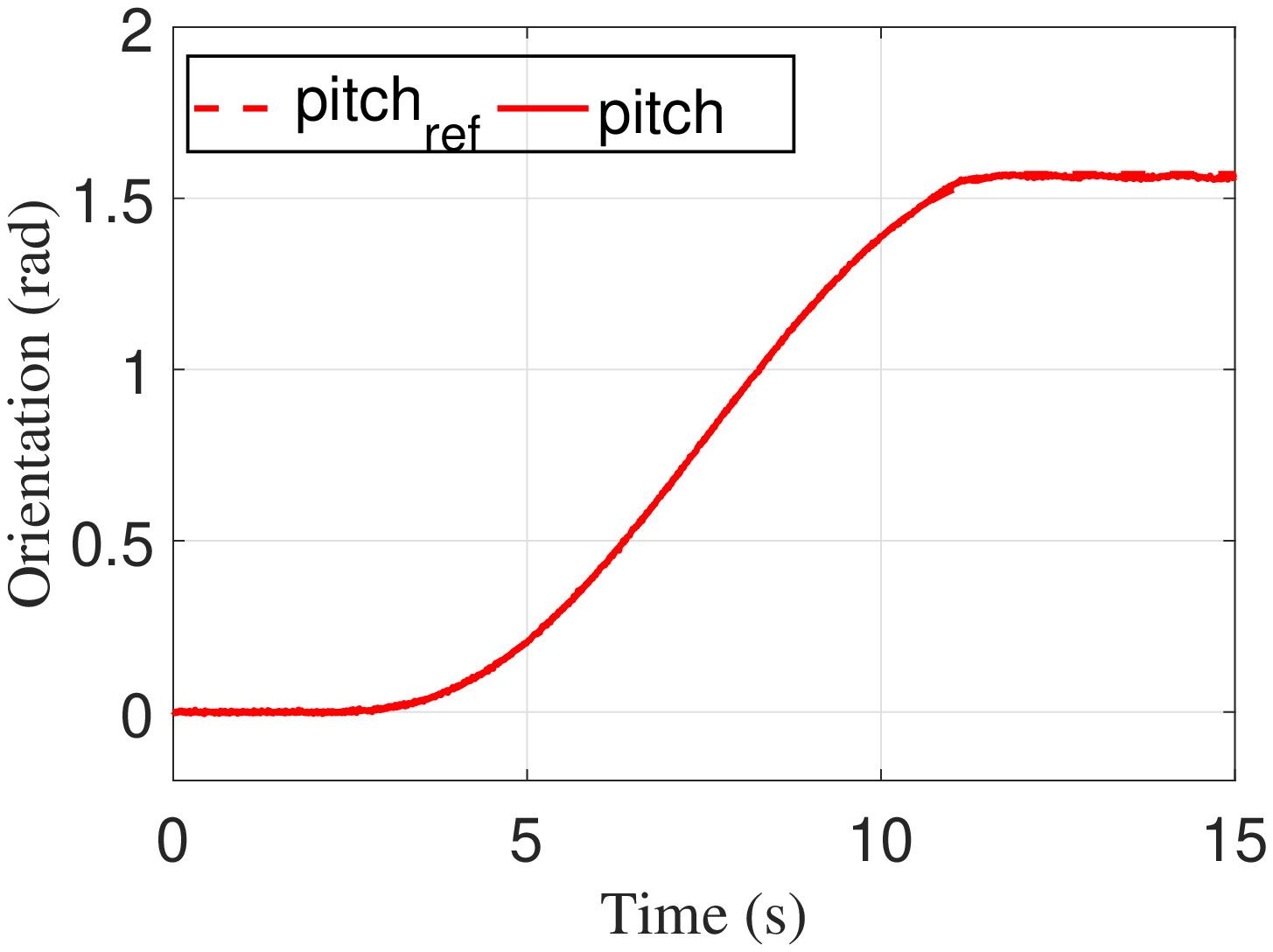}
        \caption{Six (C): Attitude.}
        \label{fig:six_sim_fail_rpy}
    \end{subfigure}%
    \begin{subfigure}{0.2\linewidth}
        \centering
        \includegraphics[width=\linewidth,trim=3.2cm 10.2cm 3cm 9.4cm, clip]{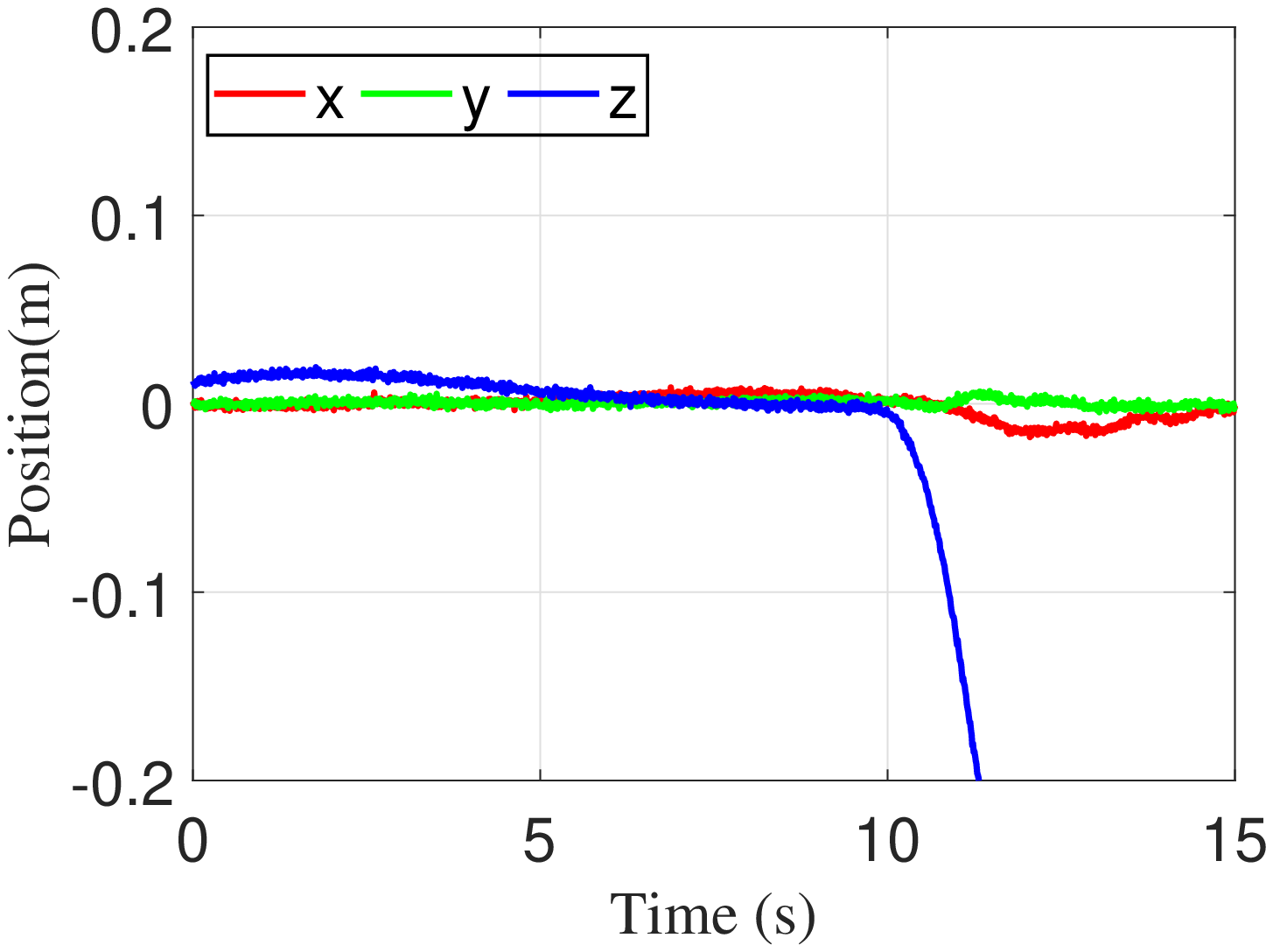}
        \caption{Six (C): Position.}
        \label{fig:six_sim_fail_xyz}
    \end{subfigure}%
    \begin{subfigure}{0.2\linewidth}
        \centering
        \includegraphics[width=\linewidth,trim=3.2cm 10.2cm 3cm 9.4cm, clip]{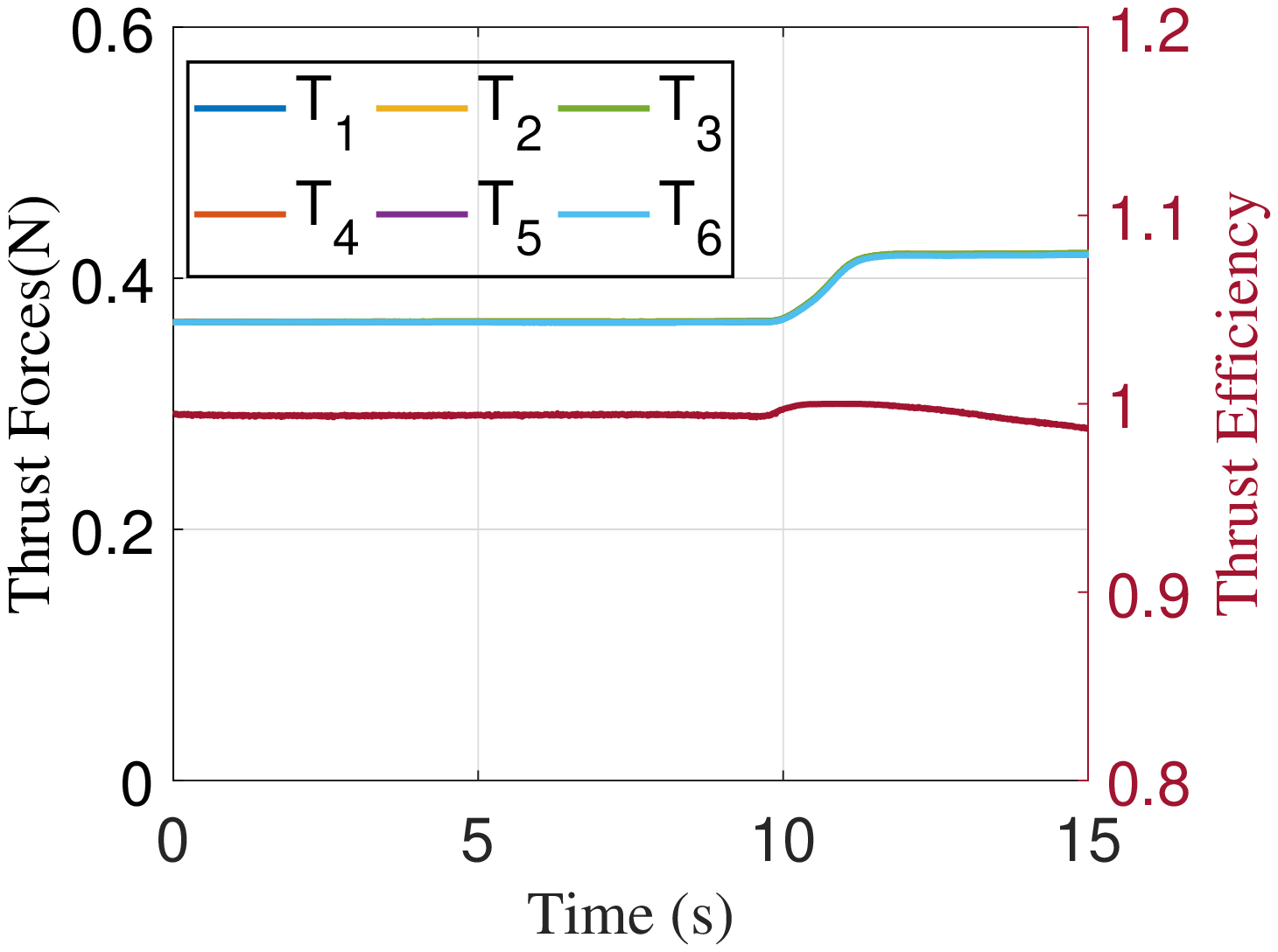}
        \caption{Six (C): Thrust.}
        \label{fig:six_sim_fail_thrust}
    \end{subfigure}%
    \begin{subfigure}{0.2\linewidth}
        \centering
        \includegraphics[width=\linewidth,trim=3.2cm 10.2cm 3cm 9.4cm, clip]{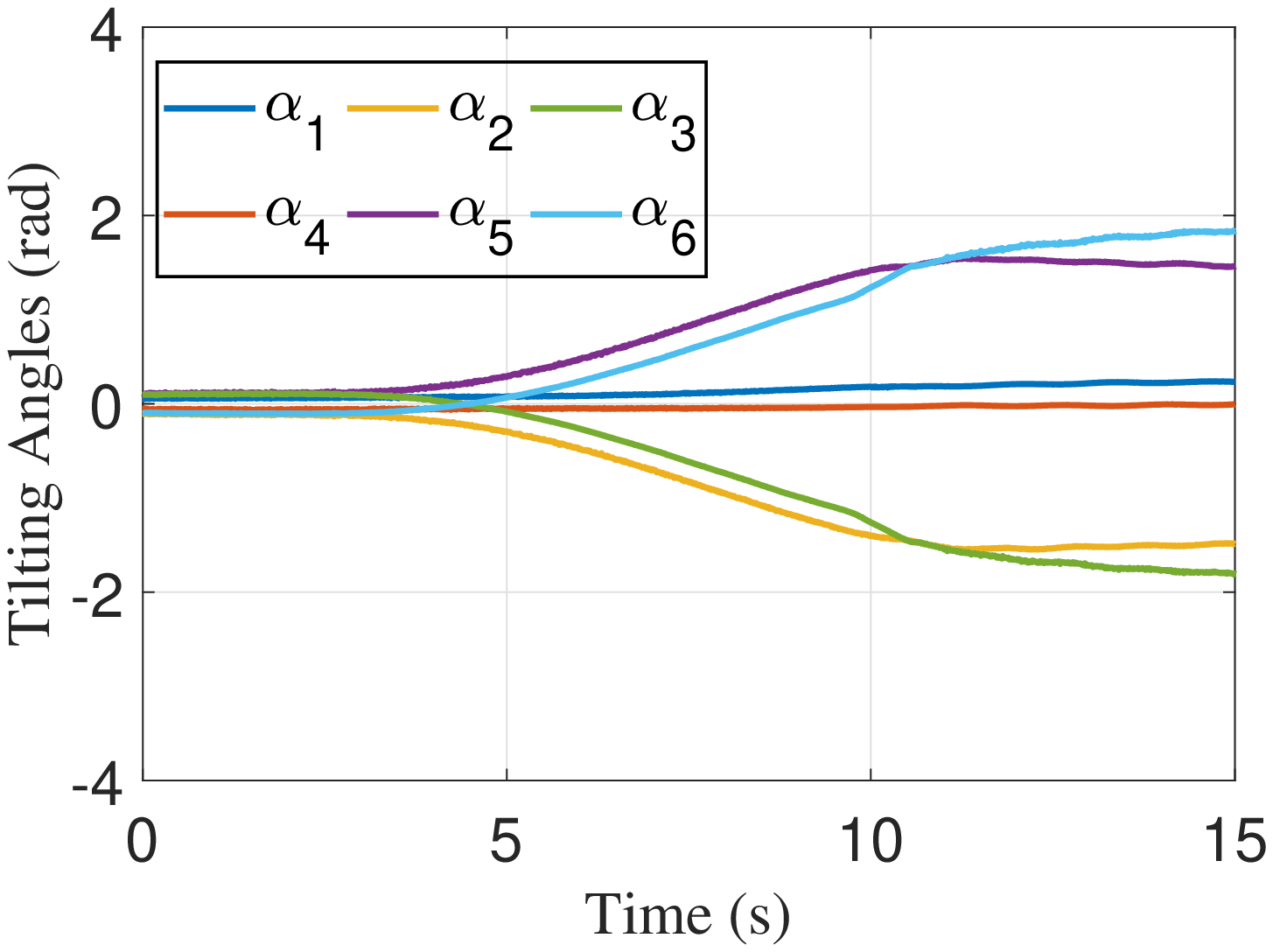}
        \caption{Six (C): Tilting Angles.}
        \label{fig:six_sim_fail_alpha}
    \end{subfigure}%
    \begin{subfigure}{0.2\linewidth}
        \centering
        \includegraphics[width=\linewidth,trim=3.2cm 10.2cm 3cm 9.4cm, clip]{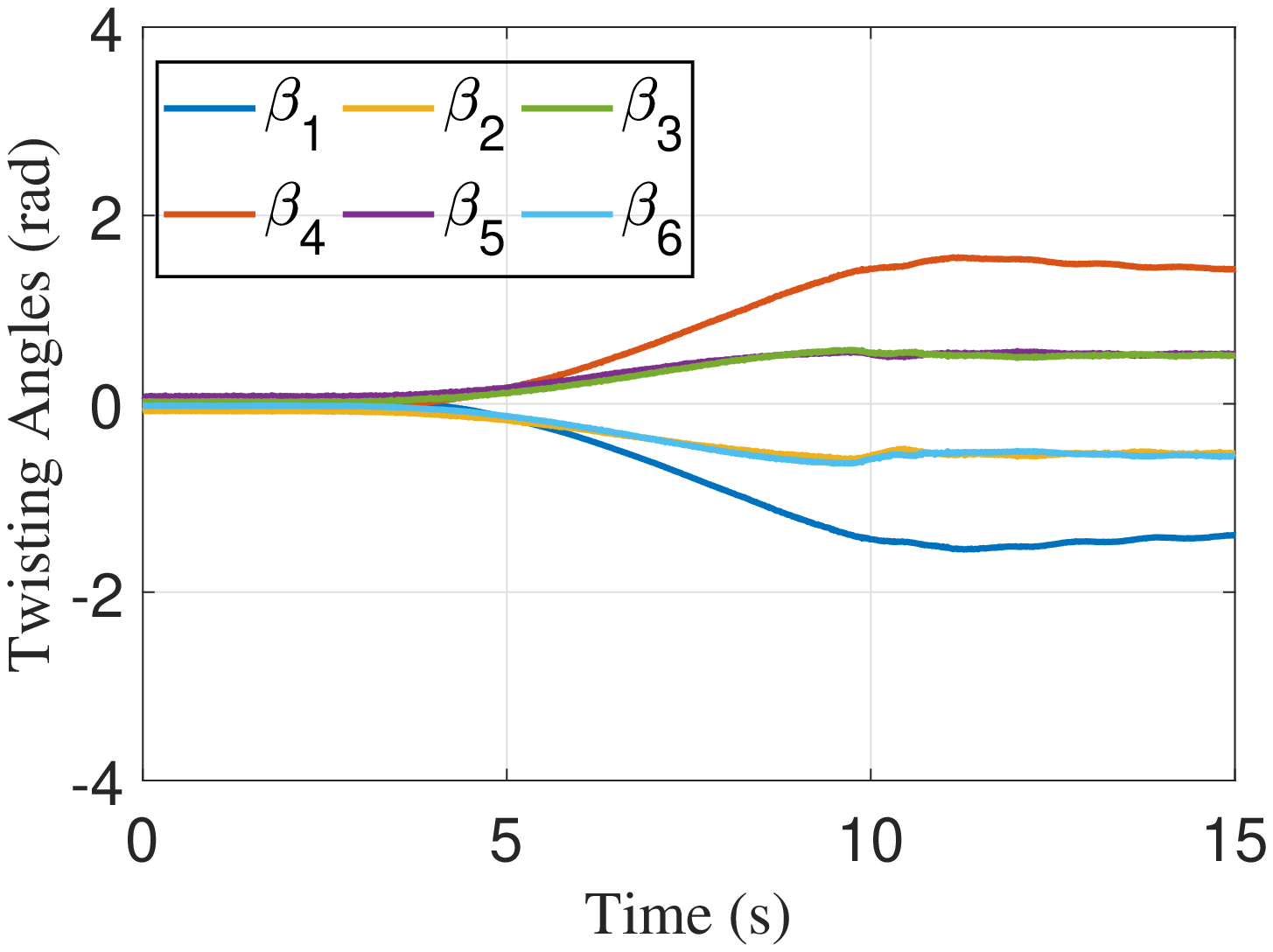}
        \caption{Six (C): Twisting Angles.}
        \label{fig:six_sim_fail_beta}
    \end{subfigure}%
    \\
    \begin{subfigure}{0.2\linewidth}
        \centering
        \includegraphics[width=\linewidth,trim=3.2cm 10.2cm 3cm 9.4cm, clip]{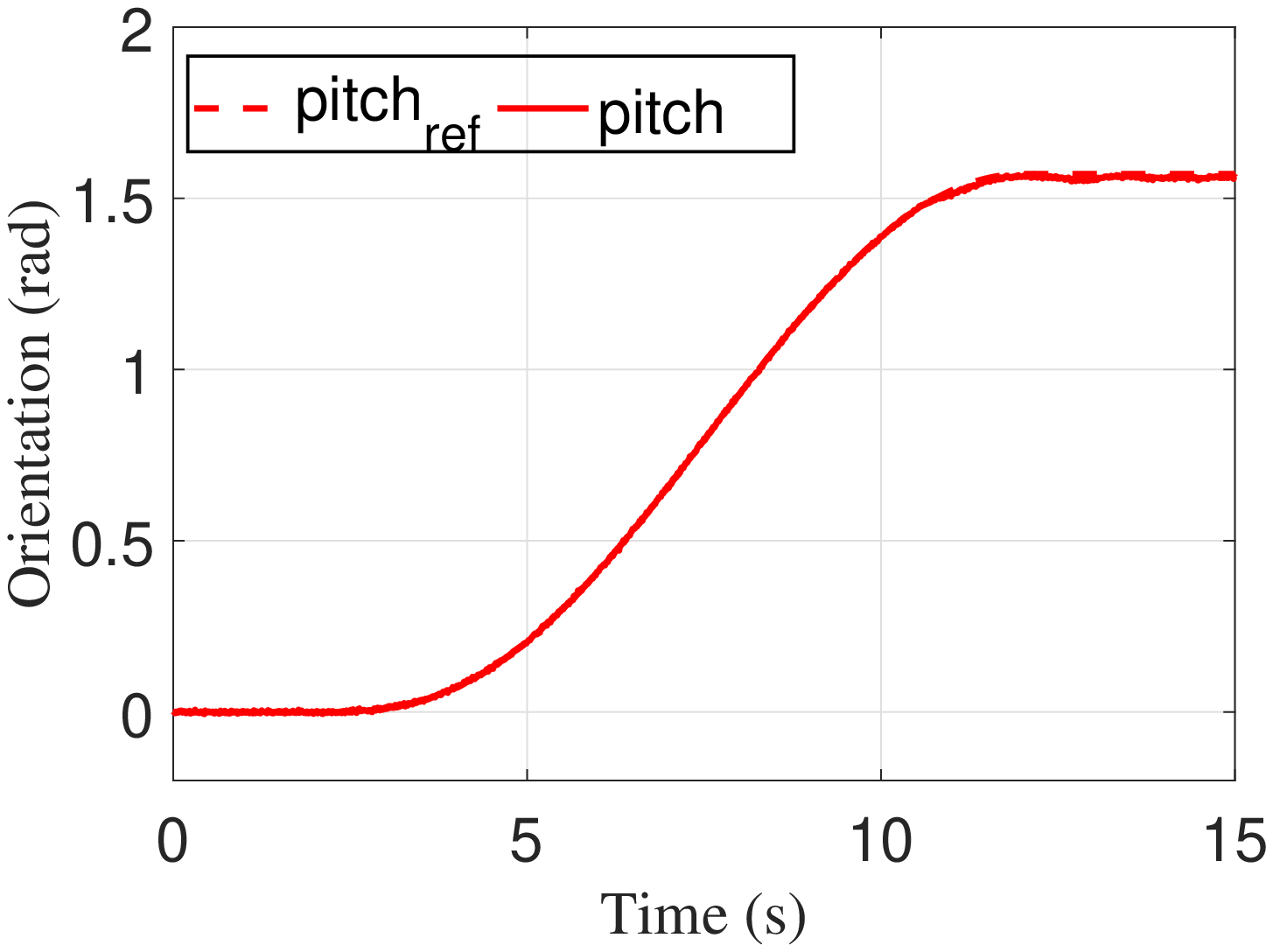}
        \caption{Six (D): Attitude.}
        \label{fig:six_sim_success_rpy}
    \end{subfigure}%
    \begin{subfigure}{0.2\linewidth}
        \centering
        \includegraphics[width=\linewidth,trim=3.2cm 10.2cm 3cm 9.4cm, clip]{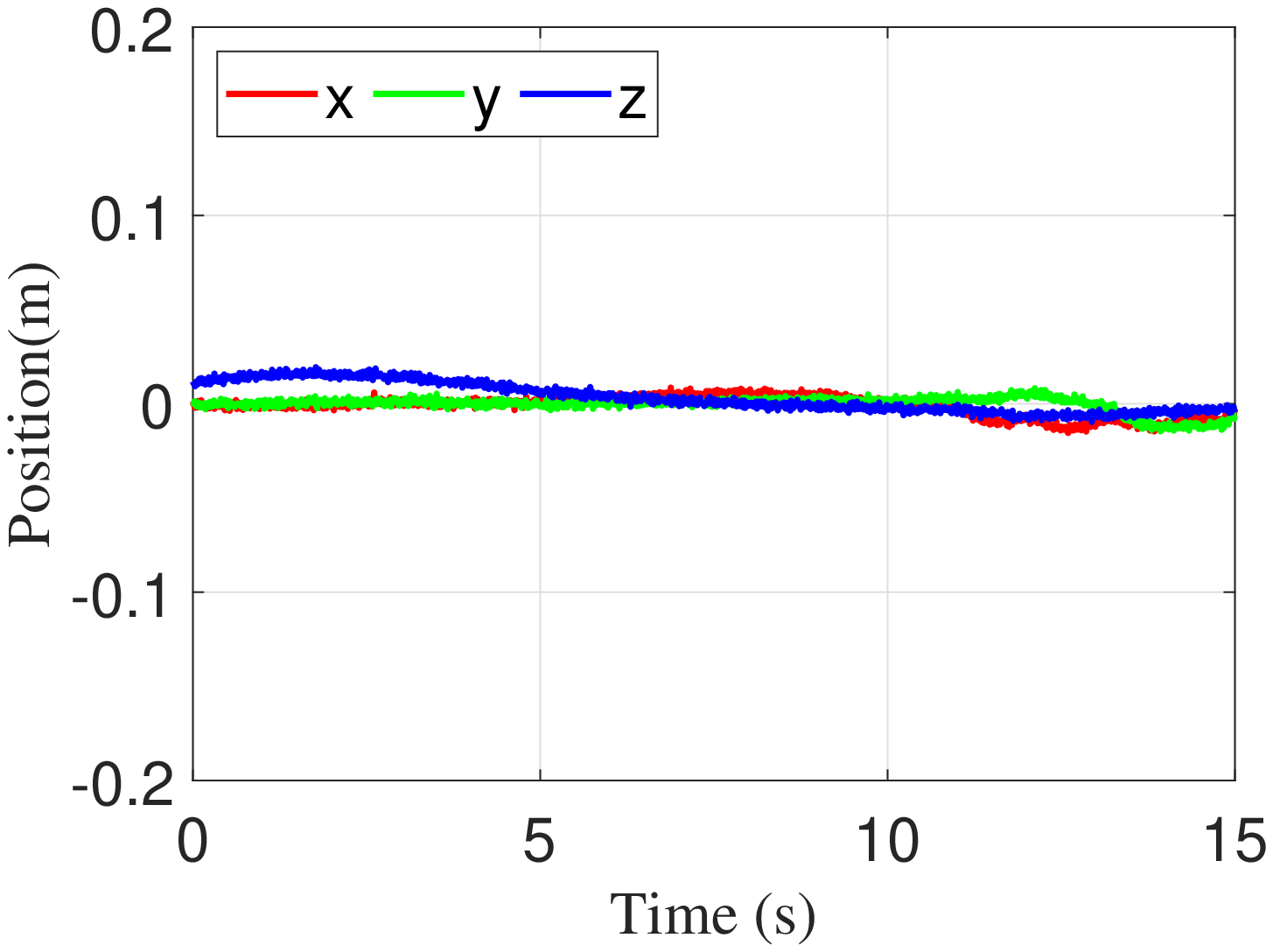}
        \caption{Six (D): Position.}
        \label{fig:six_sim_success_xyz}
    \end{subfigure}%
    \begin{subfigure}{0.2\linewidth}
        \centering
        \includegraphics[width=\linewidth,trim=3.2cm 10.2cm 3cm 9.4cm, clip]{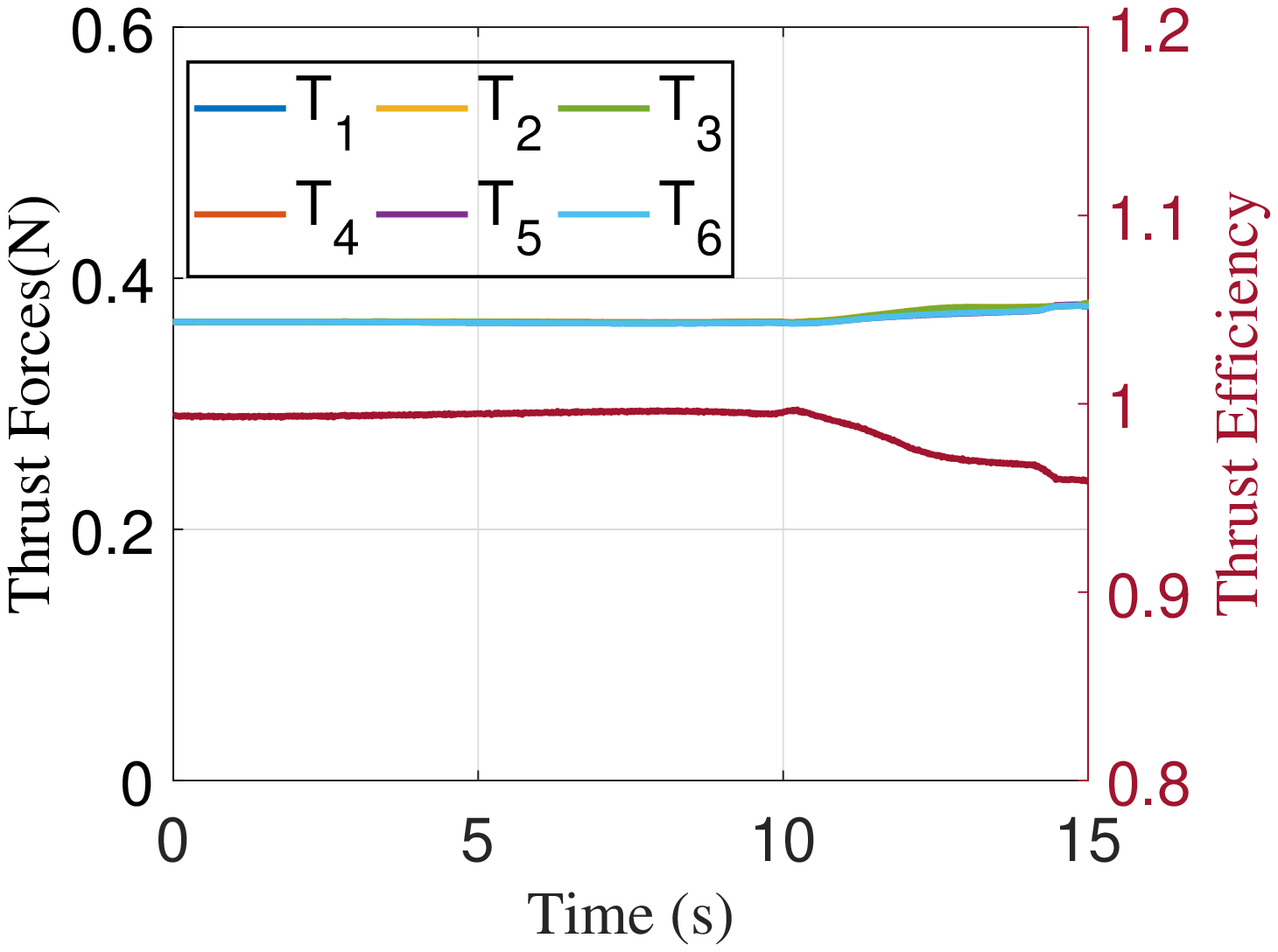}
        \caption{Six (D): Thrust.}
        \label{fig:six_sim_success_thrust}
    \end{subfigure}%
    \begin{subfigure}{0.2\linewidth}
        \centering
        \includegraphics[width=\linewidth,trim=3.2cm 10.2cm 3cm 9.4cm, clip]{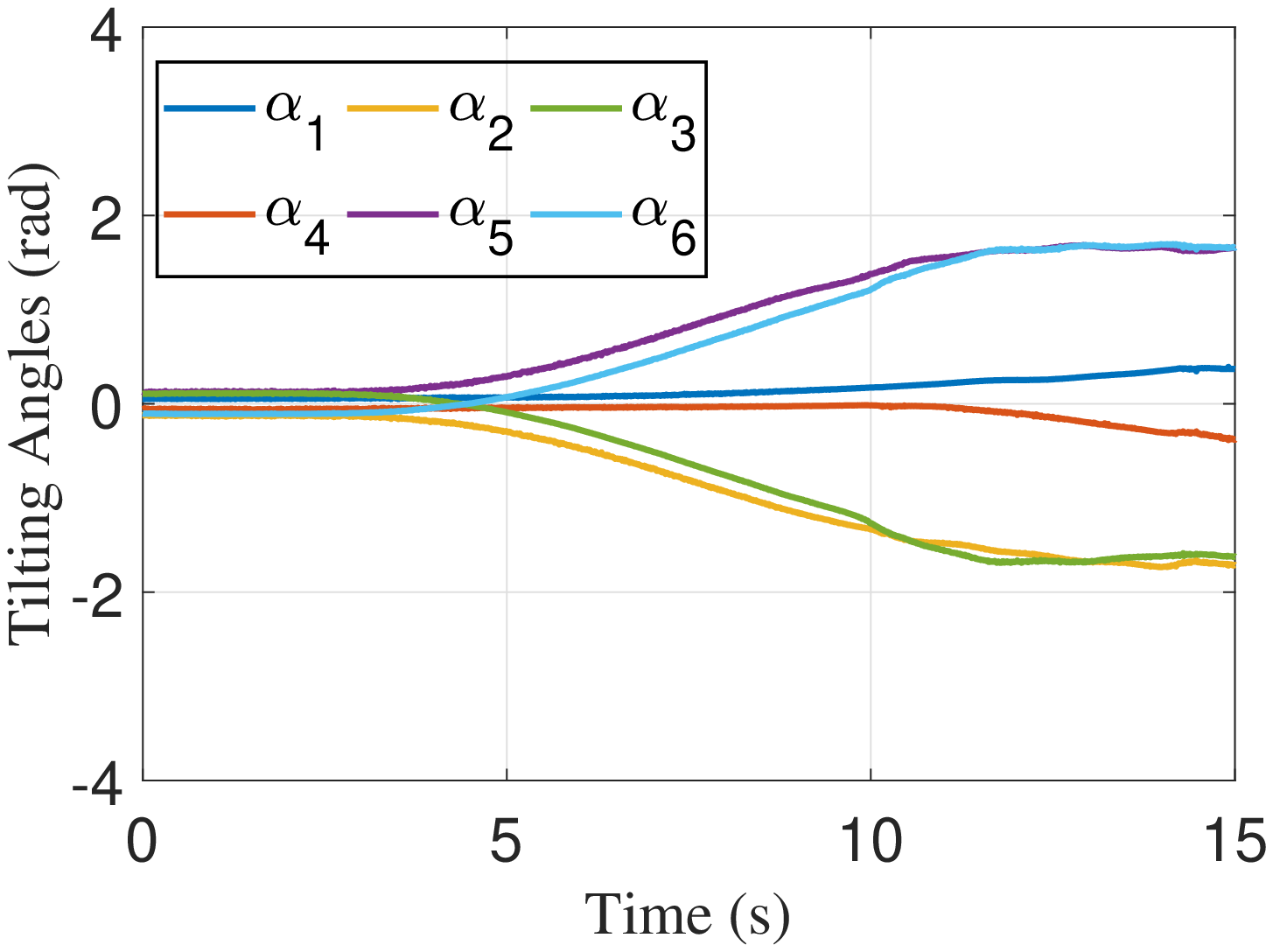}
        \caption{Six (D): Tilting Angles.}
        \label{fig:six_sim_success_alpha}
    \end{subfigure}%
    \begin{subfigure}{0.2\linewidth}
        \centering
        \includegraphics[width=\linewidth,trim=3.2cm 10.2cm 3cm 9.4cm, clip]{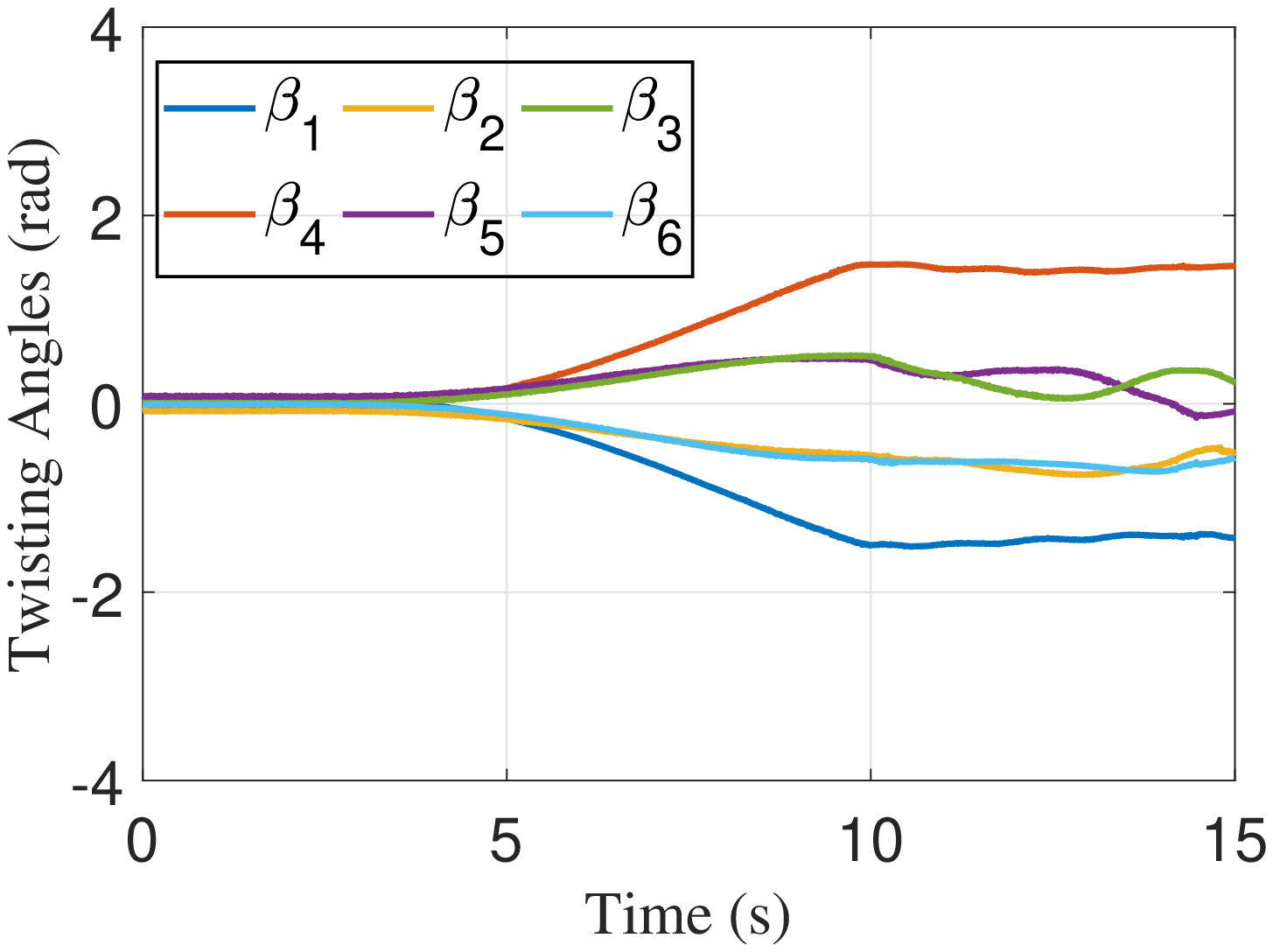}
        \caption{Six (D): Twisting Angles.}
        \label{fig:six_sim_success_beta}
    \end{subfigure}%
    \caption{\textbf{Simulation: Comparison of conventional and downwash-aware control allocation on two over-actuated \ac{uav} platforms.} C and D denotes conventional and downwash-aware control allocation, respectively.}
    \label{fig:sim_result}
\end{figure*}

\cref{fig:sim_result} summarizes the simulation results of two over-actuated \ac{uav} platforms with the proposed downwash effect model introduced in \cref{sec:downwashmodel}. For the platform that has four 3-\ac{dof} thrust generators, a reference attitude trajectory is designed where the downwash effects occur twice (\cref{fig:four_sim_fail_rpy}). As we can see, the downwash effect first appears at about 9s, when the platform is rotated at 90 degree along the axis $[-\frac{\sqrt{2}}{2},\frac{\sqrt{2}}{2},0]$; $T_4$ and $T_1$, as well as $T_3$ and $T_2$, aligned vertically (two pairs of downwash effect). With conventional allocation framework, the downwash flows significantly influence the control of the platform; we noticed a drop in Z axis with about $0.15m$, and the control performance of other 5-\ac{dof} is also deteriorated (\cref{fig:four_sim_fail_xyz}). Later, another downwash effect appears at about $16s$ ($T_4$ and $T_2$ aligned vertically), which finally makes the platform unstable. Using the proposed the downwash-aware allocation framework, the platform tracks the reference trajectory stably (\cref{fig:four_sim_success_rpy,fig:four_sim_success_xyz}) and maintain a high thrust efficiency (\cref{fig:four_sim_success_thrust}). Please see also \cref{fig:downwash_illustration} for better visualization.

For the platform that has six 3-\ac{dof} thrust generators, a 90 degree pitch reference trajectory (\cref{fig:six_sim_fail_rpy}) is utilized, and three pairs of downwash effects happen at the final attitude. With the conventional allocation framework, although the platform is still stable, we noticed a $0.3m$ drop in Z-axis with more than $5s$ to compensate for position control (\cref{fig:six_sim_fail_xyz}). Further, as we can see in \cref{fig:six_sim_fail_thrust}, this framework needs more thrusts to compensate for the downwash aerodynamics, inefficient in terms of energy. With our proposed downwash-aware allocation framework, the control in the Z-axis is maintained, and the thrust is not increased by much for downwash avoidance. In summary, by exploring the entire allocation space, the downwash effects are avoided, and the high thrust efficiency is maintained. 

\subsection{Experiment Results}

\begin{figure*}[t!]
    \centering
    \begin{subfigure}{0.2\linewidth}
        \centering
        \includegraphics[width=\linewidth,trim=3.2cm 10.2cm 3cm 9.4cm, clip]{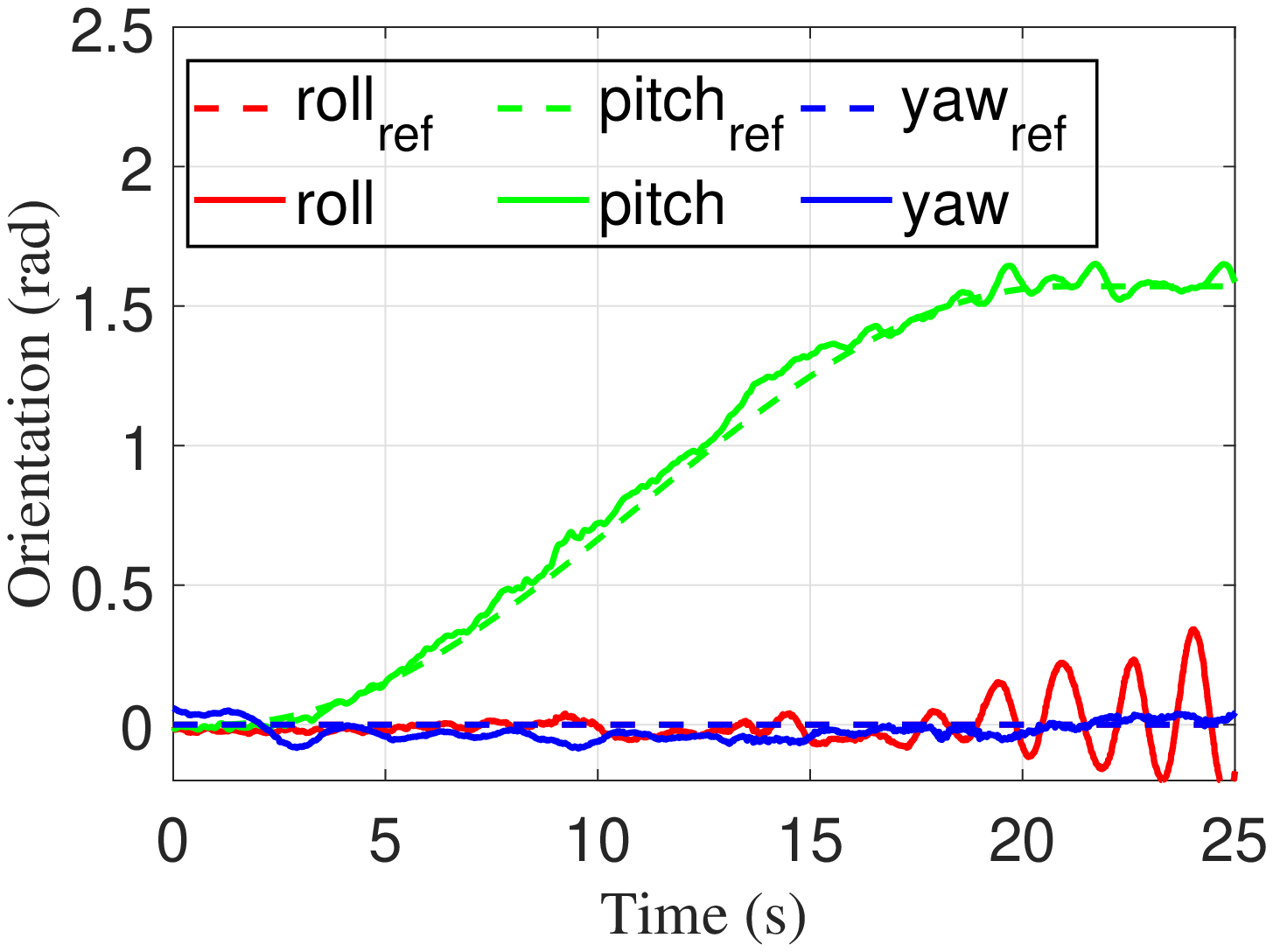}
        \caption{Four (C): Attitude.}
        \label{fig:four_bad_rpy}
    \end{subfigure}%
    \begin{subfigure}{0.2\linewidth}
        \centering
        \includegraphics[width=\linewidth,trim=3.2cm 10.2cm 3cm 9.4cm, clip]{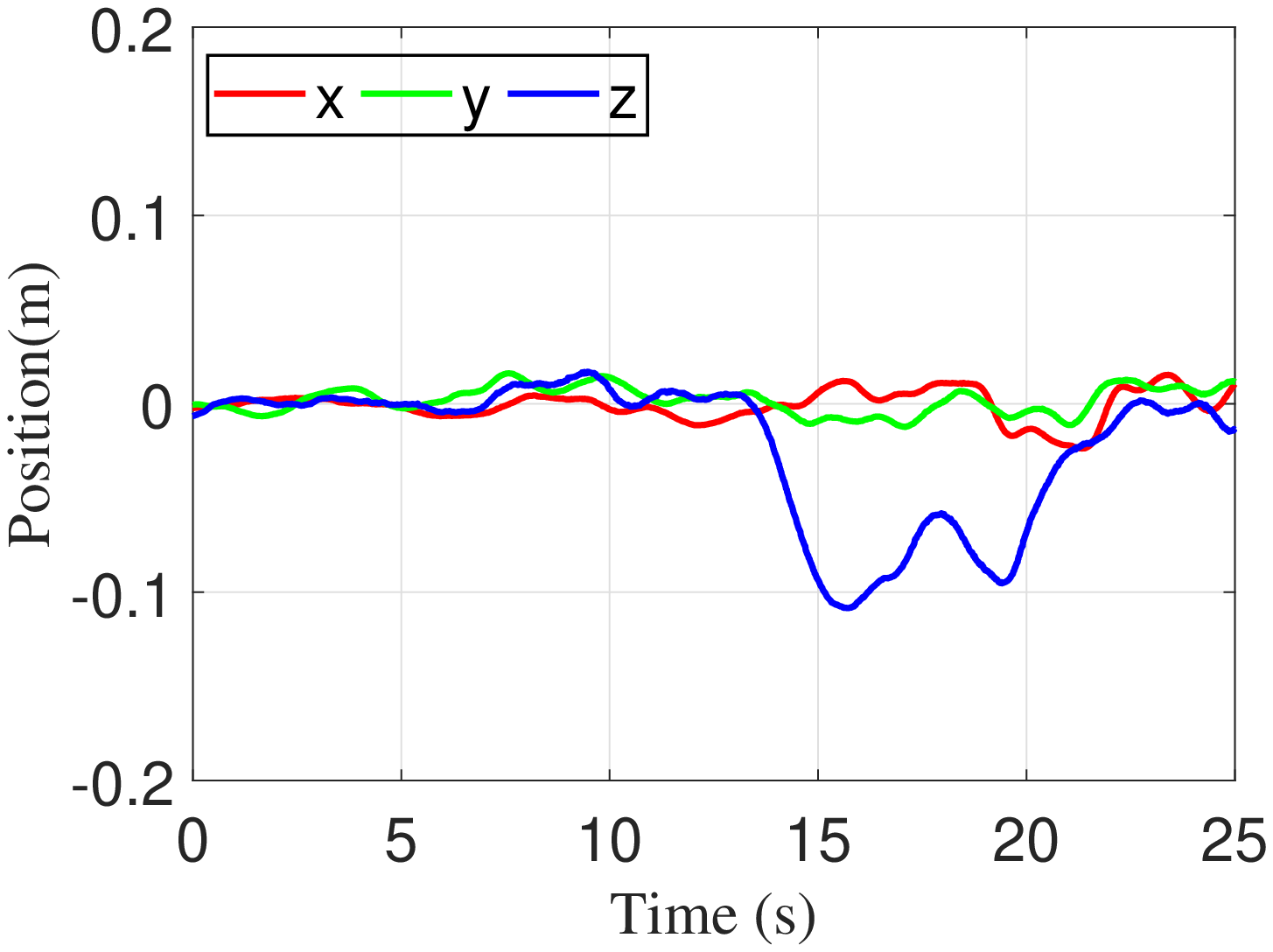}
        \caption{Four (C): Position.}
        \label{fig:four_bad_xyz}
    \end{subfigure}%
    \begin{subfigure}{0.2\linewidth}
        \centering
        \includegraphics[width=\linewidth,trim=3.2cm 10.2cm 3cm 9.4cm, clip]{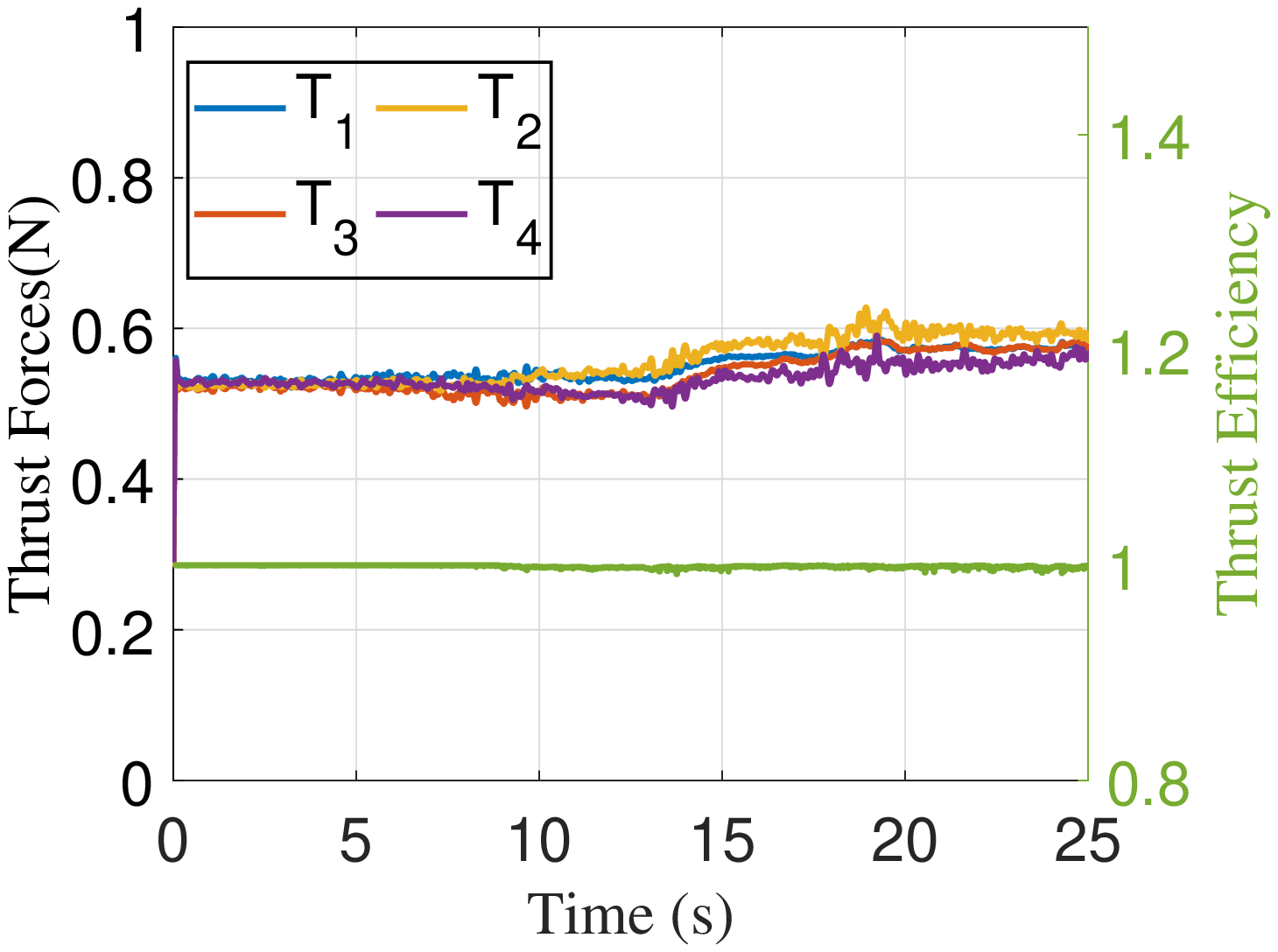}
        \caption{Four (C): Thrust.}
        \label{fig:four_bad_thrust}
    \end{subfigure}%
    \begin{subfigure}{0.2\linewidth}
        \centering
        \includegraphics[width=\linewidth,trim=3.2cm 10.2cm 3cm 9.4cm, clip]{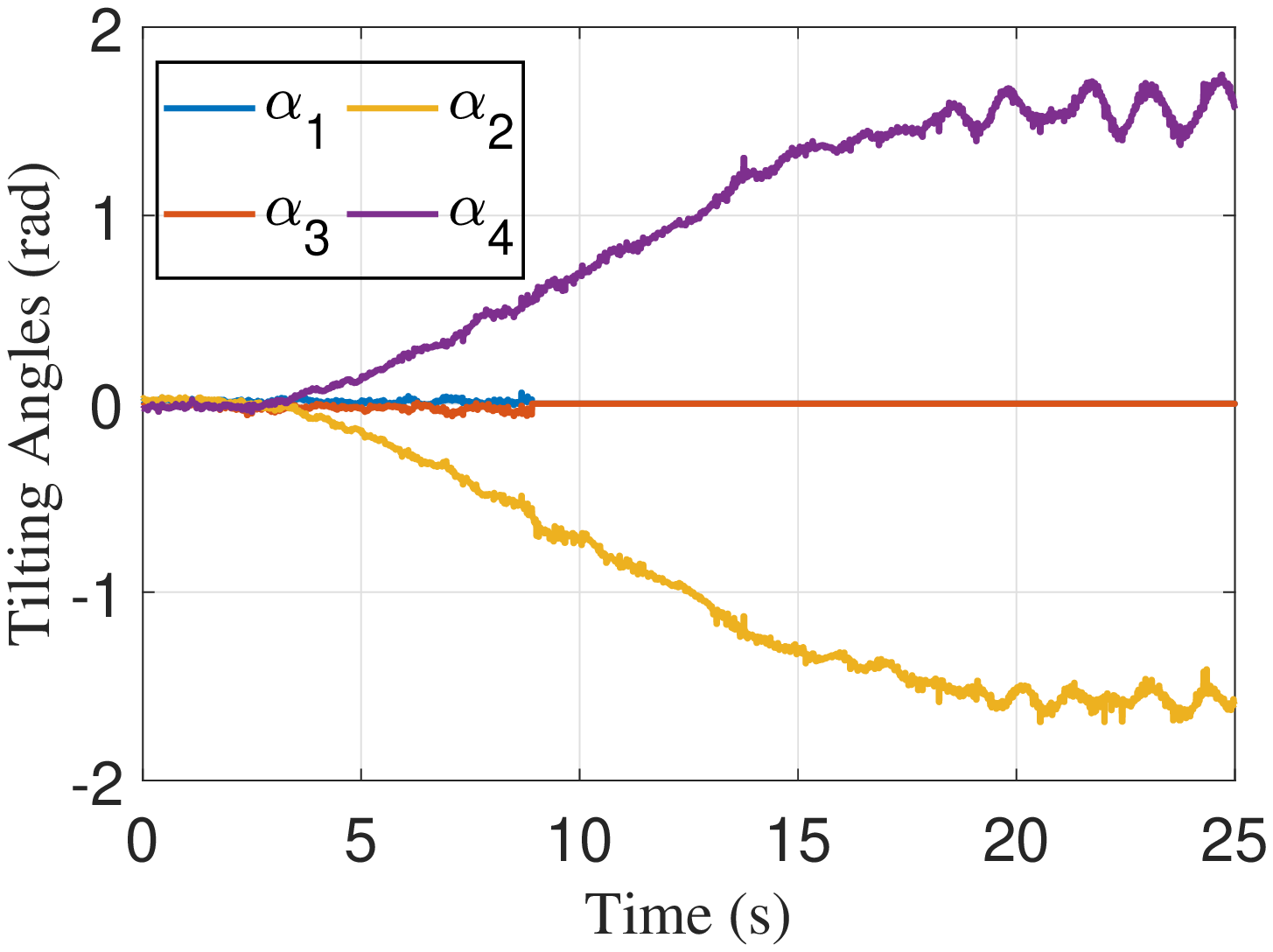}
        \caption{Four (C): Tilting Angles.}
        \label{fig:four_bad_alpha}
    \end{subfigure}%
    \begin{subfigure}{0.2\linewidth}
        \centering
        \includegraphics[width=\linewidth,trim=3.2cm 10.2cm 3cm 9.4cm, clip]{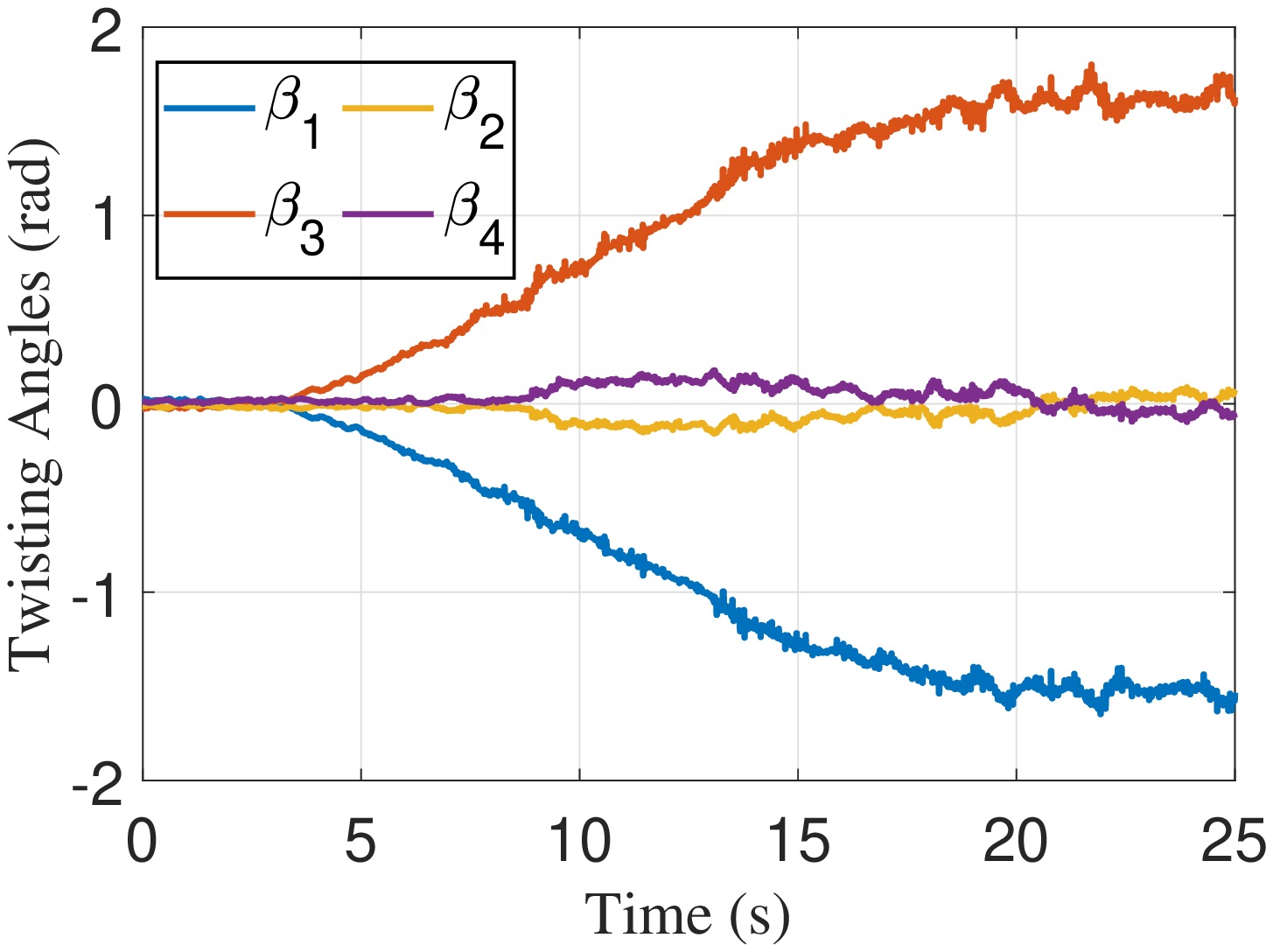}
        \caption{Four (C): Twisting Angles.}
        \label{fig:four_bad_beta}
    \end{subfigure}%
    \\
    \begin{subfigure}{0.2\linewidth}
        \centering
        \includegraphics[width=\linewidth,trim=3.2cm 10.2cm 3cm 9.4cm, clip]{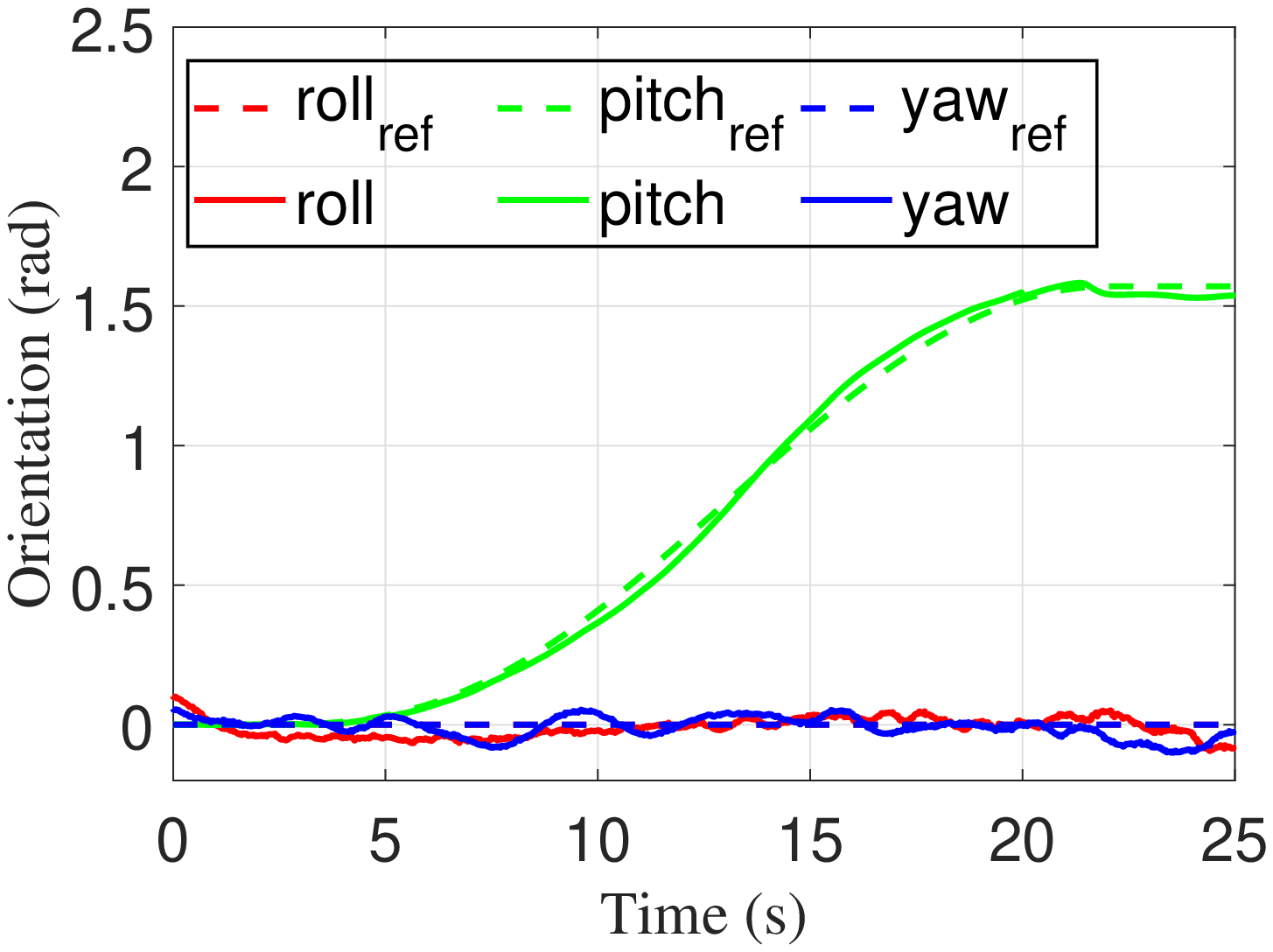}
        \caption{Four (D): Attitude.}
        \label{fig:four_success_rpy}
    \end{subfigure}%
    \begin{subfigure}{0.2\linewidth}
        \centering
        \includegraphics[width=\linewidth,trim=3.2cm 10.2cm 3cm 9.4cm, clip]{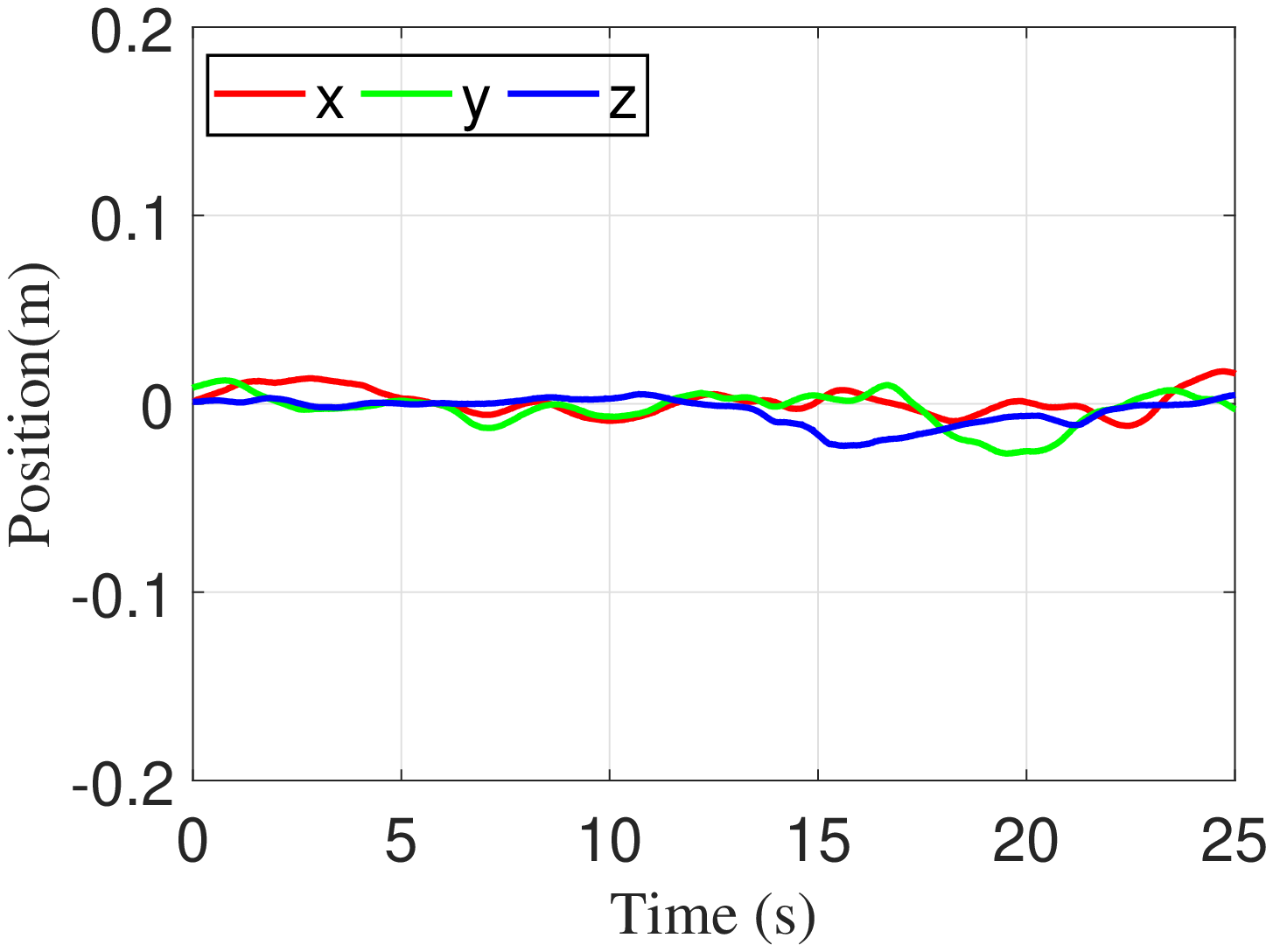}
        \caption{Four (D): Position.}
        \label{fig:four_success_xyz}
    \end{subfigure}%
    \begin{subfigure}{0.2\linewidth}
        \centering
        \includegraphics[width=\linewidth,trim=3.2cm 10.2cm 3cm 9.4cm, clip]{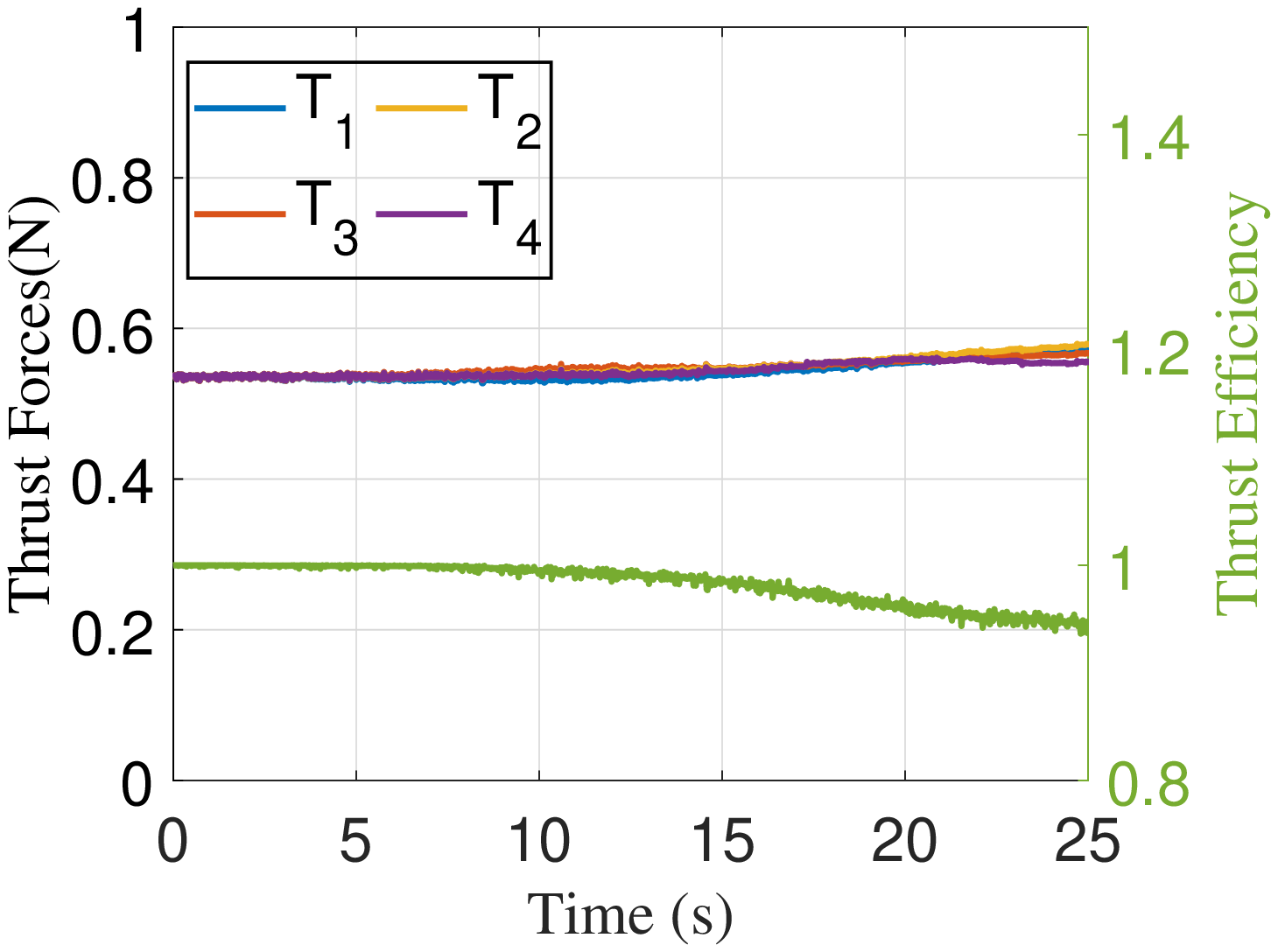}
        \caption{Four (D): Thrust.}
        \label{fig:four_success_thrust}
    \end{subfigure}%
    \begin{subfigure}{0.2\linewidth}
        \centering
        \includegraphics[width=\linewidth,trim=3.2cm 10.2cm 3cm 9.4cm, clip]{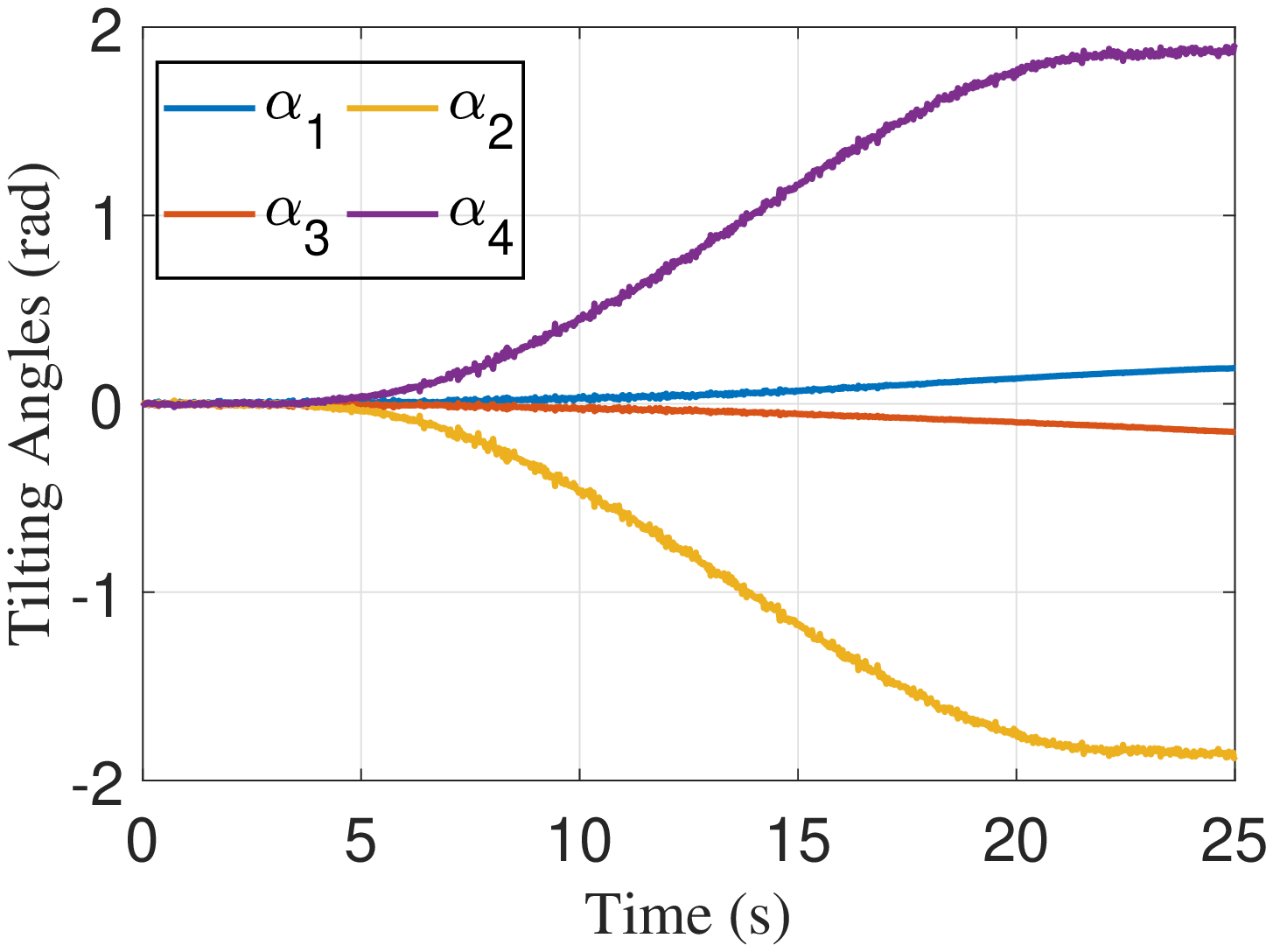}
        \caption{Four (D): Tilting Angles.}
        \label{fig:four_success_alpha}
    \end{subfigure}%
    \begin{subfigure}{0.2\linewidth}
        \centering
        \includegraphics[width=\linewidth,trim=3.2cm 10.2cm 3cm 9.4cm, clip]{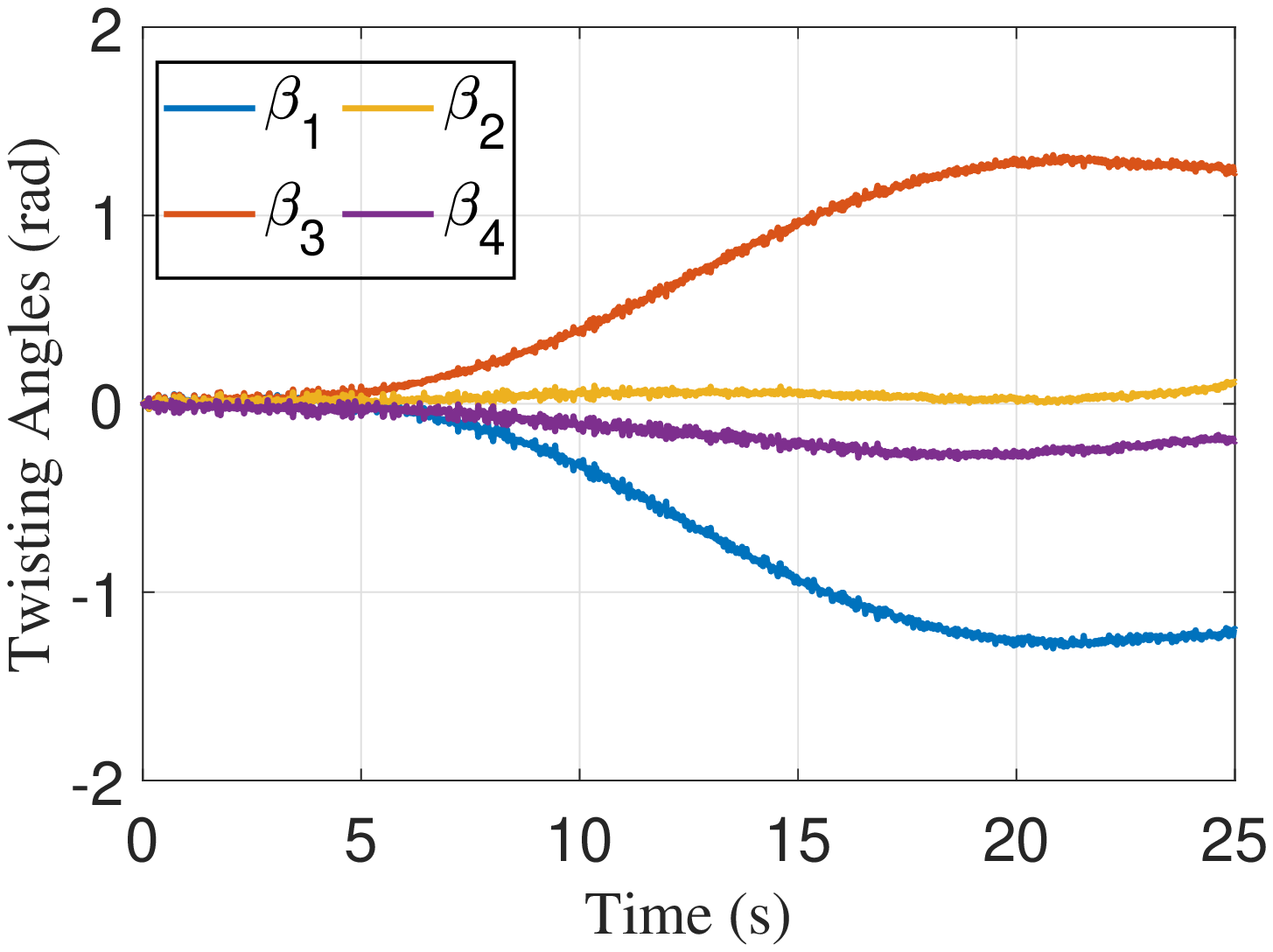}
        \caption{Four (D): Twisting Angles.}
        \label{fig:four_success_beta}
    \end{subfigure}%
    \\
    \begin{subfigure}{0.8\linewidth}
        \centering
        \includegraphics[width=\linewidth,trim=1cm 6.5cm 1cm 7cm, clip]{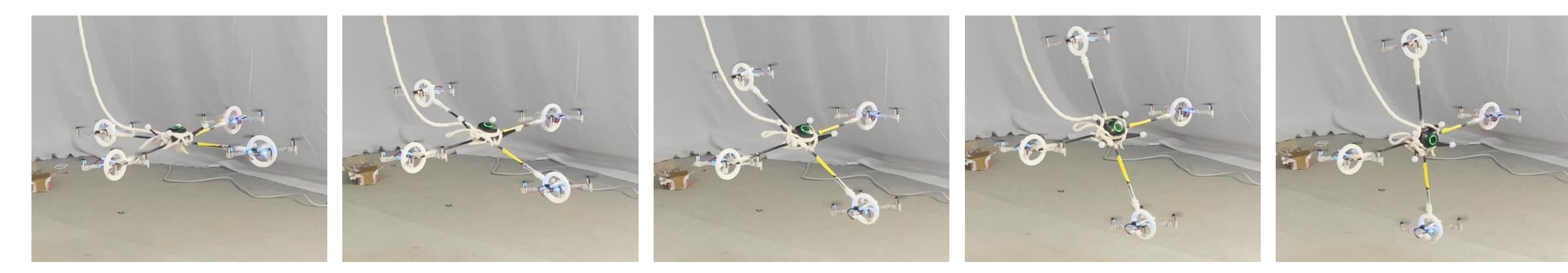}
        \caption{Four (D): Video frames.}
        \label{fig:clips}
    \end{subfigure}%
    \caption{\textbf{Experiment: Comparison of conventional and downwash-aware control allocation on the over-actuated \ac{uav} platform.}}
    \label{fig:exp_result}
\end{figure*}

We conducted experiments on the over-actuated \ac{uav} platform that has four 3-\ac{dof} thrust generators to compare the conventional allocation framework and our proposed downwash-aware allocation framework; see \cref{fig:exp_result}. Using the conventional allocation framework, the platform is controlled to track a 90 degree pitch reference trajectory (\cref{fig:four_bad_rpy}), where a pair of downwash effects appear, and an obvious drop in the Z-axis is noticed (\cref{fig:four_bad_xyz}). Although the uniformly high thrust efficiency is maintained by deploying all the thrusts in the same direction, this framework requires more thrust forces to slowly compensate downwash effects with the integrator of position controller (\cref{fig:four_bad_thrust}). Moreover, the stability of the platform is influenced by more oscillations.

Using the proposed downwash-aware allocation framework, the platform avoids the downwash effects by deploying the proper thrust forces and maintains a high thrust efficiency (\cref{fig:four_success_alpha,fig:four_success_beta,fig:four_success_thrust}). Therefore, the position and attitude tracking control performance is guaranteed along whole trajectory (\cref{fig:four_success_xyz,fig:four_success_rpy}). \cref{fig:clips} shows keyframes of the experiment.

\subsection{Discussion}

The minimum downwash avoid distance $o_{\textit{min}}$ in \cref{alg:downwash} has to be experimentally decided for different platforms. $o_{\textit{min}}=0$ means the downwash avoidance is not activated. Large $o_{\textit{min}}$ may result in no feasible solution to the downwash-aware allocation problem. Despite that small $o_{\textit{min}}$ cannot fully avoid downwash flow, it can still improve the control performance to some extent. We chose $o_{\textit{min}}=7cm$ in the experiment.

\section{Conclusion}\label{sec:conclusion}

We presented the downwash-aware control allocation framework of over-actuated \acp{uav}, which makes synergy of downwash effect avoidance and thrust efficiency maintenance. The downwash avoidance constraint and thrust efficiency index were derived and incorporated into the nullspace-based allocation framework. In simulation, the proposed downwash-aware and original nullspace-based allocation frameworks were studied and compared on two different over-actuated platforms. These frameworks were further implemented on our customized \ac{uav} platforms in experiment for demonstration. Both simulation and experiment verified that our proposed framework fully explores the allocation space and finds the desired allocation solution that could both avoid downwash effect and maintain high thrust efficiency, significantly improving the control performance.

\section*{Acknowledgement}
The authors thank Dr. Tengyu Liu, Nan Jiang, Zihang Zhao, Hao Liang, Zeyu Zhang, Zhen Chen, Yifei Dong at BIGAI for discussions and help on hardware design, motion capture system, and figures; Dr. Pengkang Yu at UCLA for his help on control framework of Crazyflie. In particular, Yao Su wants to thank the love, patience, and care from his girlfriend Mengmeng, and wishes the best of her surgery.

\balance
\setstretch{1}
\bibliographystyle{ieeetr}
\bibliography{reference.bib}

\end{document}

%% file: configs.tex
\usepackage{graphicx} \graphicspath{ {figures/} }
\usepackage{amsmath,amssymb,mathabx,amsbsy}
\usepackage{mathtools,etoolbox}
\usepackage{algorithmic}
\usepackage[ruled]{algorithm2e}
\usepackage{acronym}
\usepackage{enumitem}
\usepackage{booktabs}
\usepackage{hyperref}
\usepackage{balance}
\usepackage{xspace,setspace}
\usepackage[skip=3pt,font=small]{subcaption}
\usepackage[skip=3pt,font=small]{caption}
\usepackage[dvipsnames]{xcolor}
\usepackage[capitalise]{cleveref}
\usepackage{tabularx,colortbl,multirow,array,makecell}
\usepackage{overpic}
\usepackage{cite}
\usepackage{tikz}

\makeatletter
\DeclareRobustCommand\onedot{\futurelet\@let@token\@onedot}
\def\@onedot{\ifx\@let@token.\else.\null\fi\xspace}
\def\eg{\emph{e.g}\onedot} 
\def\ie{\emph{i.e}\onedot}

\def\wrt{w.r.t\onedot} 
\def\etal{\emph{et al}\onedot}
\makeatother

\usepackage{algorithm2e}

\makeatletter
\def\BState{\State\hskip-\ALG@thistlm}
\makeatother

\makeatletter
\renewcommand{\paragraph}{%
  \@startsection{paragraph}{4}%
  {\z@}{0ex \@plus 0ex \@minus 0ex}{-1em}%
  {\hskip\parindent\normalfont\normalsize\bfseries}%
}
\makeatother

\crefname{algocf}{alg.}{algs.}
\Crefname{algocf}{Algorithm}{Algorithms}

\definecolor{gblue}{HTML}{4285F4}
\definecolor{gred}{HTML}{DB4437}

\acrodef{qp}[QP]{Quadratic Programming}
\acrodef{fd}[FD]{Force Decomposition}
\acrodef{dof}[DoF]{Degree of Freedom}
\acrodef{ros}[ROS]{Robot Operating System}
\acrodef{uav}[UAV]{Unmanned Aerial Vehicle}
\acrodef{dof}[DoF]{Degree of Freedom}
\acrodef{com}[CoM]{center of mass}